\documentclass{article}

\PassOptionsToPackage{numbers, compress}{natbib}
\usepackage[final]{template/neurips_data_2024}

\usepackage[numbers]{natbib}
\usepackage[utf8]{inputenc} 
\usepackage[T1]{fontenc}    
\usepackage[hyperfootnotes=false]{hyperref}
\usepackage{url}            
\usepackage{booktabs}       
\usepackage{amsfonts}       
\usepackage{nicefrac}       
\usepackage{microtype}      
\usepackage{amsmath}
\usepackage{amssymb}
\usepackage{physics}        
\usepackage{delimset}       

\usepackage{caption}
\usepackage{subcaption}
\usepackage{afterpage}

\usepackage{algorithm}

\usepackage{multirow}
\usepackage{multicol}

\usepackage[toc,page,header]{appendix}
\usepackage{minitoc}

\usepackage{rotating}
\usepackage{longtable}
\usepackage{siunitx}
\usepackage{makecell}

\usepackage{wrapfig}

\usepackage{xcolor,colortbl}
\definecolor{Gray}{gray}{0.90}
\newcolumntype{g}{>{\columncolor{Gray}}c}
\definecolor{ffe1da}{RGB}{255,225,218}
\definecolor{F7E0D5}{RGB}{247,224,213}
\definecolor{darkF7E0D5}{RGB}{209,154,128}
\colorlet{Light}{white!0!F7E0D5}

\usepackage{pifont}
\newcommand{\xmark}{\ding{55}}

\usepackage{glossaries}
\newacronym{ai}{AI}{artificial intelligence}
\newacronym{dl}{DL}{deep learning}
\newacronym{ml}{ML}{machine learning}
\newacronym{ssl}{SSL}{self-supervised learning}
\newacronym{vit}{ViT}{vision transformer}
\newacronym{simclr}{SimCLR}{simple framework for contrastive learning of visual representations}
\newacronym{dino}{DINO}{self-distillation with no labels}
\newacronym{byol}{BYOL}{bootstrap your own latent}
\newacronym{ibot}{iBOT}{image BERT pre-training with online tokenizer}
\newacronym{mae}{MAE}{masked autoencoder}
\newacronym{ap}{AP}{average precision}
\newacronym{auroc}{AUROC}{area under the receiver operating characteristic curve}
\newacronym{fe}{FE}{fraction of effort}
\newacronym{afe}{AFE}{average fraction of effort}
\newacronym{LAD}{LAD}{leaves and distances}
\newacronym{CLAD}{CLAD}{continuous leaves and distances}
\glsunset{simclr}
\glsunset{dino}
\glsunset{mae}
\glsunset{byol}

\newcommand*{\set}[1]{\ensuremath{\mathcal{#1}}}

\renewcommand*{\vec}[1]{\boldsymbol{#1}}

\DeclareMathOperator{\simi}{sim}
\DeclareMathOperator{\dist}{dist}

\title{Intrinsic Self-Supervision for Data Quality Audits}

\author{%
    \textbf{Fabian Gr\"oger\thanks{Correspondence: \texttt{fabian.groeger@unibas.ch}} \ $^{1,2}$,
    \hspace{1mm}Simone Lionetti$^2$,
    \hspace{1mm}Philippe Gottfrois$^1$,
    \hspace{1mm}Alvaro Gonzalez-Jimenez$^1$,} \\ \vspace{0.2em}
    \textbf{Ludovic Amruthalingam$^2$, 
    \hspace{1mm}Labelling Consortium\thanks{Valerie Amann, Elisabeth G\"ossinger, Hazem Juratli, Beda M\"uhleisen, Alina M\"uller, and Veronika Schmidt} \ $^3$, 
    \hspace{1mm}Matthew Groh$^4$,} \\ \vspace{0.2em}
    \textbf{Alexander A.~Navarini\thanks{Joint last authorship} \ $^{1,3}$, 
    \hspace{1mm} Marc Pouly\footnotemark[3] \ $^{2}$} \\ \vspace{0.2em}
    $^1$University of Basel \; $^2$Lucerne University of Applied Sciences and Arts \\[0pt]
    $^3$University Hospital of Basel \; $^4$Northwestern University
}

\begin{document}

\doparttoc 
\faketableofcontents 

\frenchspacing

\maketitle

\begin{abstract}
Benchmark datasets in computer vision often contain off-topic images, near duplicates, and label errors, leading to inaccurate estimates of model performance.
In this paper, we revisit the task of data cleaning and formalize it as either a ranking problem, which significantly reduces human inspection effort, or a scoring problem, which allows for automated decisions based on score distributions.
We find that a specific combination of context-aware self-supervised representation learning and distance-based indicators is effective in finding issues without annotation biases.
This methodology, which we call \mbox{\textsc{SelfClean}}, surpasses state-of-the-art performance in detecting off-topic images, near duplicates, and label errors within widely-used image datasets, such as ImageNet-1k, Food-101N, and STL-10, both for synthetic issues and real contamination.
We apply the detailed method to multiple image benchmarks, identify up to 16\% of issues, and confirm an improvement in evaluation reliability upon cleaning.
The official implementation can be found at: \url{https://github.com/Digital-Dermatology/SelfClean}.
\end{abstract}

\looseness=-1
\section{Introduction}
\label{sec:Intro}
In traditional \gls*{ml}, data cleaning is essential since minor contamination in the dataset can significantly impact model performance and robustness~\citep{li2021cleanml}.
However, with the rise of \gls*{dl} and large-scale datasets, data cleaning has become less crucial as large models have shown to work relatively well even when training data has low quality~\citep{rolnick_deep_2018}.
Validating and cleaning large datasets is challenging, especially for high-dimensional data, because thorough manual verification is often not feasible.
Thus, a lot of research has been focusing on learning from noisy data~\citep{natarajan_learning_2013} rather than fixing quality issues, as the overwhelming benefits of large-scale datasets are believed to exceed the drawback of diminished control.
On the other hand, for many domains, the size of available datasets is still one of the main limiting factors for the progress of \gls*{ai}. 
In these low-data regimes, the importance of clean data is more pronounced since even fractional amounts of poor-quality samples can substantially hamper performance and possibly lead to wrong conclusions~\citep{karimi_deep_2020}.
This is especially relevant in high-stakes settings such as the medical domain, where high-quality data is needed to train robust models and validate their performance.
However, also in these domains, many practitioners rather focus on data quantity as a key performance driver and implicitly assume a high-quality collection process~\citep{pezoulas_medical_2019}.
Thus, even medical datasets are known to contain varying noise levels, which can substantially undermine the progress of \gls*{ml}~\citep{daneshjou_skincon_2022}.

The necessity to report comparable results has led \gls*{dl} practitioners to heavily rely on benchmark datasets despite them being known for containing data quality issues.
For example, an evaluation of ten of the most used benchmark datasets found them to have an average label error rate of 3.4\% in the evaluation set~\citep{northcutt_pervasive_2021}.
Such issues in benchmark sets, especially when used for evaluation, undermine the framework by which scientific progress is measured. 
Specifically, contamination in evaluation sets corrupts scores, making it unclear which methods successfully handle edge cases and obscuring their proximity to optimal performance.
This is particularly relevant since many popular benchmarks are saturating, i.e., only saw minor relative changes in performance over the last few years~\citep{ott_mapping_2022}.
Data quality issues in training sets, instead, may hinder optimization and produce suboptimal models.
Importantly, despite the need for correct evaluation data, cleaning evaluation sets can be problematic, as it may optimistically bias performance estimates. 
Ignoring known data quality issues during evaluation is, however, also incorrect, so an appropriate compromise is necessary.

\looseness=-1
In this paper, we address three types of data quality issues that illustrate these mechanisms well.
\textit{Off-topic samples}, i.e., inputs included in a dataset by mistake, add noise to evaluation metrics while slowing down and confusing training.
\textit{Near duplicates}, i.e., different views of the same object, produce arbitrary re-weighting in the evaluation set, reduce variability in the training set, and most importantly, often introduce leaks between training and evaluation sets that can lead to over-optimistic results.
\textit{Label errors}, i.e., wrongly annotated samples, result in incorrect evaluation and poison the training process.
We focus on these three data quality issues because we empirically found them to be frequent in existing image benchmark datasets and challenging to detect.
There are of course other types of data quality issues, including many that can be detected using ad-hoc rules, such as odd brightness, aspect ratio, resolution, sharpness, and entropy in the case of images.

\begin{figure}
  \centering
  \includegraphics[width=1.0\linewidth]{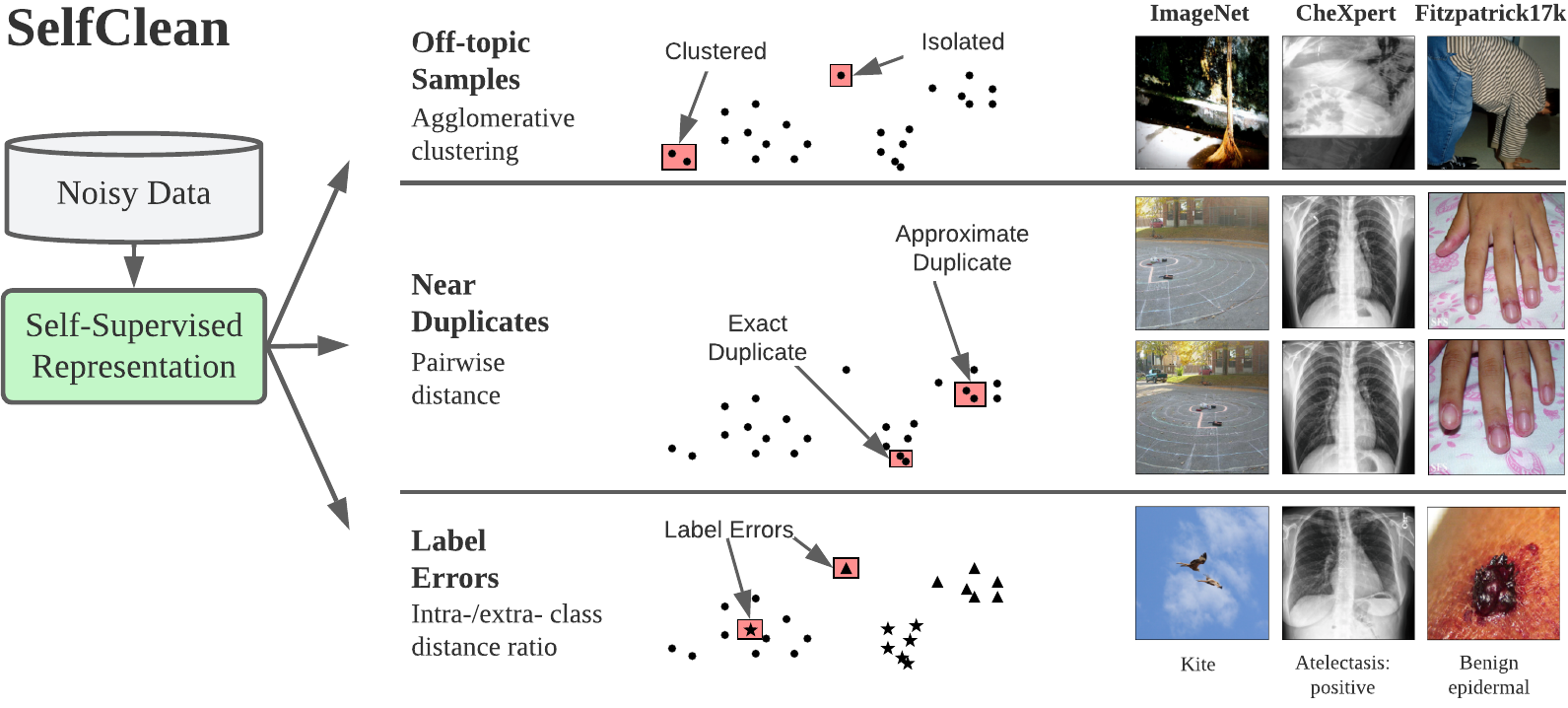}
  \caption{
    \textsc{SelfClean} first trains a self-supervised encoder on noisy data to obtain latent representations for dataset samples. 
    It then detects off-topic samples with agglomerative clustering, near duplicates based on pairwise distances, and label errors using the intra-/extra- class distance~ratio.
  }
  \label{fig:SelfClean}
\end{figure}

\looseness=-1
In this paper, we formulate dataset cleaning as a set of ranking problems, which greatly reduce the effort for manual inspection, or alternatively as a set of scoring problems, which can be used for fully automatic decisions based on score distributions.
We then find that a combination of self-supervised, dataset-specific representation learning and distance-based indicators can effectively identify multiple issues in image collections.
We apply this approach to well-known benchmark datasets in computer vision and medical imaging, and discuss implications for reliability of results across these domains.
The outlined method enables practitioners to audit data collections, increase evaluation reliability, and amend the training set to improve results.
This work contributes to \mbox{data-centric \gls*{ml}}~\citep{jarrahi2022principles} and aims to bolster confidence in both existing and newly collected datasets. 
In summary, the main contributions are:
1) A novel data cleaning procedure called \textsc{SelfClean}, which can be used to find off-topic samples, near duplicates, and label errors, and relies exclusively on the dataset itself, illustrated in figure \ref{fig:SelfClean}.
2) A detailed comparison between this cleaning method and competing approaches on synthetic and natural contamination, including validation against human experts.
3) The application of \textsc{SelfClean} to well-known benchmarks in computer vision and medical imaging and the identification of their issues.
4) A practical recommendation to clean training and evaluation splits of benchmark datasets as a reasonable trade-off between correctness and bias for more accurate performance estimates.

\section{Related work}
\label{sec:Related-Work}
\looseness=-1
Data cleaning is a core component of data analytics and a topic of interest in the data management community \citep{chu_data_2016}.
Recently, the data-centric \gls*{ai} initiative~\cite {jarrahi2022principles} brought it back to the attention of \gls*{ml} researchers, resulting in the development of data cleaning tools.
For instance, \citet{vailoppilly_all_2021} proposed an all-in-one ``data cleansing'' tool based on dimensionality reduction, a \gls*{dl} noise classifier, and a denoising model.
Tools for data cleaning also started to appear,
including CleanLab~\citep{cleanlab2017cleanlab} and CleanVision~\citep{cleanlab2022cleanvision},
Lightly~\citep{susmelj2020lightly}, and FastDup~\citep{visuallayer2022fastdup}.
Most data cleaning approaches require dimensionality reduction to work with high-dimensional data such as images. 
This includes traditional approaches such as PCA \citep{mackiewicz_principal_1993} or t-SNE \citep{maaten_visualizing_2008}, and feature extraction with deep encoders, which are usually trained on natural image databases such as ImageNet \citep{deng_imagenet_2009}. 
In the last few years, \gls*{ssl}~\citep{ozbulak_know_2023} was shown to learn more representative latent spaces compared to supervised training \citep{fernandez_watermarking_2022,wang_understanding_2020,sorscher_beyond_2022}.
Furthermore, \citet{cao_rethinking_2021} demonstrated that \gls*{ssl} can learn meaningful latent spaces even with small datasets, low resolution, and small architectures.
Inspired by these results and unlike previous works, we rely on \gls*{ssl} as a basis to detect three important types of data quality issues encountered in practice: off-topic samples, near duplicates, and label errors \citep{chu_data_2016}.
Since these sub-problems are typically addressed separately in the literature, we briefly review them in turn.

The problem of identifying off-topic samples is closely related to generalized out-of-distribution detection~\citep{yang_generalized_2022} and is akin to outlier detection, which involves both normal and anomalous samples~\citep{aggarwal_outlier_2017}.
Outlier detection can be addressed with supervised, unsupervised, and semi-supervised learning and was initially developed to fit data more smoothly \citep{boukerche_outlier_2020}.
In the realm of data cleaning, where the nature of off-topic samples is generally unknown, it is most similar to the unsupervised setting.
Outliers in low-density regions can be found using reconstruction errors~\citep{hendrycks_deep_2019,abati_latent_2019}, classification~\citep{ruff_deep_2018}, or probabilistic approaches~\citep{ren_likelihood_2019}.
For a detailed review of these methods, see~\cite{aggarwal_outlier_2017}.

Near-duplicate detection is traditionally based on representation matching \citep{lowe_distinctive_2004, ke_efcient_2004}.
Most \gls*{dl} approaches follow a similar strategy, where feature vectors are extracted by a deep network and used for content-based matching \citep{babenko_neural_2014}. 
Another option is to learn a similarity metric between samples with Siamese neural networks \citep{zbontar_computing_2015}. 
A recent approach for copy detection (i.e., near-duplicate detection)  uses a contrastive self-supervised objective with entropy regularization to ensure consistent separation of image descriptions \citep{pizzi_self-supervised_2022}. 
However, it requires a manually adapted threshold for each dataset \citep{oquab_dinov2_2023}.

The identification of label errors is generally focused on prediction-label agreement via confusion matrices and proceeds by removing samples with low recognition rate \citep{chen_understanding_2019} or parts of the minority classes \citep{zhang_imagedc_2020}. 
There are exceptions, such as recent approaches based on supervised contrastive learning for label error correction \citep{huang_contrastive_2022, lee_augment_2021}. 
Another prominent method is confident learning, which identifies label errors based on noisy data pruning, using probabilistic thresholds to estimate noise and ranking examples to train with confidence \citep{northcutt_confident_2022}.

\section{Methodology}
\label{sec:Methodology}

Let $\set{X} = \{(\vec{x}_i, l_i) \ | \ i \in\set{I}\}$ be an image classification dataset to be cleaned, where $\set{I} = \{1, \ldots, N\}$ is the index set, $\vec{x}_i$ is the $i$-th sample, and $l_i \in \{1, \ldots, L\}$ is the $i$-th label.
For each issue type, we construct a scoring function $s$ that assigns values in $[0, 1]$ to samples or pairs thereof, such that elements with a lower score are more likely to be problematic.
Sorting samples by the value obtained from the scoring function $s$ induces a ranking $R$ where more likely issues appear earlier.

\subsection{Representation learning}
As a first step, we train a deep feature extractor $f$ with parameters $\theta$ on the dataset $\set{X}$ using \acrfull*{ssl}, which learns representations by solving auxiliary tasks.
Let $\vec{e}_i = f(\vec{x}_i; \theta)\in \mathbb{R}^D$ be the representation of sample $\vec{x}_i$ obtained with $f$, where $D$ denotes the latent dimension. 
Note that \gls*{ssl} is performed on the entire dataset including data quality issues.
Any \gls*{ssl} method can be used, as investigated in appendix \ref{app:Influence-SSL}.
Here, we consider \gls*{simclr}~\citep{chen_simple_2020} and \gls*{dino}~\citep{caron_emerging_2021}, which were shown to produce meaningful latent spaces \citep{fernandez_watermarking_2022,wang_understanding_2020}.
\gls*{simclr} is a contrastive approach that compares different views of the same image against other randomly sampled ones.
\Gls*{dino} is a self-distillation method which trains a student network to match a teacher network on different views of the same image. 
For both strategies, we rely on \gls*{vit} encoders, as detailed in appendix \ref{app:Training-Details} and ablated in \ref{app:Influence-Encoder}.

As feature normalization is often built into the \gls*{ssl} training objective, it is natural to compare points in its latent space using cosine similarity, $\simi(\vec{e}_i, \vec{e}_j) = \vec{e}_i^\top \vec{e}_j / (||\vec{e}_i||_2 ||\vec{e}_j||_2)$, and the associated distance scaled to $[0, 1]$, $\dist(\vec{e}_i, \vec{e}_j) = (1 - \simi(\vec{e}_i, \vec{e}_j)) / 2$.
We explicitly include $L_2$-normalization during training and inference for strategies without normalization (e.g., \gls*{dino}), such that their latent space is a unit hypersphere of dimension $D-1$.
In appendix \ref{app:Influence-L2-Distance}, we present an ablation study of this normalization and investigate the influence of different distance functions. 

\subsection{Distance-based indicators}\label{sub:Dist-Indicators}
Dataset-specific representations based on inductive bias can be coupled with separate distance-based indicators to identify candidate issues.
Below we introduce each issue type and the corresponding indicator function used to detect them.

\looseness=-1
\textbf{Off-topic samples.}
We define samples as off-topic when they are included in the dataset by mistake.
Images from extraneous modalities, affected by device malfunctions, or without any object of interest are some examples.
Atypical samples, due e.g. to the phenomenon of hidden stratification \citep{oakden2020hidden}, that are included intentionally, are not off-topic, and although they may be revealed in the same search, they require different treatment.
We achieve off-topic sample ranking by agglomerative clustering with single linkage \citep{gower_minimum_1969} in representation space.
The idea is that the later a cluster is merged with a larger one, the more it can be considered an outlier \citep{jiang_two-phase_2001}.
The ranking is obtained by sorting the clustering dendrogram such that, at each merge, the elements of the cluster with fewer leaves appear first.
We also associate each sample with a numerical score, which takes small values for abnormal instances and is compatible with the described ranking.
In appendix \ref{app:irrelevantscoring}, we construct such a score $s_{\text{OT}}(\vec{e}_i)$ starting from the idea that merges, which happen at very different distances or between clusters of very different sizes, should produce large numerical variations.

\textbf{Near duplicates.}
We define near duplicates as pairs of images that contain different views of the same object.
In this sense, exact duplicates are a special case of near duplicates. 
We rank potential near duplicates by sorting each pair of distinct samples $(i, j), i<j$ in ascending order according to the distance between their representations in the latent space, $s_{\text{ND}}(\vec{e}_i, \vec{e}_j)=\dist(\vec{e}_i, \vec{e}_j)$. 

\textbf{Label errors.}
We define label errors as samples annotated with a wrong class label. 
We rank potential label errors by sorting samples in ascending order according to their intra-/extra- class distance ratio~\citep{ho_complexity_2002}.
For an anchor point $\vec{e}_i$, this ratio compares the distances to the nearest representation of a different label $m_{{\scriptscriptstyle \neq}}(\vec{e}_i)$ and the distance to the nearest representation of the same label $m_{{\scriptscriptstyle =}}(\vec{e}_i)$:
\begin{equation}
    \begin{array}{l}
    m_{{\scriptscriptstyle =}}(\vec{e}_i) = \min_{j\in\set{I},\,l_j=l_i} \brk[s]1{\dist(\vec{e}_i, \vec{e}_j)},\\[6pt]
    m_{{\scriptscriptstyle \neq}}(\vec{e}_i) = \min_{j\in\set{I},\,l_j\neq l_i} \brk[s]1{\dist(\vec{e}_i, \vec{e}_j)},
    \end{array}
    \qquad
    s_{\text{LE}}(\vec{e}_i) =
    \frac{m^2_{{\scriptscriptstyle \neq}}(\vec{e}_i)}{m^2_{{\scriptscriptstyle =}}(\vec{e}_i) + m^2_{{\scriptscriptstyle \neq}}(\vec{e}_i)}. 
\end{equation}

In all three cases, \textsc{SelfClean} leverages the local structure of the embedding space:
Cluster distances are computed only using the closest samples during agglomeration for off-topic samples, near duplicates are identified among sample pairs with the smallest distances, and label errors are found using only the nearest examples of the same and a different class.

\subsection{Operation modes}
\label{sub:AutoCleaning}
The criteria above rank and score candidate issues, but do not specify which ones are inferred to be actual issues.
This can be achieved with two operating modes: Human-in-the-loop or fully automatic.

\looseness=-1
\textbf{Human-in-the-loop.}
This mode leverages candidate issue rankings to facilitate human confirmation which is often infeasible exhaustively, especially when considering pairwise relationships such as near duplicates.
A human curator inspects a data sequence where issues tend to appear earlier, either confirming and correcting problems or looking for a specific rank threshold that gives the desired balance between precision and recall.
In appendix \ref{app:InspectionEffortSaved}, we estimate that for a typical dataset \textsc{SelfClean} reduces this inspection effort by a factor between 5 and 50 depending on issue type and baseline.

\looseness=-1
\textbf{Fully-automatic.}
To perform automatic cleaning, specifying a fraction of data quality issues \emph{a priori} is suboptimal, as contamination is not easy to estimate.
The scores of section \ref{sub:Dist-Indicators} empirically produce a smooth distribution for clean samples and relegate contaminated ones to significantly lower values.
Depending on the contaminated data distribution, it may then be possible to isolate problematic samples with statistical arguments based on two robust hyperparameters, the contamination rate guess~$\alpha$ and the significance level~$q$, as detailed in appendix \ref{app:automaticcleaning}.
In short, we first use a logit transformation to induce a gap between scores of normal and problematic samples.
We then set an upper bound for the left tail of the score distribution using a logistic functional form, and estimate its parameters using quantiles.
Afterward, we identify issues based on their violation of the upper probability bound.

\section{Experimental setup}
\label{sec:experimentalsetup}

\begin{figure}
  \centering
  \includegraphics[width=1.0\linewidth]{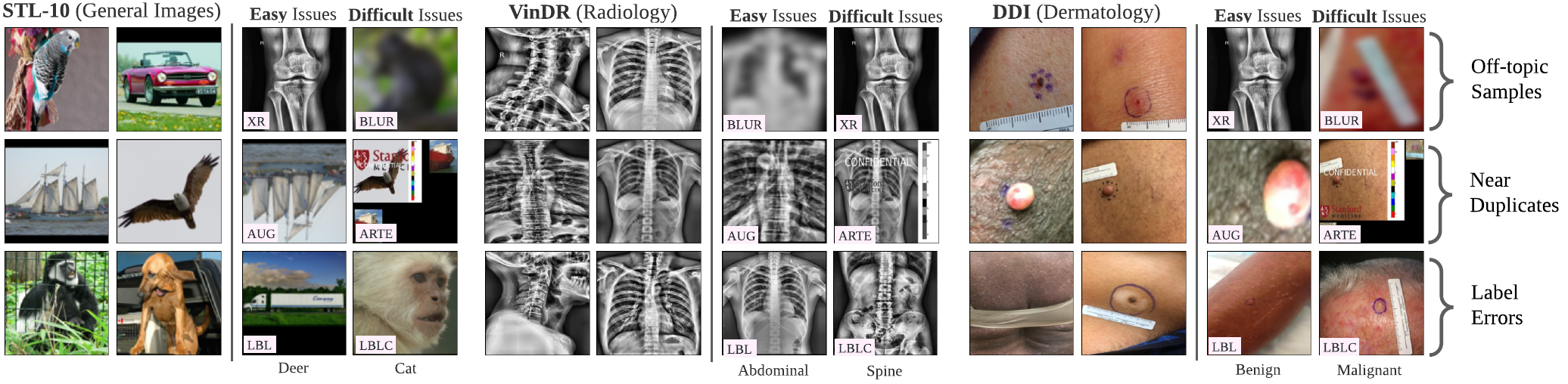}
  \caption{
    Illustration of synthetic data quality issues of all three types in STL-10, VinDR, and DDI.
  }
  \label{fig:ExperimentDesign}
\end{figure}

\textbf{Datasets.}
We experiment on a total of twelve datasets described in appendix \ref{app:Datasets}.
These are four large-scale vision benchmarks: ImageNet~\citep{deng_imagenet_2009}, STL-10~\citep{coates_analysis_2011}, CelebA~\citep{liu2015faceattributes}, and Food-101N~\citep{bossard14}, three general medical datasets of X-rays and histopathological images: CheXpert~\citep{irvin_chexpert_2019}, VinDr-BodyPartXR~\cite{pham_dicom_2021}, and PatchCamelyon~\citep{ ehteshami_bejnordi_diagnostic_2017}, and five dermatology datasets: HAM10000~\citep{tschandl_ham10000_2018}, ISIC-2019~\citep{tschandl_ham10000_2018}, Fitzpatrick17k~\citep{groh_evaluating_2021}, DDI~\citep{daneshjou_disparities_2022}, and PAD-UFES-20~\citep{pacheco_pad-ufes-20_2020}.

\textbf{Evaluation metrics.}
The evaluation in this work relies on ranking metrics, as ranking constitutes the core of \textsc{SelfClean} independently of the operation mode. 
All approaches are therefore evaluated in terms of the \gls*{auroc} and \mbox{\gls*{ap}} following standard practice \citep{rendle_evaluation_2019}. 
\Gls*{auroc} measures the likelihood that a random relevant sample is ranked higher than a random irrelevant sample. 
\Gls*{ap} measures precision across all values of recall, and is therefore sensitive to the proportion of positive and negative samples. 

\textbf{Synthetic experiment setup.}
\looseness=-1
To compare \textsc{SelfClean} against other methods, we create synthetic datasets by altering benchmarks of different modalities (i.e., STL-10, VinDr-BodyPartXR, and DDI), as illustrated in figure \ref{fig:ExperimentDesign}.
These synthetic contaminations are inspired by typical issues present in the respective dataset domains.
We consider 5\% and 10\% contamination to mimic real-world noise prevalence estimates~\citep{han_adbench_2022}.
For each issue type, we compare against other unsupervised methods that have performed well on the given task. 
A detailed description of these competing approaches can be found in appendix \ref{app:Competing-Approaches}.
Since \textsc{SelfClean} learns representations on the contaminated dataset, we train a separate encoder for every issue type, contamination level, and synthetic contamination strategy.

The first synthetic contamination strategy for off-topic samples, \textit{XR}, adds images from the ``other'' category of VinDr-BodyPartXR \citep{pham_dicom_2021}, which shows scans of lower limbs and device malfunctions.
The second strategy for off-topic samples, \textit{BLUR}, corrupts images with strong Gaussian blurring to simulate badly out-of-focus pictures.
The first contamination strategy for near duplicates, \textit{AUG}, adds samples from the original dataset after augmenting them with rotation, flipping, resizing, padding, and blurring.
The second approach for near duplicates, \textit{ARTE}, adds samples from the original dataset after including artifacts such as watermarks, color bars, and rulers, followed by scaling and composition with other images to create a collage.
For label errors, the first contamination strategy, \textit{LBL}, randomly changes a fraction of the labels choosing uniformly from incorrect ones. 
The second strategy to evaluate label errors, \textit{LBLC}, randomly changes a fraction of the labels choosing incorrect ones proportionally to class prevalence in the original dataset.
Depending on which dataset these strategies are applied to, they produce either easy or difficult problematic samples.

Different contamination strategies can be applied sequentially to create a dataset with a more realistic constellation of artificial data quality issues, resulting in a mixed-contamination strategy.
In order to consider all interactions, we start by adding off-topic samples, proceed by creating near duplicates, and finally introduce label errors.
To preserve the overall contamination rate $C$, each contamination in the sequence is added with prevalence $C_S$ such that $(1+C_S)^S=(1+C)$, where $S$ is the number of contamination steps.

\textbf{Natural experiment setup.}
We also evaluate cleaning on data quality issues naturally found in benchmark datasets.
To this end, we devise two different experiments. 
In the first experiment, we measure how well the ranking matches available metadata,
e.g., if two images show the same person or if the label was already identified as incorrect by prior work.
In a second experiment, we use \textsc{SelfClean} to propose a ranking for some datasets and evaluate it against human confirmation of issues with the statistical procedure outlined in appendix \ref{app:Human-Verification}.

\begin{table*}
    \centering
    \tiny
    \caption{
        \looseness=-1
        Performance in detecting synthetic data quality issues. 
        Evaluation is performed for each of the three considered issue types across three benchmark datasets,
        augmented with two strategies for 5\% synthetic contamination each, as illustrated in figure \ref{fig:ExperimentDesign}.
        Consult section \ref{sec:experimentalsetup} for more details on the contamination, and appendix \ref{app:Competing-Approaches} for details on competing approaches.
        Results are given in percentages (\%).
    }
    \setlength{\tabcolsep}{4pt}
    \label{tab:Results}
    \resizebox{1.0\linewidth}{!}{%
    \begin{tabular}{@{} l ll rr rr rr rr rr rr}%
        \toprule
        \parbox[t]{2mm}{\multirow{8}{*}{\rotatebox[origin=c]{90}{Off-topic Samples}}}
        &
        & & \multicolumn{2}{c}{\bfseries STL + XR}
            & \multicolumn{2}{c}{\bfseries STL + BLUR} 
            & \multicolumn{2}{c}{\bfseries VDR + BLUR}
            & \multicolumn{2}{c}{\bfseries VDR + XR} 
            & \multicolumn{2}{c}{\bfseries DDI + XR} 
            & \multicolumn{2}{c}{\bfseries DDI + BLUR} \\

        \cmidrule(lr){4-5}
        \cmidrule(lr){6-7}
        \cmidrule(lr){8-9}
        \cmidrule(lr){10-11}
        \cmidrule(lr){12-13}
        \cmidrule(lr){14-15}
            
        & \textbf{Method} & \textbf{Rep.}
            & AUROC & AP
            & AUROC & AP
            & AUROC & AP
            & AUROC & AP
            & AUROC & AP
            & AUROC & AP \\
        \cmidrule{2-15}
            & HBOS \citep{goldstein_histogram-based_2012}
            & INet
                & 66.9 & 6.6
                & 1.9 & 2.6
                & 95.7 & 36.6
                & 82.3 & 24.4
                & 93.0 & 68.0
                & 19.0 & 3.0 \\
            & ECOD \citep{li_ecod_2022}
            & INet
                & 68.4 & 7.0
                & 2.2 & 2.6
                & 95.0 & 34.1
                & 81.4 & 25.7
                & 92.8 & 68.0
                & 23.6 & 3.1 \\    
        & \cellcolor{Light}\textsc{SelfClean}
        & \cellcolor{Light}INet
                & \cellcolor{Light}11.4 & \cellcolor{Light}2.7
                & \cellcolor{Light}67.7 & \cellcolor{Light}7.3
                & \cellcolor{Light}99.9 & \cellcolor{Light}91.2
                & \cellcolor{Light}77.1 & \cellcolor{Light}32.8
                & \cellcolor{Light}98.9 & \cellcolor{Light}84.2
                & \cellcolor{Light}86.5 & \cellcolor{Light}18.2 \\
        & \cellcolor{Light}\textsc{SelfClean}
        & \cellcolor{Light}SimCLR
                & \cellcolor{Light}40.6 & \cellcolor{Light}3.9
                & \cellcolor{Light}77.4 & \cellcolor{Light}19.0
                & \cellcolor{Light}\textbf{100.0} & \cellcolor{Light}98.7
                & \cellcolor{Light}86.0 & \cellcolor{Light}35.5
                & \cellcolor{Light}99.0 & \cellcolor{Light}68.9
                & \cellcolor{Light}70.0 & \cellcolor{Light}21.9 \\
        & \cellcolor{Light}\textsc{SelfClean} 
        & \cellcolor{Light}DINO
                & \cellcolor{Light}\textbf{98.4} & \cellcolor{Light}\textbf{55.1}
                & \cellcolor{Light}\textbf{100.0} & \cellcolor{Light}\textbf{97.9}
                & \cellcolor{Light}\textbf{100.0} & \cellcolor{Light}\textbf{100.0}
                & \cellcolor{Light}\textbf{95.6} & \cellcolor{Light}\textbf{53.3}
                & \cellcolor{Light}\textbf{100.0} & \cellcolor{Light}\textbf{100.0}
                & \cellcolor{Light}\textbf{86.8} & \cellcolor{Light}\textbf{32.6} \\
        \midrule
        \parbox[t]{2mm}{\multirow{8}{*}{\rotatebox[origin=c]{90}{Near Duplicates}}} & 
        & & \multicolumn{2}{c}{\bfseries STL + AUG}
            & \multicolumn{2}{c}{\bfseries STL + ARTE}
            & \multicolumn{2}{c}{\bfseries VDR + AUG}
            & \multicolumn{2}{c}{\bfseries VDR + ARTE}
            & \multicolumn{2}{c}{\bfseries DDI + AUG} 
            & \multicolumn{2}{c}{\bfseries DDI + ARTE} \\

            \cmidrule(lr){4-5}
            \cmidrule(lr){6-7}
            \cmidrule(lr){8-9}
            \cmidrule(lr){10-11}
            \cmidrule(lr){12-13}
            \cmidrule(lr){14-15}
            
        & &
            & AUROC & AP
            & AUROC & AP
            & AUROC & AP
            & AUROC & AP
            & AUROC & AP
            & AUROC & AP \\
        \cmidrule{2-15}
            & pHashing \citep{marr_theory_1997}
            &
                & 57.8 & $<$ 0.1
                & 73.1 & 20.1
                & 47.5 & $<$ 0.1
                & 57.5 & 18.2
                & 59.4 & 0.1
                & 66.2 & 15.1 \\
            & SSIM \citep{wang_image_2004}. 
            &
                & 62.5 & 0.2
                & 83.6 & 19.9
                & 46.3 & $<$ 0.1
                & 48.4 & \textbf{22.5}
                & 57.6 & 0.2
                & 83.0 & 19.4 \\
        & \cellcolor{Light}\textsc{SelfClean}
        & \cellcolor{Light}INet
                & \cellcolor{Light}96.6 & \cellcolor{Light}7.6
                & \cellcolor{Light}96.5 & \cellcolor{Light}15.2
                & \cellcolor{Light}79.7 & \cellcolor{Light}$<$ 0.1
                & \cellcolor{Light}53.7 & \cellcolor{Light}11.1
                & \cellcolor{Light}97.6 & \cellcolor{Light}4.1
                & \cellcolor{Light}81.1 & \cellcolor{Light}34.4 \\
        & \cellcolor{Light}\textsc{SelfClean}
        & \cellcolor{Light}SimCLR
                & \cellcolor{Light}86.1 & \cellcolor{Light}0.1
                & \cellcolor{Light}93.8 & \cellcolor{Light}13.9
                & \cellcolor{Light}76.1 & \cellcolor{Light}$<$ 0.1
                & \cellcolor{Light}78.9 & \cellcolor{Light}12.6
                & \cellcolor{Light}89.8 & \cellcolor{Light}1.6
                & \cellcolor{Light}87.2 & \cellcolor{Light}0.7 \\
        & \cellcolor{Light}\textsc{SelfClean}
        & \cellcolor{Light}DINO
                & \cellcolor{Light}\textbf{100.0} & \cellcolor{Light}\textbf{43.7}
                & \cellcolor{Light}\textbf{99.9} & \cellcolor{Light}\textbf{48.0}
                & \cellcolor{Light}\textbf{98.5} & \cellcolor{Light}\textbf{0.4}
                & \cellcolor{Light}\textbf{91.6} & \cellcolor{Light}16.8
                & \cellcolor{Light}\textbf{99.7} & \cellcolor{Light}\textbf{50.8}
                & \cellcolor{Light}\textbf{98.2} & \cellcolor{Light}\textbf{48.2} \\
        \midrule
        \parbox[t]{2mm}{\multirow{8}{*}{\rotatebox[origin=c]{90}{Label Errors}}} 
        & 
            
        & & \multicolumn{2}{c}{\bfseries STL + LBL} 
            & \multicolumn{2}{c}{\bfseries STL + LBLC}
            & \multicolumn{2}{c}{\bfseries VDR + LBL}
            & \multicolumn{2}{c}{\bfseries VDR + LBLC}
            & \multicolumn{2}{c}{\bfseries DDI + LBL}
            & \multicolumn{2}{c}{\bfseries DDI + LBLC} \\

            \cmidrule(lr){4-5}
            \cmidrule(lr){6-7}
            \cmidrule(lr){8-9}
            \cmidrule(lr){10-11}
            \cmidrule(lr){12-13}
            \cmidrule(lr){14-15}
            
            & &
            & AUROC & AP
            & AUROC & AP
            & AUROC & AP
            & AUROC & AP
            & AUROC & AP
            & AUROC & AP \\
        \cmidrule{2-15}
            & CLearning \citep{northcutt_confident_2022}
            & INet
                & 86.2 & 41.6
                & 83.2 & 36.8
                & 96.7 & 79.0
                & 96.8 & 74.9
                & 67.9 & 11.0
                & 75.0 & 12.9 \\
            & FastDup \citep{visuallayer2022fastdup}
            & INet
                & 87.5 & 20.5
                & 87.0 & 19.8
                & 95.0 & 38.9
                & 94.1 & 37.8
                & 69.0 & 8.6
                & 69.9 & 11.6 \\
        & \cellcolor{Light}\textsc{SelfClean}
        & \cellcolor{Light}INet
                & \cellcolor{Light}\textbf{97.7} & \cellcolor{Light}\textbf{77.6}
                & \cellcolor{Light}\textbf{97.9} & \cellcolor{Light}\textbf{76.4}
                & \cellcolor{Light}98.5 & \cellcolor{Light}84.6
                & \cellcolor{Light}98.5 & \cellcolor{Light}84.8
                & \cellcolor{Light}67.8 & \cellcolor{Light}11.6
                & \cellcolor{Light}\textbf{79.8} & \cellcolor{Light}18.3 \\
        & \cellcolor{Light}\textsc{SelfClean}
        & \cellcolor{Light}SimCLR
                & \cellcolor{Light}79.1 & \cellcolor{Light}27.4
                & \cellcolor{Light}77.4 & \cellcolor{Light}26.5
                & \cellcolor{Light}95.0 & \cellcolor{Light}62.2
                & \cellcolor{Light}95.4 & \cellcolor{Light}64.4
                & \cellcolor{Light}64.8 & \cellcolor{Light}8.3
                & \cellcolor{Light}69.0 & \cellcolor{Light}11.1 \\
        & \cellcolor{Light}\textsc{SelfClean}
        & \cellcolor{Light}DINO
                & \cellcolor{Light}90.7 & \cellcolor{Light}54.2
                & \cellcolor{Light}91.1 & \cellcolor{Light}48.3
                & \cellcolor{Light}\textbf{99.2} & \cellcolor{Light}\textbf{88.1}
                & \cellcolor{Light}\textbf{99.0} & \cellcolor{Light}\textbf{85.6}
                & \cellcolor{Light}\textbf{71.4} & \cellcolor{Light}\textbf{13.5}
                & \cellcolor{Light}71.7 & \cellcolor{Light}\textbf{21.4} \\
        \bottomrule
    \end{tabular}
    }
\end{table*}
\begin{figure}
  \centering
  \vspace*{-0.9em}
  \includegraphics[width=1.0\linewidth]{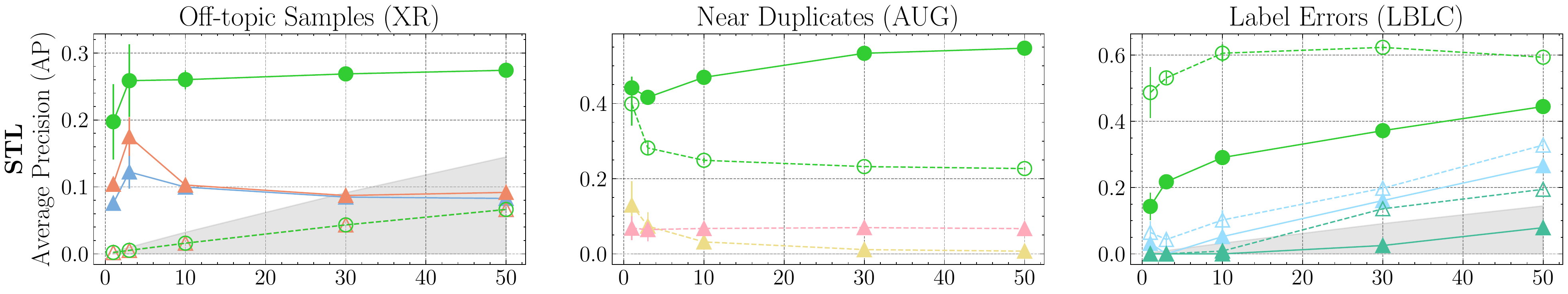}
  \includegraphics[width=1.0\linewidth]{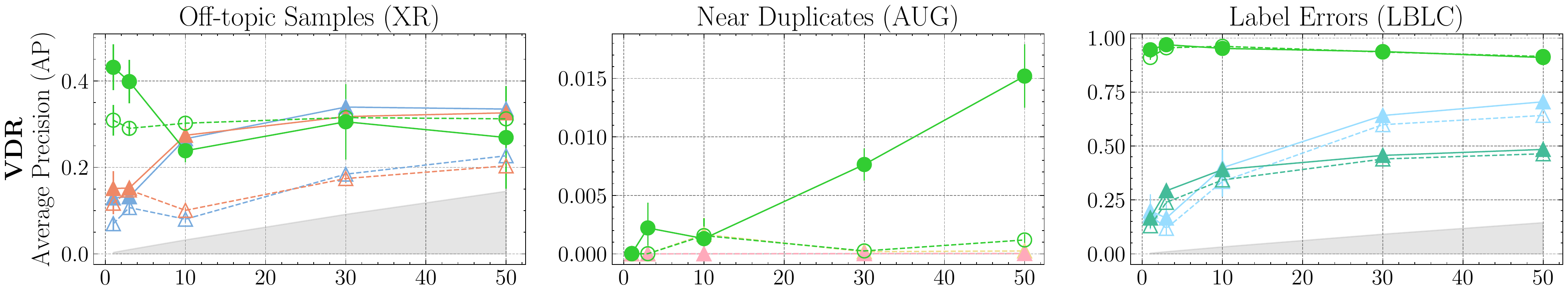}
  \includegraphics[width=1.0\linewidth]{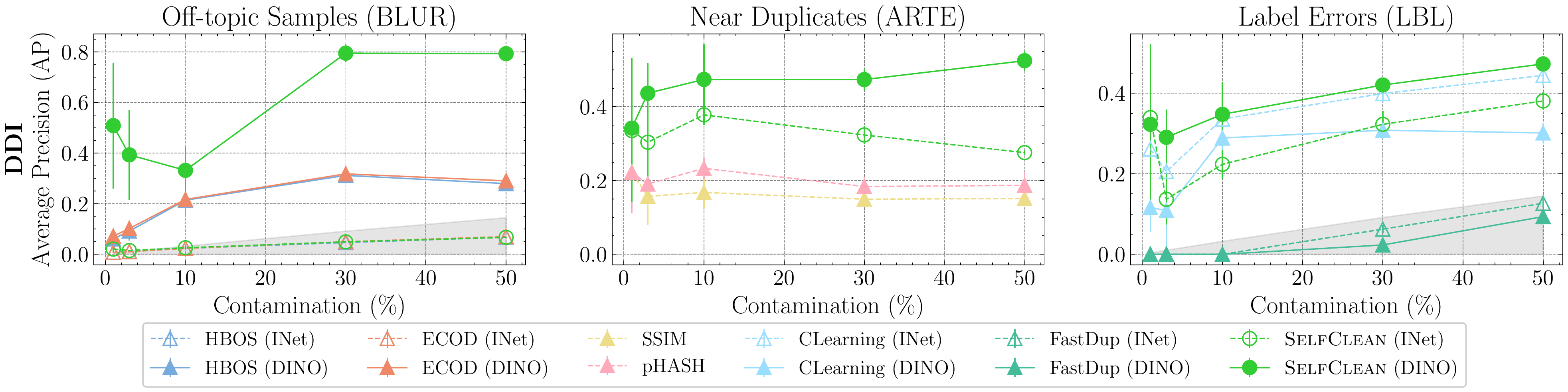}
  \caption{
      Performance of the best two approaches for each issue type to \textsc{SelfClean} across different representations for a mixed-contamination strategy at varying contamination rates.
      Gray regions indicate random performance with an \gls*{ap} equal to the respective contamination $C_S$.
  }
  \label{fig:Contamination-Experiment}
\end{figure}

\section{Results}

\subsection{Synthetic contamination}
\label{ssec:syntheticresults}

\looseness=-1
\textbf{Comparison on data quality issues.}
Table \ref{tab:Results} displays the results of \textsc{SelfClean} using either supervised ImageNet (INet), \gls*{simclr}, or \gls*{dino} pre-training, and the two best competing methods per issue type.
Performance is reported for 18 synthetic datasets based on general vision (STL), radiology (VDR), and dermatology (DDI) benchmarks described in section \ref{sec:experimentalsetup} with a contamination rate of 5\%.
Table \ref{tab:Results-Detailed} in appendix \ref{app:syntheticcleaningtable} includes results for all competing approaches for both 5\% and 10\% contamination.
\textsc{SelfClean} with \gls*{dino} pre-training outperforms all competing methods for off-topic-sample, near-duplicate, and label-error detection. 
Notably, some competing approaches for off-topic-sample detection show varying performance depending on the considered outlier type.
In contrast, \textsc{SelfClean} does not show the same behavior, mainly because the dataset-specific pre-training captures the context of the task itself.
\Gls*{simclr} and supervised ImageNet features achieve mixed performance depending on the specific dataset and issue type.
Lower performance of \gls*{simclr} is presumably caused by the small dataset size, as the batch size cannot be large enough, which is crucial for the contrastive approaches.
\Gls*{ap} for VDR with AUG is very low, likely because these synthetic issues are difficult in highly standardized settings and the dataset is not particularly clean, as further investigated in \ref{app:Investigation-VDR}.

\textbf{Influence of contamination.}
Figure \ref{fig:Contamination-Experiment} illustrates the influence of the contamination on \textsc{SelfClean} and the best two competing models.
For approaches operating on features, we compare performance using both supervised INet and self-supervised, dataset-specific \gls*{dino} training.
Central value and error bars are obtained from three random initializations resulting in different synthetic datasets.
This experiment is run on mixed-contamination datasets.
\textsc{SelfClean} outperforms competing approaches across contamination rates.
The exception is off-topic detection for VDR with high contamination, where other indicator functions on dataset-specific \gls*{ssl} features perform marginally better.
In general, dataset-specific image representations tend to outperform general-purpose ones across tasks.
For label error detection on STL, \textsc{SelfClean} performs significantly better with INet features than with \gls*{dino} features, presumably because INet features are trained with supervision on data and labels similar to STL.

\subsection{Natural contamination}
\label{ssec:naturalresults}

\textbf{Comparison with metadata.}
We validate the label error ranking in a more realistic setting using annotations from the literature, such as 5,440 verified samples of ImageNet's validation set \citep{northcutt_pervasive_2021} and 57,608 of Food-101N \citep{lee_cleannet_2018}.
\textsc{SelfClean} achieves almost double the performance in \gls*{ap} for both datasets compared to other approaches, with 8.4\% vs. 4.3\% \gls*{ap} for ImageNet and 47.8\% vs. 30.7\% for Food-101N.
We evaluate near-duplicate detection against CelebA labels that indicate images of the same celebrity.
\textsc{SelfClean} achieves 30.9\% \gls*{ap}, demonstrating it effectively learned facial recognition without supervision.
For medical datasets, we first check how well \textsc{SelfClean} can find pairs of images showing the same skin lesion.
We obtain good correspondence for HAM10000 and ISIC-2019, with an \gls*{ap} of 28.4\% and 26.6\%, respectively.
On the other hand, for PAD-UFES-20 \gls*{ap} is only 10.0\%, which we further investigate in appendix \ref{app:Detailed-Meta} and is likely caused by inaccurate metadata.
We also attempt to identify X-rays from the same patient within CheXpert and find only minor agreement with 7.5\% \gls*{ap}, suggesting again that a case-by-case investigation should be performed.
Overall, this shows that the rankings produced by \textsc{SelfClean} align with existing metadata and considerably outperform competitors.
A table with detailed results can be found in appendix \ref{app:Detailed-Meta}.

\looseness=-1
\textbf{Comparison with human annotators.}
We evaluate \textsc{SelfClean} rankings against human verification across two common vision and two medical benchmarks as described in appendix \ref{app:Human-Verification}.
Human experts confirmed significantly more data quality issues in the top 50 images ranked by \textsc{SelfClean} compared to 50 randomly sampled images, with 95\% significance in nine out of twelve tasks (table~\ref{app:HumanValidation-Table}).
We repeat the comparison for images ranked 1-25 against images ranked 26-50 and observe significance for six out of ten evaluations.
Two cases in the second comparison are excluded as only containing positive samples (i.e., data quality issues) results in undefined metrics.
These results indicate that \textsc{SelfClean} rankings align well with human assessment for these three issue types.

\setlength{\fboxsep}{2pt}
\begin{wraptable}{R}{.5\linewidth}
    \vspace{-\intextsep}
    \centering
    \caption{
        Ablation of pre-training choices in \textsc{SelfClean}.
        The upper part investigates \gls*{ssl} objective and dataset, and the lower the influence of \gls*{ssl} augmentations.
        For the different variants (lower part), we \colorbox{Light}{highlight} the differences from the default setting.
        We use a 10\% mixed-contamination dataset starting from STL and creating off-topic samples (OT) using XR, near duplicates (ND) using AUG, and label errors (LE) using LBLC.
        Performance is reported in \acrfull*{ap}.
    }
    \small
    \label{tab:Ablation}
    \setlength{\tabcolsep}{4pt}
    \resizebox{1.0\linewidth}{!}{%
    \begin{tabular}{@{} cc @{\hspace{5mm}} rrr @{}}
    \toprule
        \multicolumn{2}{@{} l}{Pre-training strategy} 
            & OT (\%) & ND (\%) & LE (\%) \\
    \midrule
        \multicolumn{2}{@{} l}{\textsc{SelfClean} (Sup. INet)} & 1.6 & 24.6 & 63.0 \\
        \multicolumn{2}{@{} l}{\textsc{SelfClean} (DINO INet)} & 13.7 & 6.1 & \textbf{69.5} \\
        \multicolumn{2}{@{} l}{\textsc{SelfClean} (DINO STL)} & \textbf{27.4} & \textbf{47.1} & 24.8 \\
    \midrule
    \midrule
	    Color+Size & Multi-Crop 
            & OT (\%) & ND (\%) & LE (\%) \\
    \midrule
        \multicolumn{5}{@{}l}{\textsc{SelfClean} (DINO)}\\
	    \checkmark & \checkmark & \textbf{27.4} & 47.1 & 24.8 \\
	    \checkmark & \colorbox{Light}{\xmark} & 2.8 & 17.5 & \textbf{39.6} \\
	    \colorbox{Light}{\xmark} & \checkmark & 4.2 & \textbf{67.2} & 12.6 \\
    \midrule
        \multicolumn{5}{@{}l}{\textsc{SelfClean} (SimCLR)}\\
	    \checkmark & \colorbox{Light}{\checkmark} & \textbf{39.1} & 12.8 & \textbf{18.1} \\[1pt]
	    \checkmark & \xmark & 26.1 & 12.1 & 15.8 \\
	    \colorbox{Light}{\xmark} & \colorbox{Light}{\checkmark} & 3.9 & \textbf{21.9} & 11.7 \\
    \bottomrule
  \end{tabular}
  }
  \vspace{-\intextsep}
\end{wraptable}

\newpage
\looseness=-1
\subsection{Influence of representation learning} \label{ssec:ablation}
Table \ref{tab:Ablation} examines the influence of \gls*{ssl} objective, dataset, and augmentation on \textsc{SelfClean} by measuring performance on STL.
In the upper panel, we observe that dataset-specific representations (\mbox{DINO STL}) yield the best results for both off-topic and near duplicate detection, showcasing the strength of learning the dataset context.
This is remarkable considering that STL has only 5,000 samples compared to the 1 million available for ImageNet.
Label error detection seems instead to benefit from the larger data volume of ImageNet.
However, this amount of data is not always available with so little domain shift, and dataset-specific representations strike a good trade-off.
The lower panel investigates the influence of augmentation during pre-training.
For \gls*{dino}, removing color and size or multi-crop augmentations, the model loses its ability to reliably detect some issue types, in particular off-topic samples.
For \gls*{simclr}, adding multi-crop substantially improves data cleaning performance.
Interestingly, adding color and size augmentations alongside multi-crop seems to have a negative influence on near-duplicate detection, while isolating off-topic samples and label errors well.

In appendix \ref{app:FurtherAblation}, we further demonstrate that it is important to pre-train for sufficient epochs and to either normalize embeddings or use the cosine distance.
We also find that \gls*{dino} works best among four \gls*{ssl} objectives and investigate the effectiveness of different backbones.
Finally, we show that label-error detection deteriorates with label granularity, but \textsc{SelfClean} stays on par with other methods.

\section{Discussion}
\label{sec:discussion}

\textbf{Application to benchmark datasets.}
We apply the fully automatic mode of \textsc{SelfClean} to well-known image benchmark datasets and estimate the prevalence of data quality issues.
For the estimation, we used conservative guesses of a contamination rate of $\alpha=0.10$ and a significance level of $q=0.05$.
Detailed results can be found in appendix \ref{app:Detailed-BenchmarkIssues}.
For highly curated datasets with extensive manual verification, such as DDI, PAD-UFES-20, HAM10000, CheXpert, and ImageNet-1k, we find noise levels below 1\%.
However, for ISIC-2019 and PatchCamelyon, we estimate 5.4\% and 3.9\% of near duplicates that are not accounted for in the metadata.
When considering datasets with less manual curation, such as Fitzpatrick17k, CelebA, and Food-101N, we find less than 1\% of off-topic samples and label errors, and approximately $14.8\%$, $0.4\%$, and $1.4\%$ near duplicates, respectively.
The abundance of near duplicates in these benchmarks can often be traced back to crawling data of different pages using the same illustration or thumbnail images.
When data splits with near-duplicate data leaks are used, performance estimates on these datasets are optimistically biased. 

\begin{table*}
    \centering
    \small
    \caption{
    Influence of removing samples detected in the automatic cleaning mode with $\alpha = 0.10$ and $q = 0.05$ on downstream tasks.
    We report macro-averaged F1 scores for linear and $k$NN classifiers on \gls*{dino} features
    over 100 random training/evaluation splits with 80\% and 20\% fractions, respectively. 
    We compute paired performance differences before and after cleaning the evaluation set,
    and before and after cleaning also the training set.
    We report the median and the intervals to the 5\% (subscript) and 95\% (superscript) percentiles.
    Additionally, we indicate significance of a paired permutation test on the difference sign with $^{*}p<0.05$, $^{**}p<0.01$, and $^{***}p<0.001$.
    }
    \label{tab:cleaninginfluencemain}
    \begin{tabular}{l ll ll}
        \toprule
        & \multicolumn{2}{c}{\bfseries $\Delta$ $k$NN Classifier (\%pt.)} 
        & \multicolumn{2}{c}{\bfseries $\Delta$ Linear Classifier (\%pt.)} \\
        \cmidrule(lr){2-3}
        \cmidrule(lr){4-5}
        
        \bfseries Dataset
        & \multicolumn{1}{c}{Clean Eval} 
        & \multicolumn{1}{c}{Clean Train}
        & \multicolumn{1}{c}{Clean Eval} 
        & \multicolumn{1}{c}{Clean Train} \\
        \midrule
        DDI
            & $+1.2^{+1.9}_{-1.2} \ ^{***}$ 
            & $+0.0^{+1.7}_{-1.4} \ ^{***}$ 
            & $+1.0^{+11.1}_{-11.2}$ 
            & $-0.7^{+7.7}_{-10.8}$ \\
        HAM10000
            & $+0.2^{+0.5}_{-0.4} \ ^{***}$ 
            & $+0.2^{+1.3}_{-0.8} \ ^{**}$ 
            & $+0.1^{+3.2}_{-3.5}$
            & $-0.1^{+3.9}_{-3.6}$ \\
        Fitzpatrick17k
            & $-4.1^{+1.2}_{-1.3} \ ^{***}$ 
            & $+0.1^{+2.0}_{-1.7}$ 
            & $-0.6^{+2.9}_{-3.6} \ ^{**}$ 
            & $+0.2^{+3.3}_{-3.9} \ ^{*}$ \\
        ImageNet-1k
            & $-0.4^{+0.1}_{-0.2} \ ^{***}$ 
            & $+0.4^{+0.3}_{-0.4} \ ^{***}$ 
            & $-0.4^{+0.6}_{-0.6} \ ^{***}$
            & $-0.0^{+0.9}_{-0.5}$ \\
        Food-101N
            & $+0.1^{+0.1}_{-0.1} \ ^{***}$
            & $+0.1^{+0.2}_{-0.2} \ ^{***}$
            & $+0.2^{+0.6}_{-0.5} \ ^{***}$ 
            & $+0.1^{+0.6}_{-0.5} \ ^{**}$ \\
        \bottomrule
    \end{tabular}
\end{table*}

\looseness=-1
\textbf{Influence of dataset cleaning.}
In table~\ref{tab:cleaninginfluencemain} we examine the impact of cleaning data quality issues to better understand their relevance.
We train linear and $k$NN classifiers based on dataset-specific \gls*{ssl} representations for multiple classification benchmarks and measure the performance difference in F1 score when removing the problematic samples found above, first from the evaluation set and then also from the training set.
For most benchmark datasets, cleaning the evaluation set significantly alters scores.
Variations are either positive or negative depending on whether wrong samples were misclassified, and larger for datasets with significant data leaks.
Cleaning the training set has a significant positive impact for many benchmarks, indicating that issues in the training set hindered optimization.

The importance of each individual data quality issue type depends on the dataset and task, and identifying trends by domain and modality requires further investigation. 
For the limited number of cases in Table \ref{tab:cleaninginfluencemain}, and taking into account Table \ref{tab:EstimatedErrorsAppendix}, data leaks caused by near duplicates across splits seem to have the highest impact, followed by label errors. 
However, we argue that information on off-topic samples and near duplicates within the same data split is always valuable, even if it only serves the purpose of restoring trust.

\looseness=-1
\textbf{Recommended use.}
\textsc{SelfClean} determines context based on the dataset rather than a specific task, so the candidates it provides for correction may represent desired features (e.g., rare diseases or longitudinal data).
The identification of a data quality issue should not be automatically considered a suggestion to remove it.
Instead, discovering relationships among samples is always an advantage, as it can inform proper action.
While undesirable behavior may occur with the automatic mode,
this is similar to other cleaning methods applied without checks,
and such biases can be mitigated with the human-in-the-loop approach.

The tension between correcting data quality issues and the veto against the examination of evaluation data, mentioned in the introduction, has no easy resolution.
We suggest the following compromise
as an improvement to the current practice.
A benchmark dataset should be refined using an \gls*{ssl} model developed on the training set.
\textsc{SelfClean} can be used to clean both training and evaluation sets, but for the latter the human-in-the-loop mode is required, and labels should not be altered.
The number of problems found for each set separately and across them for near duplicates should be reported.
Even with human confirmation and refraining from correcting label errors, the cleaning procedure introduces some degree of bias due to the sampling of the candidate issues to be confirmed. 
We believe that in many practical cases, the benefit of data cleaning outweighs this bias.

\section{Conclusion and outlook}
We found a data-cleaning strategy called \textsc{SelfClean}, based on dataset-specific self-supervised learning and local, distance-based indicator functions, to be effective for detecting off-topic samples, near duplicates, and label errors.
We demonstrated this by comparing to state-of-the-art methods across multiple general vision and medical image benchmarks both with synthetic issues and with natural contamination.
\textsc{SelfClean} outperformed competing approaches for synthetic data quality issues, and demonstrated superior correspondence to metadata and expert verification in natural settings.
Notably, the detailed methodology surpassed the state-of-the-art in label-error detection, achieving a twofold increase in \gls*{ap} over existing approaches on known ImageNet-1k and Food-101N issues.
Moreover, applying the cleaning strategy to highly curated medical datasets and general vision benchmarks revealed multiple data quality issues with significant impact on model scores.
By correcting these data collections, confidence can be regained in reported benchmark performances.
In the future, we plan to incorporate \textsc{SelfClean} during annotation to collect higher quality datasets and during inference to enhance model robustness.

\medskip
\small
\bibliography{bibliography}

\begin{thebibliography}{78}
\providecommand{\natexlab}[1]{#1}
\providecommand{\url}[1]{\texttt{#1}}
\expandafter\ifx\csname urlstyle\endcsname\relax
  \providecommand{\doi}[1]{doi: #1}\else
  \providecommand{\doi}{doi: \begingroup \urlstyle{rm}\Url}\fi

\bibitem[Li et~al.(2021)Li, Rao, Blase, Zhang, Chu, and Zhang]{li2021cleanml}
Peng Li, Xi~Rao, Jennifer Blase, Yue Zhang, Xu~Chu, and Ce~Zhang.
\newblock {Cleanml: A study for evaluating the impact of data cleaning on ml classification tasks}.
\newblock In \emph{International Conference on Data Engineering}, volume~37, pages 13--24. IEEE, 2021.

\bibitem[Rolnick et~al.(2017)Rolnick, Veit, Belongie, and Shavit]{rolnick_deep_2018}
David Rolnick, Andreas Veit, Serge Belongie, and Nir Shavit.
\newblock Deep learning is robust to massive label noise.
\newblock \emph{arXiv preprint arXiv:1705.10694}, 2017.

\bibitem[Natarajan et~al.(2013)Natarajan, Dhillon, Ravikumar, and Tewari]{natarajan_learning_2013}
Nagarajan Natarajan, Inderjit~S Dhillon, Pradeep~K Ravikumar, and Ambuj Tewari.
\newblock {Learning with {Noisy} {Labels}}.
\newblock In \emph{Advances in Neural Information Processing Systems}, volume~26, pages 1196--1204, 2013.

\bibitem[Karimi et~al.(2020)Karimi, Dou, Warfield, and Gholipour]{karimi_deep_2020}
Davood Karimi, Haoran Dou, Simon~K. Warfield, and Ali Gholipour.
\newblock {Deep learning with noisy labels: {Exploring} techniques and remedies in medical image analysis}.
\newblock \emph{Medical Image Analysis}, 65:\penalty0 101759, 2020.
\newblock ISSN 13618415.
\newblock \doi{10.1016/j.media.2020.101759}.

\bibitem[Pezoulas et~al.(2019)Pezoulas, Kourou, Kalatzis, Exarchos, Venetsanopoulou, et~al.]{pezoulas_medical_2019}
Vasileios~C. Pezoulas, Konstantina~D. Kourou, Fanis Kalatzis, Themis~P. Exarchos, Aliki Venetsanopoulou, et~al.
\newblock {Medical data quality assessment: {On} the development of an automated framework for medical data curation}.
\newblock \emph{Computers in Biology and Medicine}, 107:\penalty0 270--283, 2019.
\newblock ISSN 0010-4825.
\newblock \doi{10.1016/j.compbiomed.2019.03.001}.

\bibitem[Daneshjou et~al.(2022{\natexlab{a}})Daneshjou, Yuksekgonul, Cai, Novoa, and Zou]{daneshjou_skincon_2022}
Roxana Daneshjou, Mert Yuksekgonul, Zhuo~Ran Cai, Roberto Novoa, and James~Y. Zou.
\newblock {{SkinCon}: {A} skin disease dataset densely annotated by domain experts for fine-grained debugging and analysis}.
\newblock In \emph{Advances in Neural Information Processing Systems}, volume~35, pages 18157--18167, 2022{\natexlab{a}}.

\bibitem[Northcutt et~al.(2021{\natexlab{a}})Northcutt, Athalye, and Mueller]{northcutt_pervasive_2021}
Curtis~G. Northcutt, Anish Athalye, and Jonas Mueller.
\newblock {Pervasive Label Errors in Test Sets Destabilize Machine Learning Benchmarks}.
\newblock In \emph{Advances in Neural Information Processing Systems}, volume~35, 2021{\natexlab{a}}.

\bibitem[Ott et~al.(2022)Ott, Barbosa-Silva, Blagec, Brauner, and Samwald]{ott_mapping_2022}
Simon Ott, Adriano Barbosa-Silva, Kathrin Blagec, Jan Brauner, and Matthias Samwald.
\newblock {Mapping global dynamics of benchmark creation and saturation in artificial intelligence}.
\newblock \emph{Nature Communications}, 13\penalty0 (1):\penalty0 6793, 2022.
\newblock ISSN 2041-1723.
\newblock \doi{10.1038/s41467-022-34591-0}.
\newblock Number: 1 Publisher: Nature Publishing Group.

\bibitem[Jarrahi et~al.(2023)Jarrahi, Memariani, and Guha]{jarrahi2022principles}
Mohammad~Hossein Jarrahi, Ali Memariani, and Shion Guha.
\newblock {The principles of data-centric AI (DCAI)}.
\newblock \emph{Communications of the ACM}, 66\penalty0 (8):\penalty0 84--92, 2023.

\bibitem[Chu et~al.(2016)Chu, Ilyas, Krishnan, and Wang]{chu_data_2016}
Xu~Chu, Ihab~F. Ilyas, Sanjay Krishnan, and Jiannan Wang.
\newblock {Data {Cleaning}: {Overview} and {Emerging} {Challenges}}.
\newblock In \emph{{International} {Conference} on {Management} of {Data}}, pages 2201--2206, San Francisco California USA, 2016. ACM.
\newblock ISBN 978-1-4503-3531-7.
\newblock \doi{10.1145/2882903.2912574}.

\bibitem[Vailoppilly et~al.(2021)Vailoppilly, Sakthivel, and Kumar]{vailoppilly_all_2021}
Arun~Prasad Vailoppilly, Ramkumar Sakthivel, and Resham~Sundar Kumar.
\newblock {All-in-one Data Cleansing Tool}.
\newblock In \emph{NeurIPS Data-Centric AI Workshop}, 2021.

\bibitem[Cleanlab(2018)]{cleanlab2017cleanlab}
Inc. Cleanlab.
\newblock {CleanLab}.
\newblock GitHub, 2018.
\newblock URL \url{https://github.com/cleanlab/cleanlab}.

\bibitem[Cleanlab(2022)]{cleanlab2022cleanvision}
Inc. Cleanlab.
\newblock {CleanVision}.
\newblock GitHub, 2022.
\newblock URL \url{https://github.com/cleanlab/cleanvision}.

\bibitem[AG(2020)]{susmelj2020lightly}
Lightly AG.
\newblock {Lightly}.
\newblock GitHub, 2020.
\newblock URL \url{https://github.com/lightly-ai/lightly}.

\bibitem[Layer(2022)]{visuallayer2022fastdup}
Visual Layer.
\newblock {FastDup}.
\newblock GitHub, 2022.
\newblock URL \url{https://github.com/visual-layer/fastdup}.

\bibitem[Maćkiewicz and Ratajczak(1993)]{mackiewicz_principal_1993}
Andrzej Maćkiewicz and Waldemar Ratajczak.
\newblock {Principal components analysis ({PCA})}.
\newblock \emph{Computers \& Geosciences}, 19\penalty0 (3):\penalty0 303--342, 1993.
\newblock ISSN 0098-3004.
\newblock \doi{10.1016/0098-3004(93)90090-R}.

\bibitem[Maaten and Hinton(2008)]{maaten_visualizing_2008}
Laurens van~der Maaten and Geoffrey Hinton.
\newblock {Visualizing {Data} using t-{SNE}}.
\newblock \emph{Journal of Machine Learning Research}, 9\penalty0 (86):\penalty0 2579--2605, 2008.
\newblock ISSN 1533-7928.

\bibitem[Deng et~al.(2009)Deng, Dong, Socher, Li, Li, and Fei-Fei]{deng_imagenet_2009}
Jia Deng, Wei Dong, Richard Socher, Li-Jia Li, Kai Li, and Li~Fei-Fei.
\newblock {ImageNet: A large-scale hierarchical image database}.
\newblock In \emph{IEEE/CVF {Conference} on {Computer} {Vision} and {Pattern} {Recognition}}, pages 248--255, 2009.
\newblock \doi{10.1109/CVPR.2009.5206848}.

\bibitem[Ozbulak et~al.(2023)Ozbulak, Lee, Boga, Anzaku, Park, Van~Messem, De~Neve, and Vankerschaver]{ozbulak_know_2023}
Utku Ozbulak, Hyun~Jung Lee, Beril Boga, Esla~Timothy Anzaku, Homin Park, Arnout Van~Messem, Wesley De~Neve, and Joris Vankerschaver.
\newblock {Know Your Self-supervised Learning: A Survey on Image-based Generative and Discriminative Training}.
\newblock \emph{Transactions on Machine Learning Research}, 2023.
\newblock ISSN 2835-8856.

\bibitem[Fernandez et~al.(2022)Fernandez, Sablayrolles, Furon, J{\'e}gou, and Douze]{fernandez_watermarking_2022}
Pierre Fernandez, Alexandre Sablayrolles, Teddy Furon, Herv{\'e} J{\'e}gou, and Matthijs Douze.
\newblock {Watermarking images in self-supervised latent spaces}.
\newblock In \emph{IEEE International Conference on Acoustics, Speech and Signal Processing}, pages 3054--3058. IEEE, 2022.

\bibitem[Wang and Isola(2020)]{wang_understanding_2020}
Tongzhou Wang and Phillip Isola.
\newblock {Understanding {Contrastive} {Representation} {Learning} through {Alignment} and {Uniformity} on the {Hypersphere}}.
\newblock In \emph{{International} {Conference} on {Machine} {Learning}}, volume~37, pages 9929--9939. PMLR, 2020.
\newblock ISSN: 2640-3498.

\bibitem[Sorscher et~al.(2022)Sorscher, Geirhos, Shekhar, Ganguli, and Morcos]{sorscher_beyond_2022}
Ben Sorscher, Robert Geirhos, Shashank Shekhar, Surya Ganguli, and Ari Morcos.
\newblock {Beyond neural scaling laws: beating power law scaling via data pruning}.
\newblock In \emph{Advances in Neural Information Processing Systems}, volume~35, pages 19523--19536, 2022.

\bibitem[Cao and Wu(2021)]{cao_rethinking_2021}
Yun-Hao Cao and Jianxin Wu.
\newblock {Rethinking self-supervised learning: Small is beautiful}.
\newblock \emph{arXiv preprint arXiv:2103.13559}, 2021.

\bibitem[Yang et~al.(2024)Yang, Zhou, Li, and Liu]{yang_generalized_2022}
Jingkang Yang, Kaiyang Zhou, Yixuan Li, and Ziwei Liu.
\newblock {Generalized out-of-distribution detection: A survey}.
\newblock \emph{International Journal of Computer Vision}, pages 1--28, 2024.

\bibitem[Aggarwal(2017)]{aggarwal_outlier_2017}
Charu~C. Aggarwal.
\newblock \emph{An introduction to outlier analysis}.
\newblock Springer, 2017.
\newblock ISBN 978-3-319-47578-3.
\newblock \doi{10.1007/978-3-319-47578-3_1}.

\bibitem[Boukerche et~al.(2020)Boukerche, Zheng, and Alfandi]{boukerche_outlier_2020}
Azzedine Boukerche, Lining Zheng, and Omar Alfandi.
\newblock {Outlier {Detection}: {Methods}, {Models}, and {Classification}}.
\newblock \emph{ACM Computing Surveys}, 53\penalty0 (3):\penalty0 55:1--55:37, 2020.
\newblock ISSN 0360-0300.
\newblock \doi{10.1145/3381028}.

\bibitem[Hendrycks et~al.(2019)Hendrycks, Mazeika, and Dietterich]{hendrycks_deep_2019}
Dan Hendrycks, Mantas Mazeika, and Thomas Dietterich.
\newblock {Deep Anomaly Detection with Outlier Exposure}.
\newblock In \emph{International Conference on Learning Representations}, 2019.

\bibitem[Abati et~al.(2019)Abati, Porrello, Calderara, and Cucchiara]{abati_latent_2019}
Davide Abati, Angelo Porrello, Simone Calderara, and Rita Cucchiara.
\newblock {Latent {Space} {Autoregression} for {Novelty} {Detection}}.
\newblock In \emph{IEEE/CVF {Conference} on {Computer} {Vision} and {Pattern} {Recognition}}, pages 481--490, Long Beach, CA, USA, 2019.
\newblock ISBN 978-1-72813-293-8.
\newblock \doi{10.1109/CVPR.2019.00057}.

\bibitem[Ruff et~al.(2018)Ruff, Vandermeulen, Goernitz, Deecke, Siddiqui, et~al.]{ruff_deep_2018}
Lukas Ruff, Robert Vandermeulen, Nico Goernitz, Lucas Deecke, Shoaib~Ahmed Siddiqui, et~al.
\newblock {Deep {One}-{Class} {Classification}}.
\newblock In \emph{{International} {Conference} on {Machine} {Learning}}, volume~35, pages 4393--4402. PMLR, 2018.
\newblock ISSN: 2640-3498.

\bibitem[Ren et~al.(2019)Ren, Liu, Fertig, Snoek, Poplin, et~al.]{ren_likelihood_2019}
Jie Ren, Peter~J. Liu, Emily Fertig, Jasper Snoek, Ryan Poplin, et~al.
\newblock {Likelihood {Ratios} for {Out}-of-{Distribution} {Detection}}.
\newblock In \emph{Advances in Neural Information Processing Systems}, volume~32, pages 14707--14718, 2019.

\bibitem[Lowe(2004)]{lowe_distinctive_2004}
David~G. Lowe.
\newblock {Distinctive {Image} {Features} from {Scale}-{Invariant} {Keypoints}}.
\newblock \emph{International Journal of Computer Vision}, 60\penalty0 (2):\penalty0 91--110, 2004.
\newblock ISSN 0920-5691.
\newblock \doi{10.1023/B:VISI.0000029664.99615.94}.

\bibitem[Ke et~al.(2004)Ke, Sukthankar, Huston, Ke, and Sukthankar]{ke_efcient_2004}
Yan Ke, Rahul Sukthankar, Larry Huston, Yan Ke, and Rahul Sukthankar.
\newblock {Efficient near-duplicate detection and sub-image retrieval}.
\newblock In \emph{ACM multimedia}, volume~4, page~5. Citeseer, 2004.

\bibitem[Babenko et~al.(2014)Babenko, Slesarev, Chigorin, and Lempitsky]{babenko_neural_2014}
Artem Babenko, Anton Slesarev, Alexandr Chigorin, and Victor Lempitsky.
\newblock {Neural codes for image retrieval}.
\newblock In \emph{European Conference on Computer Vision}, volume~13, pages 584--599. Springer, 2014.

\bibitem[Žbontar and LeCun(2015)]{zbontar_computing_2015}
Jure Žbontar and Yann LeCun.
\newblock {Computing the {Stereo} {Matching} {Cost} with a {Convolutional} {Neural} {Network}}.
\newblock In \emph{IEEE/CVF {Conference} on {Computer} {Vision} and {Pattern} {Recognition}}, pages 1592--1599, 2015.
\newblock \doi{10.1109/CVPR.2015.7298767}.

\bibitem[Pizzi et~al.(2022)Pizzi, Roy, Ravindra, Goyal, and Douze]{pizzi_self-supervised_2022}
Ed~Pizzi, Sreya~Dutta Roy, Sugosh~Nagavara Ravindra, Priya Goyal, and Matthijs Douze.
\newblock {A {Self}-{Supervised} {Descriptor} for {Image} {Copy} {Detection}}.
\newblock In \emph{IEEE/CVF {Conference} on {Computer} {Vision} and {Pattern} {Recognition}}, pages 14512--14522, New Orleans, LA, USA, 2022.
\newblock ISBN 978-1-66546-946-3.
\newblock \doi{10.1109/CVPR52688.2022.01413}.

\bibitem[Oquab et~al.(2024)Oquab, Darcet, Moutakanni, Vo, Szafraniec, Khalidov, Fernandez, HAZIZA, Massa, El-Nouby, Assran, Ballas, Galuba, Howes, Huang, Li, Misra, Rabbat, Sharma, Synnaeve, Xu, Jegou, Mairal, Labatut, Joulin, and Bojanowski]{oquab_dinov2_2023}
Maxime Oquab, Timoth{\'e}e Darcet, Th{\'e}o Moutakanni, Huy~V. Vo, Marc Szafraniec, Vasil Khalidov, Pierre Fernandez, Daniel HAZIZA, Francisco Massa, Alaaeldin El-Nouby, Mido Assran, Nicolas Ballas, Wojciech Galuba, Russell Howes, Po-Yao Huang, Shang-Wen Li, Ishan Misra, Michael Rabbat, Vasu Sharma, Gabriel Synnaeve, Hu~Xu, Herve Jegou, Julien Mairal, Patrick Labatut, Armand Joulin, and Piotr Bojanowski.
\newblock {{DINO}v2: Learning Robust Visual Features without Supervision}.
\newblock \emph{Transactions on Machine Learning Research}, 2024.
\newblock ISSN 2835-8856.

\bibitem[Chen et~al.(2019)Chen, Liao, Chen, and Zhang]{chen_understanding_2019}
Pengfei Chen, Ben~Ben Liao, Guangyong Chen, and Shengyu Zhang.
\newblock {Understanding and utilizing deep neural networks trained with noisy labels}.
\newblock In \emph{International conference on machine learning}, pages 1062--1070. PMLR, 2019.

\bibitem[Zhang et~al.(2020)Zhang, Jin, Liu, Zhu, Mu, et~al.]{zhang_imagedc_2020}
Yun Zhang, Zongze Jin, Fan Liu, Weilin Zhu, Weimin Mu, et~al.
\newblock {{ImageDC}: {Image} {Data} {Cleaning} {Framework} {Based} on {Deep} {Learning}}.
\newblock In \emph{{International} {Conference} on {Artificial} {Intelligence} and {Information} {Systems}}, pages 748--752. IEEE, 2020.
\newblock \doi{10.1109/ICAIIS49377.2020.9194803}.

\bibitem[Huang et~al.(2022)Huang, Lin, and Xu]{huang_contrastive_2022}
Bin Huang, Yaohai Lin, and Chaoyang Xu.
\newblock {Contrastive label correction for noisy label learning}.
\newblock \emph{Information Sciences}, 611:\penalty0 173--184, 2022.
\newblock ISSN 0020-0255.
\newblock \doi{10.1016/j.ins.2022.08.060}.

\bibitem[Lee et~al.(2021)Lee, Kwon, Lee, Kim, Lee, and Kim]{lee_augment_2021}
Youngjune Lee, Oh~Joon Kwon, Haeju Lee, Joonyoung Kim, Kangwook Lee, and Kee-Eung Kim.
\newblock {Augment \& Valuate: A Data Enhancement Pipeline for Data-Centric AI}.
\newblock In \emph{NeurIPS Data-Centric AI Workshop}, 2021.

\bibitem[Northcutt et~al.(2021{\natexlab{b}})Northcutt, Jiang, and Chuang]{northcutt_confident_2022}
Curtis Northcutt, Lu~Jiang, and Isaac Chuang.
\newblock {Confident learning: Estimating uncertainty in dataset labels}.
\newblock \emph{Journal of Artificial Intelligence Research}, 70:\penalty0 1373--1411, 2021{\natexlab{b}}.

\bibitem[Chen et~al.(2020)Chen, Kornblith, Norouzi, and Hinton]{chen_simple_2020}
Ting Chen, Simon Kornblith, Mohammad Norouzi, and Geoffrey Hinton.
\newblock {A {Simple} {Framework} for {Contrastive} {Learning} of {Visual} {Representations}}.
\newblock In \emph{{International} {Conference} on {Machine} {Learning}}, volume~37. PMLR, 2020.

\bibitem[Caron et~al.(2021)Caron, Touvron, Misra, J{\'e}gou, Mairal, Bojanowski, and Joulin]{caron_emerging_2021}
Mathilde Caron, Hugo Touvron, Ishan Misra, Herv{\'e} J{\'e}gou, Julien Mairal, Piotr Bojanowski, and Armand Joulin.
\newblock {Emerging properties in self-supervised vision transformers}.
\newblock In \emph{International Conference on Computer Vision}, pages 9650--9660. IEEE, 2021.

\bibitem[Oakden-Rayner et~al.(2020)Oakden-Rayner, Dunnmon, Carneiro, and R{\'e}]{oakden2020hidden}
Luke Oakden-Rayner, Jared Dunnmon, Gustavo Carneiro, and Christopher R{\'e}.
\newblock {Hidden stratification causes clinically meaningful failures in machine learning for medical imaging}.
\newblock In \emph{ACM Conference on Health, Inference, and Learning}, pages 151--159, 2020.

\bibitem[Gower and Ross(1969)]{gower_minimum_1969}
J.~C. Gower and G.~J.~S. Ross.
\newblock {Minimum {Spanning} {Trees} and {Single} {Linkage} {Cluster} {Analysis}}.
\newblock \emph{Journal of the Royal Statistical Society. Series C (Applied Statistics)}, 18\penalty0 (1):\penalty0 54--64, 1969.
\newblock ISSN 0035-9254.
\newblock \doi{10.2307/2346439}.
\newblock Publisher: [Wiley, Royal Statistical Society].

\bibitem[Jiang et~al.(2001)Jiang, Tseng, and Su]{jiang_two-phase_2001}
M.F Jiang, S.S Tseng, and C.M Su.
\newblock {Two-phase clustering process for outliers detection}.
\newblock \emph{Pattern Recognition Letters}, 22\penalty0 (6-7):\penalty0 691--700, 2001.
\newblock ISSN 01678655.
\newblock \doi{10.1016/S0167-8655(00)00131-8}.

\bibitem[Ho and Basu(2002)]{ho_complexity_2002}
Tin~Kam Ho and M.~Basu.
\newblock {Complexity measures of supervised classification problems}.
\newblock \emph{IEEE Transactions on Pattern Analysis and Machine Intelligence}, 24\penalty0 (3):\penalty0 289--300, 2002.
\newblock ISSN 1939-3539.
\newblock \doi{10.1109/34.990132}.
\newblock Conference Name: IEEE Transactions on Pattern Analysis and Machine Intelligence.

\bibitem[Coates et~al.(2011)Coates, Ng, and Lee]{coates_analysis_2011}
Adam Coates, Andrew Ng, and Honglak Lee.
\newblock {An {Analysis} of {Single}-{Layer} {Networks} in {Unsupervised} {Feature} {Learning}}.
\newblock In \emph{{International} {Conference} on {Artificial} {Intelligence} and {Statistics}}, volume~14, pages 215--223. JMLR Workshop and Conference Proceedings, 2011.
\newblock ISSN: 1938-7228.

\bibitem[Liu et~al.(2015)Liu, Luo, Wang, and Tang]{liu2015faceattributes}
Ziwei Liu, Ping Luo, Xiaogang Wang, and Xiaoou Tang.
\newblock {Deep Learning Face Attributes in the Wild}.
\newblock In \emph{International Conference on Computer Vision}. IEEE, 2015.

\bibitem[Bossard et~al.(2014)Bossard, Guillaumin, and Van~Gool]{bossard14}
Lukas Bossard, Matthieu Guillaumin, and Luc Van~Gool.
\newblock {Food-101 -- Mining Discriminative Components with Random Forests}.
\newblock In \emph{European Conference on Computer Vision}, volume~13, 2014.

\bibitem[Irvin et~al.(2019)Irvin, Rajpurkar, Ko, Yu, Ciurea-Ilcus, Chute, Marklund, Haghgoo, Ball, Shpanskaya, et~al.]{irvin_chexpert_2019}
Jeremy Irvin, Pranav Rajpurkar, Michael Ko, Yifan Yu, Silviana Ciurea-Ilcus, Chris Chute, Henrik Marklund, Behzad Haghgoo, Robyn Ball, Katie Shpanskaya, et~al.
\newblock {Chexpert: A large chest radiograph dataset with uncertainty labels and expert comparison}.
\newblock In \emph{AAAI Conference on Artificial Intelligence}, volume~33, pages 590--597, 2019.

\bibitem[Pham et~al.(2021)Pham, Do, and Nguyen]{pham_dicom_2021}
Hieu~H Pham, Dung~V Do, and Ha~Q Nguyen.
\newblock {Dicom imaging router: An open deep learning framework for classification of body parts from dicom x-ray scans}.
\newblock \emph{arXiv preprint arXiv:2108.06490}, 2021.

\bibitem[Ehteshami~Bejnordi et~al.(2017)Ehteshami~Bejnordi, Veta, Johannes~van Diest, van Ginneken, Karssemeijer, et~al.]{ehteshami_bejnordi_diagnostic_2017}
Babak Ehteshami~Bejnordi, Mitko Veta, Paul Johannes~van Diest, Bram van Ginneken, Nico Karssemeijer, et~al.
\newblock {Diagnostic {Assessment} of {Deep} {Learning} {Algorithms} for {Detection} of {Lymph} {Node} {Metastases} in {Women} {With} {Breast} {Cancer}}.
\newblock \emph{JAMA}, 318\penalty0 (22):\penalty0 2199--2210, 2017.
\newblock ISSN 0098-7484.
\newblock \doi{10.1001/jama.2017.14585}.

\bibitem[Tschandl et~al.(2018)Tschandl, Rosendahl, and Kittler]{tschandl_ham10000_2018}
Philipp Tschandl, Cliff Rosendahl, and Harald Kittler.
\newblock {The {HAM10000} dataset, a large collection of multi-source dermatoscopic images of common pigmented skin lesions}.
\newblock \emph{Scientific Data}, 5\penalty0 (1):\penalty0 180161, 2018.
\newblock ISSN 2052-4463.
\newblock \doi{10.1038/sdata.2018.161}.
\newblock Number: 1 Publisher: Nature Publishing Group.

\bibitem[Groh et~al.(2021)Groh, Harris, Soenksen, Lau, Han, Kim, Koochek, and Badri]{groh_evaluating_2021}
Matthew Groh, Caleb Harris, Luis Soenksen, Felix Lau, Rachel Han, Aerin Kim, Arash Koochek, and Omar Badri.
\newblock {Evaluating deep neural networks trained on clinical images in dermatology with the fitzpatrick 17k dataset}.
\newblock In \emph{IEEE/CVF {Conference} on {Computer} {Vision} and {Pattern} {Recognition}}, pages 1820--1828, 2021.

\bibitem[Daneshjou et~al.(2022{\natexlab{b}})Daneshjou, Vodrahalli, Novoa, Jenkins, Liang, et~al.]{daneshjou_disparities_2022}
Roxana Daneshjou, Kailas Vodrahalli, Roberto~A. Novoa, Melissa Jenkins, Weixin Liang, et~al.
\newblock {Disparities in dermatology {AI} performance on a diverse, curated clinical image set}.
\newblock \emph{Science Advances}, 8\penalty0 (32):\penalty0 eabq6147, 2022{\natexlab{b}}.
\newblock \doi{10.1126/sciadv.abq6147}.
\newblock Publisher: American Association for the Advancement of Science.

\bibitem[Pacheco et~al.(2020)Pacheco, Lima, Salomão, Krohling, Biral, et~al.]{pacheco_pad-ufes-20_2020}
Andre G.~C. Pacheco, Gustavo~R. Lima, Amanda~S. Salomão, Breno Krohling, Igor~P. Biral, et~al.
\newblock {{PAD}-{UFES}-20: {A} skin lesion dataset composed of patient data and clinical images collected from smartphones}.
\newblock \emph{Data in Brief}, 32:\penalty0 106221, 2020.
\newblock ISSN 2352-3409.
\newblock \doi{10.1016/j.dib.2020.106221}.

\bibitem[Rendle(2019)]{rendle_evaluation_2019}
Steffen Rendle.
\newblock {Evaluation metrics for item recommendation under sampling}.
\newblock \emph{arXiv preprint arXiv:1912.02263}, 2019.

\bibitem[Han et~al.(2022)Han, Hu, Huang, Jiang, and Zhao]{han_adbench_2022}
Songqiao Han, Xiyang Hu, Hailiang Huang, Minqi Jiang, and Yue Zhao.
\newblock {ADBench: Anomaly detection benchmark}.
\newblock In \emph{Advances in Neural Information Processing Systems}, volume~35, pages 32142--32159, 2022.

\bibitem[Goldstein and Dengel(2012)]{goldstein_histogram-based_2012}
Markus Goldstein and Andreas Dengel.
\newblock {Histogram-based outlier score (hbos): A fast unsupervised anomaly detection algorithm}.
\newblock \emph{KI-2012: poster and demo track}, 1:\penalty0 59--63, 2012.

\bibitem[Li et~al.(2022)Li, Zhao, Hu, Botta, Ionescu, et~al.]{li_ecod_2022}
Zheng Li, Yue Zhao, Xiyang Hu, Nicola Botta, Cezar Ionescu, et~al.
\newblock {{ECOD}: {Unsupervised} {Outlier} {Detection} {Using} {Empirical} {Cumulative} {Distribution} {Functions}}.
\newblock \emph{IEEE Transactions on Knowledge and Data Engineering}, pages 1--1, 2022.
\newblock ISSN 1041-4347, 1558-2191, 2326-3865.
\newblock \doi{10.1109/TKDE.2022.3159580}.
\newblock arXiv:2201.00382 [cs, stat].

\bibitem[Marr et~al.(1997)Marr, Hildreth, and Brenner]{marr_theory_1997}
D.~Marr, E.~Hildreth, and Sydney Brenner.
\newblock {Theory of edge detection}.
\newblock \emph{Proceedings of the Royal Society of London. Series B. Biological Sciences}, 207\penalty0 (1167):\penalty0 187--217, 1997.
\newblock \doi{10.1098/rspb.1980.0020}.
\newblock Publisher: Royal Society.

\bibitem[Wang et~al.(2004)Wang, Bovik, Sheikh, and Simoncelli]{wang_image_2004}
Zhou Wang, A.C. Bovik, H.R. Sheikh, and E.P. Simoncelli.
\newblock {Image quality assessment: from error visibility to structural similarity}.
\newblock \emph{IEEE Transactions on Image Processing}, 13\penalty0 (4):\penalty0 600--612, 2004.
\newblock ISSN 1941-0042.
\newblock \doi{10.1109/TIP.2003.819861}.
\newblock Conference Name: IEEE Transactions on Image Processing.

\bibitem[Lee et~al.(2018)Lee, He, Zhang, and Yang]{lee_cleannet_2018}
Kuang-Huei Lee, Xiaodong He, Lei Zhang, and Linjun Yang.
\newblock {Cleannet: Transfer learning for scalable image classifier training with label noise}.
\newblock In \emph{IEEE/CVF {Conference} on {Computer} {Vision} and {Pattern} {Recognition}}, pages 5447--5456, 2018.

\bibitem[Dosovitskiy(2020)]{dosovitskiy_image_2021}
Alexey Dosovitskiy.
\newblock {An image is worth 16x16 words: Transformers for image recognition at scale}.
\newblock \emph{arXiv preprint arXiv:2010.11929}, 2020.

\bibitem[Paszke et~al.(2019)Paszke, Gross, Massa, Lerer, Bradbury, Chanan, Killeen, Lin, Gimelshein, Antiga, et~al.]{paszke_pytorch_2019}
Adam Paszke, Sam Gross, Francisco Massa, Adam Lerer, James Bradbury, Gregory Chanan, Trevor Killeen, Zeming Lin, Natalia Gimelshein, Luca Antiga, et~al.
\newblock {Pytorch: An imperative style, high-performance deep learning library}.
\newblock In \emph{Advances in Neural Information Processing Systems}, volume~32, pages 8026--8037, 2019.

\bibitem[Peng et~al.(2021)Peng, Mathur, and Narayanan]{peng2021mitigating}
Kenny Peng, Arunesh Mathur, and Arvind Narayanan.
\newblock {Mitigating dataset harms requires stewardship: Lessons from 1000 papers}.
\newblock In \emph{Advances in Neural Information Processing Systems}, volume~34, 2021.

\bibitem[Veeling et~al.(2018)Veeling, Linmans, Winkens, Cohen, and Welling]{Veeling2018-qh}
Bastiaan~S Veeling, Jasper Linmans, Jim Winkens, Taco Cohen, and Max Welling.
\newblock {Rotation equivariant CNNs for digital pathology}.
\newblock In \emph{Medical Image Computing and Computer Assisted Intervention--MICCAI 2018: 21st International Conference, Granada, Spain, September 16-20, 2018, Proceedings, Part II 11}, pages 210--218. Springer, 2018.

\bibitem[Combalia et~al.(2019)Combalia, Codella, Rotemberg, Helba, Vilaplana, Reiter, Carrera, Barreiro, Halpern, Puig, et~al.]{combalia_bcn20000_2019}
Marc Combalia, Noel~CF Codella, Veronica Rotemberg, Brian Helba, Veronica Vilaplana, Ofer Reiter, Cristina Carrera, Alicia Barreiro, Allan~C Halpern, Susana Puig, et~al.
\newblock {Bcn20000: Dermoscopic lesions in the wild}.
\newblock \emph{arXiv preprint arXiv:1908.02288}, 2019.

\bibitem[Liu et~al.(2008)Liu, Ting, and Zhou]{liu_isolation_2008}
Fei~Tony Liu, Kai~Ming Ting, and Zhi-Hua Zhou.
\newblock {Isolation {Forest}}.
\newblock In \emph{2008 {Eighth} {IEEE} {International} {Conference} on {Data} {Mining}}, pages 413--422, 2008.
\newblock \doi{10.1109/ICDM.2008.17}.
\newblock ISSN: 2374-8486.

\bibitem[Freund et~al.(1999)Freund, Schapire, and Abe]{freund_short_1999}
Yoav Freund, Robert Schapire, and Naoki Abe.
\newblock {A short introduction to boosting}.
\newblock \emph{Journal-Japanese Society For Artificial Intelligence}, 14\penalty0 (771-780):\penalty0 1612, 1999.

\bibitem[Sharma et~al.(2020)Sharma, Donmez, Luo, Liu, and Yalniz]{sharma_noiserank_2020}
Karishma Sharma, Pinar Donmez, Enming Luo, Yan Liu, and I~Zeki Yalniz.
\newblock {Noiserank: Unsupervised label noise reduction with dependence models}.
\newblock In \emph{European Conference on Computer Vision}, volume~16, pages 737--753. Springer, 2020.

\bibitem[Dubois et~al.(2022)Dubois, Ermon, Hashimoto, and Liang]{dubois_improving_2022}
Yann Dubois, Stefano Ermon, Tatsunori~B. Hashimoto, and Percy~S. Liang.
\newblock {Improving {Self}-{Supervised} {Learning} by {Characterizing} {Idealized} {Representations}}.
\newblock In \emph{Advances in Neural Information Processing Systems}, volume~35, pages 11279--11296, 2022.

\bibitem[E and Wojtowytsch(2022)]{e_emergence_2022}
Weinan E and Stephan Wojtowytsch.
\newblock {On the emergence of simplex symmetry in the final and penultimate layers of neural network classifiers}.
\newblock In \emph{{Mathematical} and {Scientific} {Machine} {Learning} {Conference}}, volume~2, pages 270--290. PMLR, 2022.
\newblock ISSN: 2640-3498.

\bibitem[Grill et~al.(2020)Grill, Strub, Altché, Tallec, Richemond, Buchatskaya, Doersch, Avila~Pires, Guo, Gheshlaghi~Azar, Piot, kavukcuoglu, Munos, and Valko]{grill_bootstrap_2020}
Jean-Bastien Grill, Florian Strub, Florent Altché, Corentin Tallec, Pierre Richemond, Elena Buchatskaya, Carl Doersch, Bernardo Avila~Pires, Zhaohan Guo, Mohammad Gheshlaghi~Azar, Bilal Piot, koray kavukcuoglu, Remi Munos, and Michal Valko.
\newblock {Bootstrap {Your} {Own} {Latent} - {A} {New} {Approach} to {Self}-{Supervised} {Learning}}.
\newblock In \emph{Advances in Neural Information Processing Systems}, volume~33, pages 21271--21284, 2020.

\bibitem[He et~al.(2022)He, Chen, Xie, Li, Doll{\'a}r, and Girshick]{he_masked_2021}
Kaiming He, Xinlei Chen, Saining Xie, Yanghao Li, Piotr Doll{\'a}r, and Ross Girshick.
\newblock {Masked autoencoders are scalable vision learners}.
\newblock In \emph{IEEE/CVF {Conference} on {Computer} {Vision} and {Pattern} {Recognition}}, pages 16000--16009, 2022.

\bibitem[He et~al.(2016)He, Zhang, Ren, and Sun]{he_deep_2016}
Kaiming He, Xiangyu Zhang, Shaoqing Ren, and Jian Sun.
\newblock {Deep residual learning for image recognition}.
\newblock In \emph{IEEE/CVF {Conference} on {Computer} {Vision} and {Pattern} {Recognition}}, pages 770--778, 2016.

\bibitem[Tokuda et~al.(2022)Tokuda, Comin, and Costa]{tokuda_revisiting_2020}
Eric~K Tokuda, Cesar~H Comin, and Luciano da~F Costa.
\newblock {Revisiting agglomerative clustering}.
\newblock \emph{Physica A: Statistical mechanics and its applications}, 585:\penalty0 126433, 2022.

\end{thebibliography}

\section*{Checklist}


\begin{enumerate}

\item For all authors...
\begin{enumerate}
  \item Do the main claims made in the abstract and introduction accurately reflect the paper's contributions and scope?
    \answerYes{All claims reflect the contributions.}
  \item Did you describe the limitations of your work?
    \answerYes{See section~\ref{sec:Limitations}.}
  \item Did you discuss any potential negative societal impacts of your work?
    \answerYes{See section~\ref{sec:BroaderImpact} and~\ref{sec:Limitations}.}
  \item Have you read the ethics review guidelines and ensured that your paper conforms to them?
    \answerYes{The paper conforms to all ethical guidelines.}
\end{enumerate}

\item If you are including theoretical results...
\begin{enumerate}
    \item Did you state the full set of assumptions of all theoretical results?
        \answerNA{We have no theoretical results.}
    \item Did you include complete proofs of all theoretical results?
        \answerNA{We have no theoretical results.}
\end{enumerate}

\item If you ran experiments (e.g. for benchmarks)...
\begin{enumerate}
    \item Did you include the code, data, and instructions needed to reproduce the main experimental results (either in the supplemental material or as a URL)?
        \answerYes{A library link is provided in section~\ref{app:Training-Details} and specific paper reproduction code is attached as supplementary material.}
    \item Did you specify all the training details (e.g., data splits, hyperparameters, how they were chosen)?
        \answerYes{See section~\ref{sec:experimentalsetup} for experimental setup, section~\ref{app:Training-Details} for detailed hyperparameters, and the provided code.}
    \item Did you report error bars (e.g., with respect to the random seed after running experiments multiple times)?
        \answerYes{Where computationally possible, we repeated experiments for different random seeds as in section~\ref{ssec:syntheticresults}, or estimated finite-sample uncertainties as in section~\ref{app:Detailed-Meta}.}
    \item Did you include the total amount of compute and the type of resources used (e.g., type of GPUs, internal cluster, or cloud provider)?
        \answerYes{See section~\ref{app:Training-Details} on training details.}
\end{enumerate}

\item If you are using existing assets (e.g., code, data, models) or curating/releasing new assets...
\begin{enumerate}
  \item If your work uses existing assets, did you cite the creators?
    \answerYes{All assets were correctly cited.}
  \item Did you mention the license of the assets?
    \answerYes{The description of datasets in section \ref{app:Datasets} contains details on licensing.}
  \item Did you include any new assets either in the supplemental material or as a URL?
    \answerYes{Assets for reproducing results are included as a link to an anonymized repository and within the supplementary material.}
  \item Did you discuss whether and how consent was obtained from people whose data you're using/curating?
    \answerYes{See section~\ref{app:Human-Verification}.}
  \item Did you discuss whether the data you are using/curating contains personally identifiable information or offensive content?
    \answerYes{See section~\ref{app:Human-Verification}, we only collect binary confirmation labels without any sensitive information.}
\end{enumerate}

\item If you used crowdsourcing or conducted research with human subjects...
\begin{enumerate}
  \item Did you include the full text of instructions given to participants and screenshots, if applicable?
    \answerYes{See section~\ref{app:Human-Verification}.}
  \item Did you describe any potential participant risks, with links to Institutional Review Board (IRB) approvals, if applicable?
    \answerNA{See section~\ref{app:Human-Verification}.}
  \item Did you include the estimated hourly wage paid to participants and the total amount spent on participant compensation?
    \answerYes{See section~\ref{app:Human-Verification}.}
\end{enumerate}

\end{enumerate}


\newpage
\appendix
\addcontentsline{toc}{section}{Appendix} 
\part{Appendix} 
\parttoc 

\section{Broader impact} \label{sec:BroaderImpact}

\looseness=-1
\textsc{SelfClean} is a new data-cleaning procedure that can be applied to any visual data collection.
The procedure relies on \gls*{ssl} and, therefore, does not inherit annotation bias.
Practitioners can choose if they want the cleaning process to happen fully automatically or with human intervention.
Gaining insights into data collections of unknown quality can lead to the curation of more reliable benchmarks, which in turn result in performance measurements that are more accurate generalization estimates.
Moreover, reported benchmark results can be questioned if they contain substantial contamination.
\textsc{SelfClean} thus significantly contributes to clarifying which methods are the most valuable and to steering future research directions both in academia and applied innovation.

The near-duplicate detection component of \textsc{SelfClean} could be used for person re-identification and data de-anonymization, even if it was not designed for this purpose.
Although new in peer-reviewed publications for data cleaning to the best of our knowledge, this method can already be found in at least one publicly available tool~\citep{susmelj2020lightly}.
We believe that the benefits of increased awareness outweigh the increased chances of malignant use.

\section{Limitations} \label{sec:Limitations}

\textsc{SelfClean} hinges on the considered dataset and inherits biases from its intrinsic composition.
For example, given an image collection with a minority group that can be easily distinguished from the rest, the minority samples may be suggested to be off-topic.
This risk is studied in section \ref{app:minority}, where we find no evidence for this behavior for multiple datasets such as DDI, Fitzpatrick17k, and CheXpert.

From a computational perspective, the current formulation of near-duplicate detection does not scale well with dataset size.
This could be improved with approximation methods or by relying on an iterative analysis of nearest-neighbor distances.
Also, the detailed methodology requires \gls*{ssl} pre-training on the dataset in order to clean it, which requires sufficient computational power and might be a limiting factor to some.
However, training on the considered dataset is required by other methods such as confident learning \citep{northcutt_confident_2022}, although this training is supervised and requires labeled data.
In contrast, \textsc{SelfClean} does not require any annotations during training.

Currently, there is no standard protocol for evaluating data cleaning frameworks. 
To address this, we designed synthetic experiments that simulate data quality issues.
The datasets used for evaluation are, however, already contaminated (see section \ref{app:Benchmark-Inspect}), which means that performance measures are capped. 
However, since all approaches are subject to the same conditions, we expect only minor interference in their comparison.

\looseness=-1
While several mechanisms that produce data quality issues were considered (such as longitudinal studies, watermarks, blurring, and different resolutions for near duplicates), exhaustively exploring all possibilities is unfeasible. 
It is likely that in some scenarios \textsc{SelfClean} can fail. 
A hint of this behavior can be seen in section~\ref{app:Investigation-VDR}.

Finally, we acknowledge that certain data quality issue types such as ambiguous labels were not investigated in this work.
Likewise, limited investigation was carried out on how to remedy identified issues, as this is expected to strongly depend on dataset and task.

\afterpage{
\begin{table}[htbp]
    \centering
    \small
    \caption{
        Hyperparameters used for pre-training using \gls*{simclr} and \gls*{dino} on the dataset to clean. 
        Here ``-'' indicates that the respective parameter is not used for the corresponding pre-training strategy. 
        Parameters not given in this list follow the introductory paper. 
        More detailed information about the hyperparameters can be found in the open-sourced codebase.
        }
    \label{tab:HyperPrams-Pre-Training}
    \begin{tabular}{l  r r}
        \toprule
        \textbf{Hyperparameter} & \textbf{SimCLR} \cite{chen_simple_2020} & \textbf{DINO} \cite{caron_emerging_2021} \\
        \midrule
        Batch size & 90 & 64 \\
        Epochs & 500 & 500 \\
        Optimizer & Adam & AdamW \\
        Learning rate & 0.001 & 0.0005 \\
        Min. learning rate & 1e-6 & 1e-6 \\
        Weight decay & 0.04 & 0.04 \\
        Weight decay end & 0.4 & 0.4 \\
        Warmup epochs & 10 & 10 \\
        Momentum teacher & - & 0.996 \\
        Clip grad. & 3.0 & 3.0 \\
        Base model & \gls*{vit}-tiny & \gls*{vit}-tiny \\
        Model embedding dim. & 192 & 192 \\
        Model output dim. & 128 & 4096 \\
        Model patch size & 16 & 16 \\
        Model drop path rate & 0.1 & 0.1 \\
        Norm last layer & - & True \\
        Global crops scale & - & (0.7, 1.) \\
        Local crops scale & - & (0.05, 0.4) \\
        Global crops number & - & 2 \\
        Local crops number & - & 12 \\
        Warmup teacher temp. & - & 0.04 \\
        Teacher temp. & - & 0.04 \\
        Warmup teacher temp. epochs & - & 30 \\
        Contrastive temp. & 0.5 & - \\
        \bottomrule
    \end{tabular}
\end{table}
\begin{table}[htbp]
    \small
    \caption{
        Configuration of the synthetic near duplicate strategies AUG (\ref{tab:Hyper-MED}) and ARTE (\ref{tab:Hyper-ARTE}). 
    }
    \label{tab:HyperPrams-Duplicate-Augmentation}
    \hfill
    \begin{subtable}[h]{0.45\textwidth}
        \centering
        \caption{AUG}
        \label{tab:Hyper-MED}
        \begin{tabular}{l r}
        \toprule
        \textbf{Hyperparameter} & \textbf{AUG} \\
        \midrule
        Rotation probability & 0.5 \\
        Padding probability & 0.5 \\
        Blur probability & 0.5 \\
        Rotation degree range & (0, 180) \\
        Scale range & (0.5, 0.9) \\
        Padding & 3 \\
        Gaussian kernel size & 5 \\
        \bottomrule
       \end{tabular}
    \end{subtable}
    \hfill
    \begin{subtable}[h]{0.45\textwidth}
        \centering
        \caption{ARTE}
        \label{tab:Hyper-ARTE}
        \begin{tabular}{l r}
        \toprule
        \textbf{Hyperparameter} & \textbf{ARTE} \\
        \midrule
        Watermark probability & 0.5 \\
        Colorbar probability & 0.5 \\
        Collage probability & 0.5 \\
        Watermark max. scale &  0.5 \\
        Collage max. scale & 0.5 \\
        Reference size & 512 \\
        \bottomrule
        \end{tabular}
     \end{subtable}
     \hfill
\end{table}
}

\section{Training details}\label{app:Training-Details}
We use a randomly initialized \gls*{vit}-tiny \citep{dosovitskiy_image_2021} with a patch size of $16\!\times\!16$ as encoder unless otherwise specified.
The latent representation is given by the class token output from the encoder's last layer, which has dimension 192 for a \gls*{vit}-tiny.

Table \ref{tab:HyperPrams-Pre-Training} lists the hyperparameters used for pre-training with \acrshort*{dino} \cite{caron_emerging_2021} and \acrshort*{simclr} \cite{chen_simple_2020}. 
Parameter values are similar to the introductory papers of these approaches \citep{chen_simple_2020,caron_emerging_2021} with the exception that for \gls*{dino} the global crop scale is larger and we sample more local crops, which we have found to be beneficial for smaller datasets ($<$20,000) while we observed no benefit or harm for larger datasets. 
All \gls*{ssl} models were pre-trained for 500 epochs with only minor manual hyperparameter tuning to ensure proper convergence. 
We resize images to $224\!\times\!224$ pixels and normalize them using the mean and standard deviation of ImageNet~\citep{deng_imagenet_2009}.

For the synthetic experiment setup, table \ref{tab:HyperPrams-Duplicate-Augmentation} lists the hyperparameters for producing near-duplicate images. 
The configuration was chosen to mimic the natural contamination of near duplicates in benchmark datasets.

\newpage
The implementation of \textsc{SelfClean} and the code used for evaluation are based on PyTorch 1.9~\citep{paszke_pytorch_2019} and can be found at
\begin{center}
    \url{https://github.com/Digital-Dermatology/SelfClean},
    and
    \url{https://github.com/Digital-Dermatology/SelfClean-Evaluation}.
\end{center}

Experiments were performed on an Nvidia DGX station, which features eight V100 GPUs, each with 32 GB of memory, 512 GB of system memory, and 40 CPU cores, for a total of 10,800 GPU hours which roughly correspond to 1,200 kg CO$_2$.

\section{Datasets}
\label{app:Datasets}

In this study, we selected twelve well-known, open-source image datasets comprising four general-purpose vision benchmarks and eight medical datasets.
These datasets contain different modalities, such as smartphone, X-ray, histopathology, dermatoscopy, and clinical images.
The diversity of the datasets and domains should illustrate \textsc{SelfClean}'s versatility.
Furthermore, some datasets were chosen because of their high-quality standards, as their curation involved extensive manual correction, including validation by multiple domain experts. 

\subsection{Datasets from the general image domain}

\looseness=-1
\textbf{ImageNet-1k (INet)} is a well-known image benchmark with 1,000 classes \citep{deng_imagenet_2009}. 
Images were scraped by querying words from WordNet's ``synonym sets'' (synsets) on several image search engines. 
The images were labeled by Amazon Mechanical Turk workers, who were asked whether each image contained objects of a given synset. 
License: Custom (research, non-commercial).

\looseness=-1
\textbf{STL-10 (STL)} is a benchmark consisting of 10 classes, each with 500 training images, 800 test images, and an additional 100,000 unlabeled images for unsupervised learning \citep{coates_analysis_2011}. 
It focuses on higher resolution images (96x96 pixels) compared to other similar collections like CIFAR-10. 
The images in STL-10 are sourced from labeled examples in ImageNet and are chosen to represent a broad range of object categories and real-world scenarios. 
License: Custom (attribution + ImageNet license).

\textbf{Food-101N} is an image dataset that contains 310,009 images of food divided into 101 classes \citep{lee_cleannet_2018}.
Both Food-101N and the Food-101 \citep{bossard14} dataset share the same 101 classes.
However, Food-101N has a significantly larger number of images and contains more noise.
The pictures were scraped from Google, Bing, Yelp, and TripAdvisor.
60,000 of them were manually verified and used for evaluation.
The evaluation set includes information for each sample on whether or not it features a label problem.
License: CC BY 4.0.

\textbf{CelebFaces Attributes Dataset (CelebA)} is a large-scale dataset with 202,599 celebrity face images, each with 40 attribute annotations \citep{liu2015faceattributes}.
The images in this dataset cover 10,177 identities, large pose variations and mixed backgrounds.
The CelebA dataset contains images of public figures, and while it is widely used in research, it is important to consider privacy, consent, and potential biases \citep{peng2021mitigating}. 
We have ensured that our usage complies with the dataset's terms and conditions, and we advise caution to avoid perpetuating any biases inherent in the dataset. 
Our work does not involve any manipulation or generation of images that could misrepresent individuals.
License: Custom (research, non-commercial).

\subsection{Datasets from the medical domain}

\textbf{CheXpert} is a large public dataset for chest radiograph interpretation, consisting of 224,316 X-ray scans from 65,240 patients \citep{irvin_chexpert_2019}.
The authors retrospectively collected chest radiographic examinations from Stanford Hospital, performed between October 2002 and July 2017 in both inpatient and outpatient centers, along with their associated radiology reports.
Labels were extracted from the free-text radiology reports with an automated rule-based system.
The dataset further contains radiologist-labeled reference evaluation sets.
License: Stanford University School of Medicine's Research Use Agreement.

\textbf{VinDr-BodyPartXR (VDR)} consists of 16,093 X-ray images that were manually annotated for body part classification~\cite{pham_dicom_2021}.
The authors differentiate between five groups, including abdominal, adult chest, pediatric chest, spine, and other X-rays.
The ``other'' category contains X-rays of any other body part, device malfunctions, and scans of clinical tools.
License: CC BY-NC 4.0.

\textbf{PatchCamelyon} consists of 327,680 color image patches extracted from histopathologic scans of lymph node sections \citep{Veeling2018-qh} from the Camelyon16 dataset \citep{ehteshami_bejnordi_diagnostic_2017}. 
Each patch is annotated with a binary label indicating the presence of metastatic tissue. 
Camelyon16 contains 399 whole-slide images and corresponding glass slides of sentinel axillary lymph nodes, which were retrospectively sampled from 399 patients who underwent breast cancer surgery at two hospitals in the Netherlands. 
All metastases in the slides were annotated under the supervision of multiple expert pathologists.
License: CC0.


\textbf{Diverse Dermatology Images (DDI)} is a public, deeply-curated, and pathologically-confirmed image dataset with diverse skin tones \citep{daneshjou_disparities_2022}.
It contains 656 clinical images of 570 unique patients with 78 common and uncommon diseases originating from pathology reports of the Stanford Clinics.
License: Stanford University School of Medicine's Research Use Agreement.

\textbf{PAD-UFES-20} is a public benchmark dataset composed of clinical images collected from smartphone devices including patient clinical data \citep{pacheco_pad-ufes-20_2020}. 
The dataset comprises 1,373 patients, 1,641 skin lesions, and 2,298 images for six different diagnoses: three skin diseases and three skin cancers.
License: CC BY 4.0.

\textbf{HAM10000} is a public benchmark dataset consisting of 10,015 dermatoscopic images collected from different populations and institutions \citep{tschandl_ham10000_2018}. 
The collected cases include a representative sample of seven categories of pigmented lesions.
License: CC BY-NC.

\textbf{Fitzpatrick17k (FST)} is a public benchmark dataset containing 16,577 clinical images with skin condition annotations and skin type labels based on the Fitzpatrick scoring system \citep{groh_evaluating_2021}. 
The images originate from two online dermatology atlases and thus are known to contain issues \citep{daneshjou_skincon_2022}. 
In this study, we used the middle granularity level, which partitions the labels into nine disease categories.
License: CC BY-NC-SA 3.0.

\textbf{High-Quality Fitzpatrick17k (HQ-FST)} is a subset of the Fitzpatrick17k dataset used in the paper \citep{groh_evaluating_2021} as a data quality check.
It was obtained by randomly selecting 3\% of the images (504 samples) and gathering annotations by two board-certified dermatologists to evaluate diagnostic accuracy.
This subset is assumed to be of much higher quality than its original, larger counterpart.
License: CC BY-NC-SA 3.0.

\textbf{ISIC-2019} is a public benchmark dataset of 25,331 dermoscopic images with metadata split into eight diagnostic categories. 
Additionally, the test set contains an additional outlier class not represented in the training data. 
The images originate from the HAM10000 \citep{tschandl_ham10000_2018} and the BCN\_20000 \citep{combalia_bcn20000_2019} datasets.
License: CC BY-NC-SA 3.0.

\section{Competing approaches}
\label{app:Competing-Approaches}
\looseness=-1
We selected different competing approaches to detect each of the three data quality issue categories, i.e., off-topic samples, near duplicates, and label errors.
Some of these methods require to encode images in a low-dimensional latent space.
For this projection, we used a \gls*{vit}-tiny, the same architecture used for \textsc{SelfClean}, pre-trained with supervision on ImageNet or with DINO self-supervision on each dataset.
We refer to these encoders with ``(INet)'' and ``(DINO)'' respectively, after the name of each detection approach.
In this section, we briefly summarize each competing approach used in this work.

\subsection{Approaches for off-topic samples}

\textbf{Isolation Forest (IForest)} isolates observations by randomly selecting a feature and splitting the value between the minimum and maximum of the selected feature.
The number of splits required to isolate a sample corresponds to the path length from the root node to the leaf node in a tree \citep{liu_isolation_2008}. 
This path length averaged over a forest of random trees is a measure of normality, where noticeably shorter paths are produced for anomalies.

\textbf{Histogram-based outlier detection (HBOS)} is an efficient unsupervised method that creates a histogram of the feature vector for each dimension and then calculates a score based on how likely a particular data point is to fall within the histogram bins for each dimension \citep{goldstein_histogram-based_2012}. 
The higher the score, the more likely the data point is an outlier, i.e., a feature vector coming from an anomaly will occupy unlikely bins in one or several of its dimensions and thus produce a higher anomaly score.

\textbf{Empirical Cumulative Distribution Functions (ECOD)} is a parameter-free, highly-interpretable unsupervised outlier detection algorithm \citep{li_ecod_2022}.
It estimates an empirical cumulative distribution function (ECDF) for each variable in the data separately. 
To generate an outlier score for an observation, it computes the tail probability for each variable using the univariate ECDFs and multiplies them together. 
This calculation is done in log space, accounting for each dimension's left and right tails.

\subsection{Approaches for near duplicates}

\textbf{Perceptual Hash (pHashing)} is a type of locality-sensitive hash, which is similar if features of the sample are similar \citep{marr_theory_1997}. 
It relies on the discrete cosine transform (DCT) for dimensionality reduction and produces hash bits depending on whether each DCT value is above or below the average value. In this paper, we use pHash with a hash size of 8.

\textbf{Structural Similarity Index Measure (SSIM)} is a type of similarity measure to compare two images with each other based on three features, namely luminance, contrast, and structure \citep{wang_image_2004}. 
Instead of applying SSIM globally, i.e., all over the image at once, one usually applies the metrics regionally, i.e., in small sections of the image, and takes the mean overall. 
This variant of SSIM is often called ``Mean Structural Similarity Index''.
In this paper, we apply SSIM locally to 8x8 windows but still refer to the method as SSIM for simplicity.

\subsection{Approaches for label errors}

\textbf{Confident Learning (CLearning)} is a data-centric approach that focuses on label quality by characterizing and identifying label errors in datasets based on the principles of pruning noisy data, counting with probabilistic thresholds to estimate noise, and ranking examples to train with confidence \citep{northcutt_confident_2022}. 
It builds upon the assumption of a class-conditional noise process to directly estimate the joint distribution between noisy (given) and uncorrupted (unknown) labels, resulting in a generalized learning process that is provably consistent and experimentally performant.
In this study, we use AdaBoost \citep{freund_short_1999} as a classifier on top of pre-trained representations to estimate probabilities. 
We did not observe any significant performance difference when using different classifiers similarly to \citet{northcutt_confident_2022}.

\textbf{NoiseRank (Noise)} is a method for unsupervised label noise detection using Markov Random Fields \citep{sharma_noiserank_2020}. 
It constructs a dependence model to estimate the posterior probability of an instance being incorrectly labeled, given the dataset, and then ranks instances based on this probability.

\subsection{Approaches for multiple issue types}

\textbf{FastDup} is an open-source, non-peer-reviewed tool designed to rapidly extract valuable insights from image and video datasets, aiming to increase the dataset quality and reduce data operations costs at an unparalleled scale~\citep{visuallayer2022fastdup}.
It detects outliers, duplicate, and near-duplicate images and videos, and wrongly labeled samples.

\section{Further ablation studies}
\label{app:FurtherAblation}
This section presents additional ablation studies that investigate different components of \textsc{SelfClean}.
Note that we cannot consistently use the same dataset for these ablation studies, as each ablation is most meaningful for a dataset with a specific domain and degree of cleanliness, also in relationship with the considered issue type and amount of required compute.

\subsection[Influence of L2-normalization and distance functions]{Influence of $L_2$-normalization and distance functions}
\label{app:Influence-L2-Distance}

\begin{figure}[htbp]
  \includegraphics[width=\linewidth]{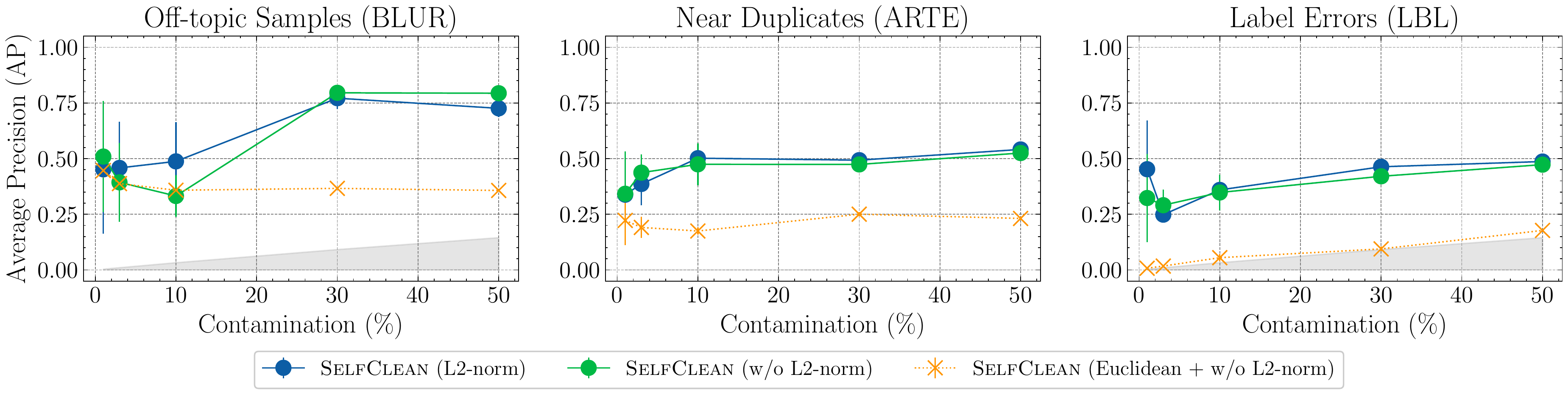}
  \caption{
    Performance of \textsc{SelfClean} when changing the distance function and removing the $L_2$-normalization.
    The performance is measured in terms of \acrfull*{ap} for a mixed-contamination strategy when varying the contamination rate.
    The artificial dataset is created from DDI by adding off-topic samples (BLUR), then injecting augmented duplicates (ARTE), and finally changing labels at random (LBL).
    Shaded regions indicate random performance.
  }
  \label{fig:Influence-L2-Distance}
\end{figure}

For \gls*{ssl} strategies without explicit normalization, we included $L_2$-normalization in the latent space during both training and inference (e.g., \gls*{dino}).
A similar explicit $L_2$-normalization for representation layers is also used in theoretical works on \gls*{ssl} \citep{dubois_improving_2022}, where it was inherited from the neural collapse literature \citep{e_emergence_2022}.
We investigate the influence of this $L_2$-normalization on the detection performance for the different dataset quality issues.
Figure~\ref{fig:Influence-L2-Distance} shows the performance of \textsc{SelfClean} with and without normalization.
The experiment is run on a 10\% mixed-contamination dataset, starting from DDI and creating off-topic samples using BLUR, near duplicates using ARTE, and label errors using LBL.
The results show that $L_2$-normalization has a mild, slightly positive effect on the performance.
One possible explanation for the improved performance is that limiting the latent space to the unit hypersphere enforces a more direct relation between the training objective and the relative distances of encoded samples.

Additionally, we examined the influence of the choice of the distance function between cosine and Euclidean distance.
Since the Euclidean and cosine distance on a $L_2$-normalized space always produce the same ranking, we only show the results of different distance functions for the non-normalized latent space.
Figure \ref{fig:Influence-L2-Distance} shows that performance is strongly influenced by the choice of distance function.
Specifically, using Euclidean distance leads to significantly lower performance.

\subsection{Influence of the number of pre-training epochs}
\label{app:Influence-Pretraining-Epochs}
We evaluate the learned representations after a different number of pre-training epochs to investigate the influence of the pre-training length.
The experiment is run on a 10\% mixed-contamination dataset, starting from DDI and creating off-topic samples using BLUR, near duplicates using ARTE, and label errors using LBL.
The performance of the representations is evaluated every 50 epochs for both representations with and without $L_2$-normalization.

\begin{figure}[htbp]
  \includegraphics[width=1.0\linewidth]{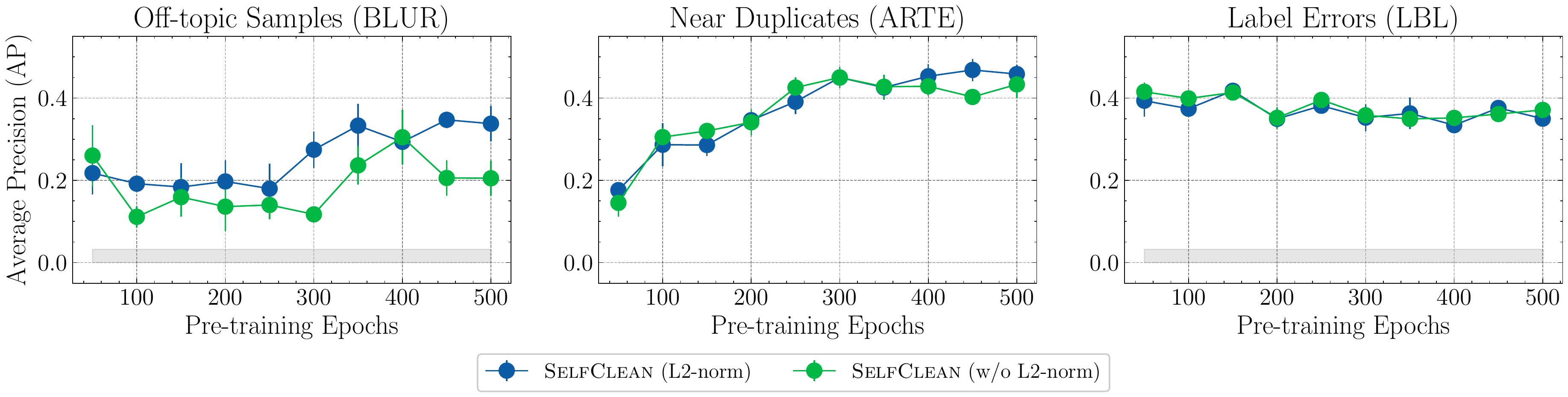}
  \caption{
    Performance of \textsc{SelfClean} during pre-training.
    The performance is measured in terms of \acrfull*{ap} for a 10\% mixed-contamination strategy.
    The artificial dataset is created from DDI by adding off-topic samples (BLUR), then injecting augmented duplicates (ARTE), and finally changing labels at random (LBL).
    Shaded regions indicate random performance.
  }
  \label{fig:Influence-Epochs}
\end{figure}

Figure \ref{fig:Influence-Epochs} shows that performance for off-topic sample and near duplicate detection increases with longer pre-training with $L_2$-normalization.
Without normalization, the performance for off-topic detection has no clear trend.
For label error detection, both methods first degrade slightly and later stabilize.
Overall, at least with $L_2$-normalization, longer pre-training leads to stronger performance.

\subsection{Influence of label granularity}
\label{app:Influence-NLabels}
We investigate the performance of label error detection on different label granularities using the high-quality Fitzpatrick17k dataset.
This dataset features three hierarchy levels with 3, 9, and 104 classes respectively, and it has only around 500 samples, which makes the task difficult.
In table \ref{tab:Results-Ablation-Label} we report results for synthetic label issues (i.e., LBL and LBLC) for 10\% contamination.
Overall, it is harder to detect label errors as granularity increases, in agreement with intuition.
We observe that \textsc{SelfClean} excels at coarse granularity, and performs similarly to other approaches for fine-grained classification.

\addtolength{\tabcolsep}{-3pt}
\begin{table*}[htbp]
    \centering
    \tiny
    \caption{
        Performance of models on the detection of label errors.
        Evaluation is performed for each of the two synthetic label error strategies
        across HQ-FST with three different label partitions.
        All scores are reported in percentages (\%).
    }
    \label{tab:Results-Ablation-Label}
    \begin{tabular}{lll  rr rr  rr rr  rr rr}%
        \toprule
        & \bfseries Method
        & \bfseries Rep.
        & \multicolumn{4}{c}{\bfseries 3-Partition} 
        & \multicolumn{4}{c}{\bfseries 9-Partition}
        & \multicolumn{4}{c}{\bfseries 104-Partition} \\
        \midrule

        \parbox[t]{2mm}{\multirow{7}{*}{\rotatebox[origin=c]{90}{Label Errors}}}
        & &
        & \multicolumn{2}{c}{\bfseries LBL} 
        & \multicolumn{2}{c}{\bfseries LBLC}
        & \multicolumn{2}{c}{\bfseries LBL} 
        & \multicolumn{2}{c}{\bfseries LBLC}
        & \multicolumn{2}{c}{\bfseries LBL}
        & \multicolumn{2}{c}{\bfseries LBLC} \\

        \cmidrule(lr){4-5}
        \cmidrule(lr){6-7}
        \cmidrule(lr){8-9}
        \cmidrule(lr){10-11}
        \cmidrule(lr){12-13}
        \cmidrule(lr){14-15}
        
        & &
            & AUROC & AP
            & AUROC & AP
            & AUROC & AP
            & AUROC & AP
            & AUROC & AP
            & AUROC & AP \\
        \cmidrule{2-15}
            & CLearning \citep{northcutt_confident_2022}
            & INet
                & 79.7 & 30.7
                & 80.2 & 27.5
                & \textbf{84.6} & \textbf{40.4}
                & 69.0 & 26.2
                & 46.2 & 11.5
                & 51.7 & 23.2 \\
            & NoiseRank \citep{sharma_noiserank_2020}        
            & INet
                & 50.8 & 10.4
                & 57.0 & 12.1
                & 53.4 & 17.7
                & 60.4 & 23.8
                & 65.1 & \textbf{34.9}
                & 58.6 & 35.6 \\
            & FastDup \citep{visuallayer2022fastdup}
            & INet
                & 64.8 & 14.6
                & 74.1 & 21.4
                & 70.2 & 18.6
                & 68.3 & 21.1
                & 52.6 & 12.4
                & 55.8 & 20.2 \\
        & \cellcolor{Light}\textsc{SelfClean}
        & \cellcolor{Light}INet
                & \cellcolor{Light}76.8 & \cellcolor{Light}28.7
                & \cellcolor{Light}79.2 & \cellcolor{Light}29.9
                & \cellcolor{Light}81.2 & \cellcolor{Light}30.3
                & \cellcolor{Light}78.3 & \cellcolor{Light}31.9
                & \cellcolor{Light}67.1 & \cellcolor{Light}20.8
                & \cellcolor{Light}\textbf{73.4} & \cellcolor{Light}\textbf{44.3} \\
        & \cellcolor{Light}\textsc{SelfClean}
        & \cellcolor{Light}DINO
                & \cellcolor{Light}\textbf{80.9} & \cellcolor{Light}\textbf{35.4}
                & \cellcolor{Light}\textbf{85.3} & \cellcolor{Light}\textbf{40.0}
                & \cellcolor{Light}80.3 & \cellcolor{Light}28.3
                & \cellcolor{Light}\textbf{78.7} & \cellcolor{Light}\textbf{32.8}
                & \cellcolor{Light}\textbf{68.8} & \cellcolor{Light}19.2
                & \cellcolor{Light}70.5 & \cellcolor{Light}39.5 \\
        \bottomrule
    \end{tabular}
\end{table*}
\addtolength{\tabcolsep}{+3pt}

\subsection{Influence of the type of features}
\label{app:Influence-Features}

We investigate the influence of different types of features, i.e., general, domain-specific, and dataset-specific features.
The experiment is run on a 10\% mixed-contamination dataset, starting from VDR and DDI.
Table~\ref{tab:Ablation-Features} shows that for VDR, domain-specific features have the strongest overall performance followed by dataset-specific features.
General supervised and self-supervised features both fail at near duplicates even if they show strong performance on label errors.
For DDI, dataset-specific features yield the best performance, followed by domain-specific, self-supervised general, and supervised general.
For both datasets, these results show the importance of learning representations that successfully capture the task's context in order to achieve good detection performance.

\addtolength{\tabcolsep}{-3pt}
\begin{table*}[htbp]
    \centering
    \caption{
        Ablation of the feature types, i.e., general, domain-specific, and dataset-specific features.
        We use a different 10\% mixed contaminated dataset starting from VDR and DDI.
        Scores are \acrfull*{ap} percentages, and aggregated across the three tasks using the harmonic mean.
    }
    \scriptsize
    \label{tab:Ablation-Features}
    \begin{tabular}{@{} l l @{\hspace{5mm}} rrr @{\hspace{5mm}} r @{}}
    \toprule
	   Dataset & Pre-training 
            & OT (\%) & ND (\%) & LE (\%) 
            & H. Mean \\
   \midrule
        \multicolumn{6}{@{}l}{\textit{VDR (XR+AUG+LBLC)}}\\
        INet & Supervised
            & 28.5 & $<$ 0.1 & \textbf{95.8} 
            & $<$ 0.1 \\
        INet & \gls*{dino}
            & 28.7 & $<$ 0.1 & 95.5 
            & $<$ 0.1 \\
        CheXpert & \gls*{dino}
            & \textbf{31.3} & \textbf{0.3} & 94.8 
            & \textbf{0.9} \\
        VDR & \gls*{dino}
            & 23.3 & 0.1 & 95.0
            & 0.3 \\
    \midrule
        \multicolumn{6}{@{}l}{\textit{DDI (BLUR+ARTE+LBL)}}\\
        INet & Supervised
            & 2.6 & 37.8 & 22.4
            & 6.6 \\
        INet & \gls*{dino}
            & 18.6 & 28.8 & \textbf{36.7}
            & 25.9 \\
        HAM10000 & \gls*{dino} 
            & 29.1 & 27.1 & 31.2
            & 29.0 \\
        DDI & \gls*{dino}
            & \textbf{33.2} & \textbf{47.4} & 34.8
            & \textbf{37.5} \\
    \bottomrule
  \end{tabular}
\end{table*}
\addtolength{\tabcolsep}{+3pt}

\subsection[Influence of the self-supervised learning objective]{Influence of the \acrlong*{ssl} objective}
\label{app:Influence-SSL}
We investigate further \gls*{ssl} objectives for detecting data quality issues.
In addition to \gls*{simclr} and \gls*{dino}, which are used throughout the paper, we include \gls*{byol} \citep{grill_bootstrap_2020} and \gls*{mae} \citep{he_masked_2021}.
The experiment is run on a 10\% mixed-contamination dataset, starting from STL and creating off-topic samples using XR, near duplicates using AUG, and label errors using LBLC.
Table \ref{tab:Ablation-SSL} shows that \gls*{dino} has the strongest overall performance.
Some \gls*{ssl} objectives only obtain strong results for specific issue types.
This is the case for \gls*{byol}, which separates off-topic samples well but fails on near duplicates and label errors.
Other methods, such as \gls*{mae} and \gls*{simclr}, achieve similar performance across issue types, although significantly lower than \gls*{dino}.

\addtolength{\tabcolsep}{-3pt}
\begin{table*}[htbp]
    \centering
    \caption{
        Ablation of the pre-training strategy.
        We use a 10\% mixed contaminated dataset starting from STL and creating off-topic samples (OT) using XR, near duplicates (ND) using AUG, and label errors (LE) using LBLC.
        Scores are \acrfull*{ap} percentages, and aggregated across the three tasks using the harmonic mean.
    }
    \scriptsize
    \label{tab:Ablation-SSL}
    \begin{tabular}{@{} l l @{\hspace{5mm}} rrr @{\hspace{5mm}} r @{}}
    \toprule
	  Pre-training & Dataset
            & OT (\%) & ND (\%) & LE (\%) 
            & H. Mean \\
   \midrule
        \gls*{simclr} \cite{chen_simple_2020} & STL
            & 26.1 & 12.1 & 15.8 
            & 16.3\\
        \gls*{byol} \cite{grill_bootstrap_2020} & STL
            & \textbf{29.7} & $<$ 0.1 & 3.5 
            & $<$ 0.1 \\
        \gls*{mae} \cite{he_masked_2021} & STL
            & 8.3 & 18.1 & 17.7 
            & 12.9 \\
        \gls*{dino} \cite{caron_emerging_2021} & STL
            & 27.4 & \textbf{47.1} & \textbf{24.8} 
            & \textbf{30.6} \\
    \bottomrule
  \end{tabular}
\end{table*}
\addtolength{\tabcolsep}{+3pt}

\subsection{Influence of the encoder architecture}
\label{app:Influence-Encoder}

We investigate further encoder architectures in table \ref{tab:Ablation-encoder}.
In addition to the \gls*{vit}-tiny with a patch size of $16\!\times\!16$ used throughout the paper, we include larger and different types of architectures, i.e., \glspl*{vit} \cite{dosovitskiy_image_2021} and ResNets \cite{he_deep_2016}.
The experiment is run on a 10\% mixed-contamination dataset, starting from STL and creating off-topic samples using XR, near duplicates using AUG, and label errors using LBLC.
Results indicate that the smaller models (i.e., \gls*{vit}-tiny and ResNet-18) produce stable results with \gls*{dino} pre-training, although \glspl*{vit} show overall superior performance, similar as found in \cite{caron_emerging_2021}.
Label error detection works best with supervised training as already observed in section \ref{ssec:ablation}, presumably because ImageNet and STL have very similar contexts.
Furthermore, for label errors, performance with supervised pre-training increases with model size.
Larger models (i.e., \gls*{vit}-small and ResNet-50) show mixed results, likely because of the small pre-training dataset of 5,000 samples.

\addtolength{\tabcolsep}{-3pt}
\begin{table*}[htbp]
    \centering
    \caption{
        Ablation of the encoder architecture, i.e., \gls*{vit} \cite{dosovitskiy_image_2021} and ResNet \cite{he_deep_2016}.
        We use a 10\% mixed contaminated dataset starting from STL and creating off-topic samples (OT) using XR, near duplicates (ND) using AUG, and label errors (LE) using LBLC.
        Scores are \acrfull*{ap} percentages, and aggregated across the three tasks using the harmonic mean.
    }
    \scriptsize
    \label{tab:Ablation-encoder}
    \begin{tabular}{@{} llll @{\hspace{5mm}} rrr @{\hspace{5mm}} r @{}}
    \toprule
	  Encoder & N.o. Parameters & Pre-training & Dataset 
            & OT (\%) & ND (\%) & LE (\%) 
            & H. Mean \\
   \midrule
        \multirow{2}{*}{\gls*{vit}-tiny $16\!\times\!16$} & \multirow{2}{*}{5.5 Mio.} & Supervised & INet 
            & 1.6 & 24.6 & \textbf{63.0} 
            & 4.4 \\
        & & \gls*{dino} & STL 
            & \textbf{27.4} & \textbf{47.1} & 24.8
            & \textbf{30.6} \\
   \midrule
        \multirow{2}{*}{ResNet-18} & \multirow{2}{*}{11.7 Mio.} & Supervised & INet 
            & 4.6 & 4.4 & \textbf{94.8}
            & 6.6 \\
        & & \gls*{dino} & STL 
            & \textbf{14.8} & \textbf{22.9} & 30.1
            & \textbf{20.8} \\
   \midrule
        \multirow{2}{*}{\gls*{vit}-small $16\!\times\!16$} & \multirow{2}{*}{21.7 Mio.} & Supervised & INet 
            & \textbf{2.4} & \textbf{20.9} & \textbf{94.5}
            & \textbf{6.3} \\
        & & \gls*{dino} & STL 
            & 1.8 & 20.7 & 44.0
            & 4.8 \\
   \midrule
        \multirow{2}{*}{ResNet-50} & \multirow{2}{*}{25.6 Mio.} & Supervised & INet 
            & \textbf{10.1} & 1.7 & \textbf{96.5}
            & 4.3 \\
        & & \gls*{dino} & STL 
            & 4.1 & \textbf{25.1} & 69.1
            & \textbf{10.1} \\
    \bottomrule
  \end{tabular}
\end{table*}
\addtolength{\tabcolsep}{+3pt}

\newpage
\section{Detailed dataset cleaning results}

This section provides extended tables with performance results related to dataset cleaning.
More precisely, section \ref{app:syntheticcleaningtable} investigates synthetic contamination detection with different methods, metrics, and contamination levels, expanding on section \ref{ssec:syntheticresults}.
Section \ref{app:Detailed-Meta} presents in tabular form the comparison of \textsc{SelfClean} with available metadata as discussed in section \ref{ssec:naturalresults}.
Section \ref{app:Detailed-DatasetCleaning} extends table \ref{tab:cleaninginfluencemain} in section \ref{sec:discussion} by including information on the performances used to compute paired differences.

\subsection{Detailed comparison on synthetic data quality issues}
\label{app:syntheticcleaningtable}

Table \ref{tab:Results-Detailed} details results of the comparison of synthetic data quality issues.
Conclusions are drawn in section \ref{ssec:syntheticresults}.

\addtolength{\tabcolsep}{-3pt}
\begin{table*}[htbp]
    \centering
    \tiny
    \caption{
        Performance of various models on the detection of synthetic data quality issues.
        Evaluation is performed for each of the three considered issue types
        across three benchmark datasets, STL, VDR, and DDI,
        augmented with different strategies for synthetic contamination (XR, BLUR, AUG, ARTE, LBL, and LBLC).
        All scores are reported in percentages (\%).
    }
    \label{tab:Results-Detailed}
    \resizebox{0.9\linewidth}{!}{%
    \begin{tabular}{@{} l ll rr rr rr rr rr rr}%
        \toprule
        & \bfseries Method
        & \bfseries Rep.
        & \multicolumn{12}{c}{\bfseries Contamination 5\%} \\
        \midrule
        \parbox[t]{2mm}{\multirow{9}{*}{\rotatebox[origin=c]{90}{Off-topic Samples}}}
        &
        & & \multicolumn{2}{c}{\bfseries STL + XR}
            & \multicolumn{2}{c}{\bfseries STL + BLUR} 
            & \multicolumn{2}{c}{\bfseries VDR + BLUR}
            & \multicolumn{2}{c}{\bfseries VDR + XR} 
            & \multicolumn{2}{c}{\bfseries DDI + XR} 
            & \multicolumn{2}{c}{\bfseries DDI + BLUR} \\

        \cmidrule(lr){4-5}
        \cmidrule(lr){6-7}
        \cmidrule(lr){8-9}
        \cmidrule(lr){10-11}
        \cmidrule(lr){12-13}
        \cmidrule(lr){14-15}
            
        & &
            & AUROC & AP
            & AUROC & AP
            & AUROC & AP
            & AUROC & AP
            & AUROC & AP
            & AUROC & AP \\
        \cmidrule{2-15}
            & IForest \citep{liu_isolation_2008}
            & INet
                & 68.2 & 7.0
                & 1.6 & 2.6
                & 94.0 & 31.1
                & 81.2 & 22.6
                & 93.3 & 59.5
                & 21.5 & 3.1 \\
            & HBOS \citep{goldstein_histogram-based_2012}
            & INet
                & 66.9 & 6.6
                & 1.9 & 2.6
                & 95.7 & 36.6
                & 82.3 & 24.4
                & 93.0 & 68.0
                & 19.0 & 3.0 \\
            & ECOD \citep{li_ecod_2022}
            & INet
                & 68.4 & 7.0
                & 2.2 & 2.6
                & 95.0 & 34.1
                & 81.4 & 25.7
                & 92.8 & 68.0
                & 23.6 & 3.1 \\
            & FastDup \citep{visuallayer2022fastdup}
            & INet
                & 4.1 & 2.5
                & 8.3 & 2.6
                & 25.1 & 7.7
                & 69.6 & 20.0
                & 53.5 & 29.5
                & 19.7 & 3.1 \\
                
        & \cellcolor{Light}\textsc{SelfClean}
        & \cellcolor{Light}INet
                & \cellcolor{Light}11.4 & \cellcolor{Light}2.7
                & \cellcolor{Light}67.7 & \cellcolor{Light}7.3
                & \cellcolor{Light}99.9 & \cellcolor{Light}91.2
                & \cellcolor{Light}77.1 & \cellcolor{Light}32.8
                & \cellcolor{Light}98.9 & \cellcolor{Light}84.2
                & \cellcolor{Light}86.5 & \cellcolor{Light}18.2 \\
        & \cellcolor{Light}\textsc{SelfClean}
        & \cellcolor{Light}SimCLR
                & \cellcolor{Light}40.6 & \cellcolor{Light}3.9
                & \cellcolor{Light}77.4 & \cellcolor{Light}19.0
                & \cellcolor{Light}\textbf{100.0} & \cellcolor{Light}98.7
                & \cellcolor{Light}86.0 & \cellcolor{Light}35.5
                & \cellcolor{Light}99.0 & \cellcolor{Light}68.9
                & \cellcolor{Light}70.0 & \cellcolor{Light}21.9 \\
        & \cellcolor{Light}\textsc{SelfClean} 
        & \cellcolor{Light}DINO
                & \cellcolor{Light}\textbf{98.4} & \cellcolor{Light}\textbf{55.1}
                & \cellcolor{Light}\textbf{100.0} & \cellcolor{Light}\textbf{97.9}
                & \cellcolor{Light}\textbf{100.0} & \cellcolor{Light}\textbf{100.0}
                & \cellcolor{Light}\textbf{95.6} & \cellcolor{Light}\textbf{53.3}
                & \cellcolor{Light}\textbf{100.0} & \cellcolor{Light}\textbf{100.0}
                & \cellcolor{Light}\textbf{86.8} & \cellcolor{Light}\textbf{32.6} \\
        \midrule
        \parbox[t]{2mm}{\multirow{8}{*}{\rotatebox[origin=c]{90}{Near Duplicates}}} & 
        & & \multicolumn{2}{c}{\bfseries STL + AUG}
            & \multicolumn{2}{c}{\bfseries STL + ARTE}
            & \multicolumn{2}{c}{\bfseries VDR + AUG}
            & \multicolumn{2}{c}{\bfseries VDR + ARTE}
            & \multicolumn{2}{c}{\bfseries DDI + AUG} 
            & \multicolumn{2}{c}{\bfseries DDI + ARTE} \\

            \cmidrule(lr){4-5}
            \cmidrule(lr){6-7}
            \cmidrule(lr){8-9}
            \cmidrule(lr){10-11}
            \cmidrule(lr){12-13}
            \cmidrule(lr){14-15}
            
        & &
            & AUROC & AP
            & AUROC & AP
            & AUROC & AP
            & AUROC & AP
            & AUROC & AP
            & AUROC & AP \\
        \cmidrule{2-15}
            & pHashing \citep{marr_theory_1997}
            &
                & 57.8 & $<$ 0.1
                & 73.1 & 20.1
                & 47.5 & $<$ 0.1
                & 57.5 & 18.2
                & 59.4 & 0.1
                & 66.2 & 15.1 \\
            & SSIM \citep{wang_image_2004}. 
            &
                & 62.5 & 0.2
                & 83.6 & 19.9
                & 46.3 & $<$ 0.1
                & 48.4 & \textbf{22.5}
                & 57.6 & 0.2
                & 83.0 & 19.4 \\
            & FastDup \citep{visuallayer2022fastdup}
            & INet
                & 50.2 & 2.2
                & 49.2 & 3.3
                & 37.6 & $<$ 0.1
                & 40.1 & 2.9
                & 56.2 & 4.8
                & 44.6 & 7.1 \\
                
        & \cellcolor{Light}\textsc{SelfClean}
        & \cellcolor{Light}INet
                & \cellcolor{Light}96.6 & \cellcolor{Light}7.6
                & \cellcolor{Light}96.5 & \cellcolor{Light}15.2
                & \cellcolor{Light}79.7 & \cellcolor{Light}$<$ 0.1
                & \cellcolor{Light}53.7 & \cellcolor{Light}11.1
                & \cellcolor{Light}97.6 & \cellcolor{Light}4.1
                & \cellcolor{Light}81.1 & \cellcolor{Light}34.4 \\
        & \cellcolor{Light}\textsc{SelfClean}
        & \cellcolor{Light}SimCLR
                & \cellcolor{Light}86.1 & \cellcolor{Light}0.1
                & \cellcolor{Light}93.8 & \cellcolor{Light}13.9
                & \cellcolor{Light}76.1 & \cellcolor{Light}$<$ 0.1
                & \cellcolor{Light}78.9 & \cellcolor{Light}12.6
                & \cellcolor{Light}89.8 & \cellcolor{Light}1.6
                & \cellcolor{Light}87.2 & \cellcolor{Light}0.7 \\
        & \cellcolor{Light}\textsc{SelfClean}
        & \cellcolor{Light}DINO
                & \cellcolor{Light}\textbf{100.0} & \cellcolor{Light}\textbf{43.7}
                & \cellcolor{Light}\textbf{99.9} & \cellcolor{Light}\textbf{48.0}
                & \cellcolor{Light}\textbf{98.5} & \cellcolor{Light}\textbf{0.4}
                & \cellcolor{Light}\textbf{91.6} & \cellcolor{Light}16.8
                & \cellcolor{Light}\textbf{99.7} & \cellcolor{Light}\textbf{50.8}
                & \cellcolor{Light}\textbf{98.2} & \cellcolor{Light}\textbf{48.2} \\
        \midrule
        \parbox[t]{2mm}{\multirow{8}{*}{\rotatebox[origin=c]{90}{Label Errors}}} 
        & 
            
        & & \multicolumn{2}{c}{\bfseries STL + LBL} 
            & \multicolumn{2}{c}{\bfseries STL + LBLC}
            & \multicolumn{2}{c}{\bfseries VDR + LBL}
            & \multicolumn{2}{c}{\bfseries VDR + LBLC}
            & \multicolumn{2}{c}{\bfseries DDI + LBL}
            & \multicolumn{2}{c}{\bfseries DDI + LBLC} \\

            \cmidrule(lr){4-5}
            \cmidrule(lr){6-7}
            \cmidrule(lr){8-9}
            \cmidrule(lr){10-11}
            \cmidrule(lr){12-13}
            \cmidrule(lr){14-15}
            
            & &
            & AUROC & AP
            & AUROC & AP
            & AUROC & AP
            & AUROC & AP
            & AUROC & AP
            & AUROC & AP \\
        \cmidrule{2-15}
            & CLearning \citep{northcutt_confident_2022}
            & INet
                & 86.2 & 41.6
                & 83.2 & 36.8
                & 96.7 & 79.0
                & 96.8 & 74.9
                & 67.9 & 11.0
                & 75.0 & 12.9 \\
            & NoiseRank \citep{sharma_noiserank_2020}        
            & INet
                & 49.5 & 5.0
                & 51.4 & 5.4
                & 48.9 & 5.3
                & 51.8 & 5.3
                & 51.4 & 5.8
                & 52.0 & 6.1 \\
            & FastDup \citep{visuallayer2022fastdup}
            & INet
                & 87.5 & 20.5
                & 87.0 & 19.8
                & 95.0 & 38.9
                & 94.1 & 37.8
                & 69.0 & 8.6
                & 69.9 & 11.6 \\
                
        & \cellcolor{Light}\textsc{SelfClean}
        & \cellcolor{Light}INet
                & \cellcolor{Light}\textbf{97.7} & \cellcolor{Light}\textbf{77.6}
                & \cellcolor{Light}\textbf{97.9} & \cellcolor{Light}\textbf{76.4}
                & \cellcolor{Light}98.5 & \cellcolor{Light}84.6
                & \cellcolor{Light}98.5 & \cellcolor{Light}84.8
                & \cellcolor{Light}67.8 & \cellcolor{Light}11.6
                & \cellcolor{Light}\textbf{79.8} & \cellcolor{Light}18.3 \\
        & \cellcolor{Light}\textsc{SelfClean}
        & \cellcolor{Light}SimCLR
                & \cellcolor{Light}79.1 & \cellcolor{Light}27.4
                & \cellcolor{Light}77.4 & \cellcolor{Light}26.5
                & \cellcolor{Light}95.0 & \cellcolor{Light}62.2
                & \cellcolor{Light}95.4 & \cellcolor{Light}64.4
                & \cellcolor{Light}64.8 & \cellcolor{Light}8.3
                & \cellcolor{Light}69.0 & \cellcolor{Light}11.1 \\
        & \cellcolor{Light}\textsc{SelfClean}
        & \cellcolor{Light}DINO
                & \cellcolor{Light}90.7 & \cellcolor{Light}54.2
                & \cellcolor{Light}91.1 & \cellcolor{Light}48.3
                & \cellcolor{Light}\textbf{99.2} & \cellcolor{Light}\textbf{88.1}
                & \cellcolor{Light}\textbf{99.0} & \cellcolor{Light}\textbf{85.6}
                & \cellcolor{Light}\textbf{71.4} & \cellcolor{Light}\textbf{13.5}
                & \cellcolor{Light}71.7 & \cellcolor{Light}\textbf{21.4} \\
        \midrule
        & & & \multicolumn{12}{c}{\bfseries Contamination 10\%} \\
        \midrule
        \parbox[t]{2mm}{\multirow{9}{*}{\rotatebox[origin=c]{90}{Off-topic Samples}}}
        &
        & & \multicolumn{2}{c}{\bfseries STL + XR}
            & \multicolumn{2}{c}{\bfseries STL + BLUR} 
            & \multicolumn{2}{c}{\bfseries VDR + BLUR}
            & \multicolumn{2}{c}{\bfseries VDR + XR} 
            & \multicolumn{2}{c}{\bfseries DDI + XR} 
            & \multicolumn{2}{c}{\bfseries DDI + BLUR} \\

        \cmidrule(lr){4-5}
        \cmidrule(lr){6-7}
        \cmidrule(lr){8-9}
        \cmidrule(lr){10-11}
        \cmidrule(lr){12-13}
        \cmidrule(lr){14-15}
            
        & &
            & AUROC & AP
            & AUROC & AP
            & AUROC & AP
            & AUROC & AP
            & AUROC & AP
            & AUROC & AP \\
        \cmidrule{2-15}
            & IForest \citep{liu_isolation_2008}
            & INet
                & 45.5 & 7.6
                & 1.6 & 5.2
                & 81.4 & 21.9
                & 75.3 & 30.6
                & 91.6 & 61.5
                & 12.3 & 5.6 \\
            & HBOS \citep{goldstein_histogram-based_2012}
            & INet
                & 48.7 & 7.9
                & 1.2 & 5.2
                & 87.6 & 28.8
                & 81.0 & 35.4
                & 95.8 & 78.0
                & 13.5 & 5.7 \\
            & ECOD \citep{li_ecod_2022}
            & INet
                & 52.6 & 8.6
                & 1.6 & 5.2
                & 89.1 & 32.5
                & 79.4 & 33.6
                & 95.1 & 76.7
                & 19.3 & 5.9 \\
            & FastDup \citep{visuallayer2022fastdup}
            & INet
                & 2.9 & 4.7
                & 7.9 & 5.4
                & 22.1 & 7.7
                & 64.8 & 24.2
                & 31.7 & 22.5
                & 14.5 & 5.9 \\
                
        & \cellcolor{Light}\textsc{SelfClean}
        & \cellcolor{Light}INet
                & \cellcolor{Light}0.7 & \cellcolor{Light}4.7
                & \cellcolor{Light}47.3 & \cellcolor{Light}9.1
                & \cellcolor{Light}99.8 & \cellcolor{Light}93.3
                & \cellcolor{Light}74.0 & \cellcolor{Light}38.8
                & \cellcolor{Light}94.6 & \cellcolor{Light}72.8
                & \cellcolor{Light}82.3 & \cellcolor{Light}22.4 \\
        & \cellcolor{Light}\textsc{SelfClean}
        & \cellcolor{Light}SimCLR
                & \cellcolor{Light}36.0 & \cellcolor{Light}7.1
                & \cellcolor{Light}75.4 & \cellcolor{Light}46.3
                & \cellcolor{Light}99.8 & \cellcolor{Light}96.4
                & \cellcolor{Light}86.5 & \cellcolor{Light}43.5
                & \cellcolor{Light}96.4 & \cellcolor{Light}58.6
                & \cellcolor{Light}80.4 & \cellcolor{Light}32.2 \\
        & \cellcolor{Light}\textsc{SelfClean} 
        & \cellcolor{Light}DINO
                & \cellcolor{Light}\textbf{97.6} & \cellcolor{Light}\textbf{60.8}
                & \cellcolor{Light}\textbf{100.0} & \cellcolor{Light}\textbf{97.8}
                & \cellcolor{Light}\textbf{100.0} & \cellcolor{Light}\textbf{100.0}
                & \cellcolor{Light}\textbf{96.9} & \cellcolor{Light}\textbf{62.1}
                & \cellcolor{Light}\textbf{100.0} & \cellcolor{Light}\textbf{100.0}
                & \cellcolor{Light}\textbf{88.4} & \cellcolor{Light}\textbf{55.4} \\
        \midrule
        \parbox[t]{2mm}{\multirow{8}{*}{\rotatebox[origin=c]{90}{Near Duplicates}}} & 
        & & \multicolumn{2}{c}{\bfseries STL + AUG}
            & \multicolumn{2}{c}{\bfseries STL + ARTE}
            & \multicolumn{2}{c}{\bfseries VDR + AUG}
            & \multicolumn{2}{c}{\bfseries VDR + ARTE}
            & \multicolumn{2}{c}{\bfseries DDI + AUG} 
            & \multicolumn{2}{c}{\bfseries DDI + ARTE} \\

            \cmidrule(lr){4-5}
            \cmidrule(lr){6-7}
            \cmidrule(lr){8-9}
            \cmidrule(lr){10-11}
            \cmidrule(lr){12-13}
            \cmidrule(lr){14-15}
            
        & &
            & AUROC & AP
            & AUROC & AP
            & AUROC & AP
            & AUROC & AP
            & AUROC & AP
            & AUROC & AP \\
        \cmidrule{2-15}
            & pHashing \citep{marr_theory_1997}
            &
                & 53.9 & 0.1
                & 73.2 & 22.5
                & 46.2 & $<$ 0.1
                & 56.9 & 19.3
                & 56.5 & 0.5
                & 72.6 & 25.5 \\
            & SSIM \citep{wang_image_2004}. 
            &
                & 62.8 & 0.2
                & 83.8 & 22.5
                & 49.4 & 0.1
                & 50.2 & \textbf{22.3}
                & 57.3 & 0.9
                & 80.6 & 26.3 \\
            & FastDup \citep{visuallayer2022fastdup}
            & INet
                & 54.5 & 3.3
                & 54.9 & 5.5
                & 37.1 & $<$ 0.1
                & 44.5 & 3.9
                & 58.8 & 3.3
                & 54.7 & 4.9 \\
                
        & \cellcolor{Light}\textsc{SelfClean}
        & \cellcolor{Light}INet
                & \cellcolor{Light}96.2 & \cellcolor{Light}17.9
                & \cellcolor{Light}96.8 & \cellcolor{Light}17.9
                & \cellcolor{Light}80.9 & \cellcolor{Light}$<$ 0.1
                & \cellcolor{Light}54.9 & \cellcolor{Light}13.4
                & \cellcolor{Light}98.2 & \cellcolor{Light}12.5
                & \cellcolor{Light}82.0 & \cellcolor{Light}25.9 \\
        & \cellcolor{Light}\textsc{SelfClean}
        & \cellcolor{Light}SimCLR
                & \cellcolor{Light}81.7 & \cellcolor{Light}0.1
                & \cellcolor{Light}93.5 & \cellcolor{Light}10.3
                & \cellcolor{Light}70.4 & \cellcolor{Light}$<$ 0.1
                & \cellcolor{Light}77.5 & \cellcolor{Light}12.9
                & \cellcolor{Light}89.0 & \cellcolor{Light}0.3
                & \cellcolor{Light}61.1 & \cellcolor{Light}$<$ 0.1 \\
        & \cellcolor{Light}\textsc{SelfClean}
        & \cellcolor{Light}DINO
                & \cellcolor{Light}\textbf{100.0} & \cellcolor{Light}\textbf{51.0}
                & \cellcolor{Light}\textbf{99.9} & \cellcolor{Light}\textbf{46.4}
                & \cellcolor{Light}\textbf{98.7} & \cellcolor{Light}\textbf{0.3}
                & \cellcolor{Light}\textbf{88.5} & \cellcolor{Light}14.3
                & \cellcolor{Light}\textbf{99.3} & \cellcolor{Light}\textbf{49.0}
                & \cellcolor{Light}\textbf{97.4} & \cellcolor{Light}\textbf{49.8} \\
        \midrule
        \parbox[t]{2mm}{\multirow{8}{*}{\rotatebox[origin=c]{90}{Label Errors}}} 
        & 
            
        & & \multicolumn{2}{c}{\bfseries STL + LBL} 
            & \multicolumn{2}{c}{\bfseries STL + LBLC}
            & \multicolumn{2}{c}{\bfseries VDR + LBL}
            & \multicolumn{2}{c}{\bfseries VDR + LBLC}
            & \multicolumn{2}{c}{\bfseries DDI + LBL}
            & \multicolumn{2}{c}{\bfseries DDI + LBLC} \\

            \cmidrule(lr){4-5}
            \cmidrule(lr){6-7}
            \cmidrule(lr){8-9}
            \cmidrule(lr){10-11}
            \cmidrule(lr){12-13}
            \cmidrule(lr){14-15}
            
            & &
            & AUROC & AP
            & AUROC & AP
            & AUROC & AP
            & AUROC & AP
            & AUROC & AP
            & AUROC & AP \\
        \cmidrule{2-15}
            & CLearning \citep{northcutt_confident_2022}
            & INet
                & 83.5 & 47.9
                & 85.0 & 46.4
                & \textbf{97.4} & \textbf{85.1}
                & 97.4 & 84.5
                & 73.7 & 25.6
                & 72.5 & 24.9 \\
            & NoiseRank \citep{sharma_noiserank_2020}        
            & INet
                & 49.4 & 10.0
                & 50.0 & 10.3
                & 51.5 & 10.5
                & 51.5 & 10.8
                & 51.7 & 11.1
                & 50.3 & 10.4 \\
            & FastDup \citep{visuallayer2022fastdup}
            & INet
                & 2.9 & 4.7
                & 0.3 & 5.2
                & 3.3 & 6.0
                & 64.8 & 24.2
                & 31.7 & 22.5
                & 9.3 & 5.5 \\
                
        & \cellcolor{Light}\textsc{SelfClean}
        & \cellcolor{Light}INet
                & \cellcolor{Light}\textbf{97.1} & \cellcolor{Light}\textbf{80.0}
                & \cellcolor{Light}\textbf{96.8} & \cellcolor{Light}\textbf{80.0}
                & \cellcolor{Light}96.2 & \cellcolor{Light}81.6
                & \cellcolor{Light}97.1 & \cellcolor{Light}83.0
                & \cellcolor{Light}70.6 & \cellcolor{Light}24.5
                & \cellcolor{Light}77.3 & \cellcolor{Light}28.3 \\
        & \cellcolor{Light}\textsc{SelfClean}
        & \cellcolor{Light}SimCLR
                & \cellcolor{Light}73.3 & \cellcolor{Light}27.6
                & \cellcolor{Light}74.7 & \cellcolor{Light}31.6
                & \cellcolor{Light}91.8 & \cellcolor{Light}61.6
                & \cellcolor{Light}92.3 & \cellcolor{Light}65.2
                & \cellcolor{Light}68.3 & \cellcolor{Light}24.1
                & \cellcolor{Light}68.5 & \cellcolor{Light}22.1 \\
        & \cellcolor{Light}\textsc{SelfClean}
        & \cellcolor{Light}DINO
                & \cellcolor{Light}89.5 & \cellcolor{Light}57.6
                & \cellcolor{Light}89.0 & \cellcolor{Light}56.5
                & \cellcolor{Light}97.5 & \cellcolor{Light}84.1
                & \cellcolor{Light}\textbf{97.8} & \cellcolor{Light}\textbf{86.3}
                & \cellcolor{Light}\textbf{75.9} & \cellcolor{Light}\textbf{27.6}
                & \cellcolor{Light}\textbf{78.3} & \cellcolor{Light}\textbf{29.1} \\
        \bottomrule
    \end{tabular}
    }
\end{table*}
\addtolength{\tabcolsep}{+3pt}

\newpage
\subsection{Detailed comparison with metadata} \label{app:Detailed-Meta}
Table \ref{tab:Detailed-Meta-Full} details the comparison of the \textsc{SelfClean} ranking with metadata from multiple benchmark datasets, as discussed in section \ref{ssec:naturalresults}.

For PAD-UFES-20, we investigated \textsc{SelfClean}'s relatively low performance, as discussed in ``Comparision with metadata'' of section \ref{ssec:naturalresults}.
We provided the first 200 near-duplicate candidates of PAD-UFES-20 to three practicing dermatologists and asked them to verify whether the given samples were near duplicates.
The experts reached a good inter-annotator agreement with Krippendorff's alpha $> 0.6$.
Of the samples they unanimously agreed to be near duplicates (56 samples), 32\% had faulty metadata where the lesion ID was not correctly maintained.
Thus, we find evidence that the poor alignment of \textsc{SelfClean} and the metadata of PAD-UFES-20 is likely caused by imperfect metadata.

\begin{table}[htbp]
    \centering
    \caption{
        Comparison of \textsc{SelfClean} and competitor rankings with metadata from multiple benchmark datasets.
        We include the proportion of positive samples, which corresponds to the baseline AP.
        Consult section \ref{ssec:naturalresults} for interpretation.
    }
    \label{tab:Detailed-Meta-Full}
    \resizebox{1.0\linewidth}{!}{%
    \begin{tabular}{l l c ll rr}
        \toprule
        \bfseries Dataset &
        \bfseries Metadata & 
        \bfseries $\text{Positive Samples (\%)}$ & 
        \bfseries Method &
        \bfseries Rep. &
        \bfseries $\text{AUROC (\%)}$ & 
        \bfseries $\text{AP (\%)}$ \\
        
        \midrule
        \multirow{3}{*}{PAD-UFES-20} & \multirow{3}{*}{Same Lesion} & \multirow{3}{*}{$0.06$} 
            & pHashing \cite{marr_theory_1997} & & 56.6 & 0.2 \\
        & & 
            & SSIM \cite{wang_image_2004} & & 63.7 & 0.3 \\
        & & 
            & \cellcolor{Light}\textsc{SelfClean} & \cellcolor{Light}DINO & \cellcolor{Light}\textbf{71.0} & \cellcolor{Light}\textbf{10.0} \\
        
        \midrule
        \multirow{3}{*}{HAM10000} & \multirow{3}{*}{Same Lesion} & \multirow{3}{*}{$0.01$} 
            & pHashing \cite{marr_theory_1997} & & n.a.\footnotemark[4] & n.a.\footnotemark[4] \\
        & & 
            & SSIM \cite{wang_image_2004} & & n.a.\footnotemark[4] & n.a.\footnotemark[4] \\
        & & 
            & \cellcolor{Light}\textsc{SelfClean} & \cellcolor{Light}DINO & \cellcolor{Light}\textbf{98.7} & \cellcolor{Light}\textbf{28.4} \\
        
        \midrule
        \multirow{3}{*}{ISIC-2019} & \multirow{3}{*}{Same Lesion} & \multirow{3}{*}{$0.01$} 
            & pHashing \cite{marr_theory_1997} & & n.a.\footnotemark[4] & n.a.\footnotemark[4] \\
        & & 
            & SSIM \cite{wang_image_2004} & & n.a.\footnotemark[4] & n.a.\footnotemark[4] \\
        & & 
            & \cellcolor{Light}\textsc{SelfClean} & \cellcolor{Light}DINO & \cellcolor{Light}\textbf{98.2} & \cellcolor{Light}\textbf{26.6} \\
        
        \midrule
        \multirow{3}{*}{CheXpert} & \multirow{3}{*}{Same Patient} & \multirow{3}{*}{$0.01$} 
            & pHashing \citep{marr_theory_1997} & & n.a.\footnotemark[4] & n.a.\footnotemark[4] \\
        & & 
            & SSIM \cite{wang_image_2004} & & n.a.\footnotemark[4] & n.a.\footnotemark[4] \\
        & & 
            & \cellcolor{Light}\textsc{SelfClean} & \cellcolor{Light}DINO & \cellcolor{Light}\textbf{70.5} & \cellcolor{Light}\textbf{7.5} \\
        
        \midrule
        \multirow{3}{*}{CelebA} & \multirow{3}{*}{Same Person} & \multirow{3}{*}{$0.02$} 
            & pHashing \citep{marr_theory_1997} & & n.a.\footnotemark[4] & n.a.\footnotemark[4] \\
        & & 
            & SSIM \citep{wang_image_2004} & & n.a.\footnotemark[4] & n.a.\footnotemark[4] \\
        & & 
            & \cellcolor{Light}\textsc{SelfClean} & \cellcolor{Light}DINO & \cellcolor{Light}\textbf{78.8} & \cellcolor{Light}\textbf{30.9} \\
        
        \midrule
        \multirow{3}{*}{ImageNet-1k} & \multirow{3}{*}{\makecell{Verified \\ Label Errors\footnotemark[2]}} & \multirow{3}{*}{$4.38$} 
            & CLearning \citep{northcutt_confident_2022} & INet & 46.6 & 4.3 \\
        & & 
            & FastDup \citep{visuallayer2022fastdup} & INet & 42.6 & 3.6\\
        & & 
            & \cellcolor{Light}\textsc{SelfClean} & \cellcolor{Light}DINO & \cellcolor{Light}\textbf{67.7} & \cellcolor{Light}\textbf{8.7} \\
        
        \midrule
        \multirow{3}{*}{Food-101N} & \multirow{3}{*}{\makecell{Verified \\ Label Errors\footnotemark[3]}} & \multirow{3}{*}{$18.51$} 
            & CLearning \citep{northcutt_confident_2022} & INet & 61.0 & 25.2 \\
        & & 
            & FastDup \cite{visuallayer2022fastdup} & INet & 72.1 & 30.7 \\
        & & 
            & \cellcolor{Light}\textsc{SelfClean} & \cellcolor{Light}DINO & \cellcolor{Light}\textbf{79.8} & \cellcolor{Light}\textbf{47.8} \\
        \midrule
        \midrule
        \multicolumn{2}{l}{\bfseries Subsampled results\footnotemark[4]} \\
        \midrule
        \multirow{3}{*}{HAM10000} & \multirow{3}{*}{Same Lesion} & \multirow{3}{*}{$0.01$} 
            & pHashing \cite{marr_theory_1997} & & 71.3 \ [69.9, 74.2] & 2.7 \ [2.6, 7.6] \\
        & & 
            & SSIM \cite{wang_image_2004} & & 67.3 \ [66.3, 72.4] & 7.7 \ [4.4, 7.8] \\
        & & 
            & \cellcolor{Light}\textsc{SelfClean} & \cellcolor{Light}DINO & \cellcolor{Light}\textbf{98.7 \ [97.9, 99.0]} & \cellcolor{Light}\textbf{30.0 \ [23.9, 34.3]} \\
        
        \midrule
        \multirow{3}{*}{ISIC-2019} & \multirow{3}{*}{Same Lesion} & \multirow{3}{*}{$0.01$} 
            & pHashing \cite{marr_theory_1997} & & 62.3 \ [58.5, 63.4] & 0.1 \ [$<0.1$, 2.1] \\
        & & 
            & SSIM \cite{wang_image_2004} & & 69.1 \ [66.3, 70.0] & 1.3 \ [0.3, 1.6] \\
        & & 
            & \cellcolor{Light}\textsc{SelfClean} & \cellcolor{Light}DINO & \cellcolor{Light}\textbf{98.9 \ [97.4, 98.9]} & \cellcolor{Light}\textbf{28.6 \ [26.6, 29.2]} \\
        
        \midrule
        \multirow{3}{*}{CheXpert} & \multirow{3}{*}{Same Patient} & \multirow{3}{*}{$0.01$} 
            & pHashing \citep{marr_theory_1997} & & 54.7 \ [53.8, 57.0] & 0.2 \ [0.1, 0.4] \\
        & & 
            & SSIM \cite{wang_image_2004} & & 65.7 \ [64.7, 66.1] & 0.2 \ [0.2, 0.3] \\
        & & 
            & \cellcolor{Light}\textsc{SelfClean} & \cellcolor{Light}DINO & \cellcolor{Light}\textbf{86.5 \ [85.5, 88.1]} & \cellcolor{Light}\textbf{1.9 \ [0.3, 2.3]} \\
        
        \midrule
        \multirow{3}{*}{CelebA} & \multirow{3}{*}{Same Person} & \multirow{3}{*}{$0.02$} 
            & pHashing \citep{marr_theory_1997} & & 53.3 \ [52.8, 54.7] & $<$ 0.1 \ [$<$ 0.1, $<$ 0.1] \\
        & & 
            & SSIM \citep{wang_image_2004} & & 56.3 \ [55.3, 58.3] & $<$ 0.1 \ [$<$ 0.1, $<$ 0.1] \\
        & & 
            & \cellcolor{Light}\textsc{SelfClean} & \cellcolor{Light}DINO & \cellcolor{Light}\textbf{81.0 \ [80.6, 81.2]} & \cellcolor{Light}\textbf{0.6 \ [0.6, 0.6]} \\
        \bottomrule
    \end{tabular}
    }
\end{table}

\footnotetext[2]{Refers to the subset of ImageNet-1k validation set which was verified by \citet{northcutt_pervasive_2021}.}
\footnotetext[3]{Refers to the subset of Food-101N set which was verified by \citet{lee_cleannet_2018}.}
\footnotetext[4]{
    As the number of near duplicates for comparison exceeds memory limitations for the baseline methods (as indicated by ``n.a.'' in the upper panel), they were subsampled three times with the same percentage of positive samples to 2,000 samples (i.e., 1,999,000 comparisons). We report the median and the min-max variation in brackets.
}

\newpage
\subsection{Potential bias of off-topic ranking}
\label{app:minority}

There is a chance that off-topic sample detection may exacerbate data distribution biases because underrepresented samples are more likely to be proposed as candidate issues.
Therefore, we investigate if some specific dataset attributes correlate with the off-topic sample ranking, assessing, for example, if pigment-rich skin is more often suggested to be off-topic.
For this experiment, we focus on the demographics of CheXpert and skin types in DDI and Fitzpatrick17k.
We compare the ranking of the feature attribute using \gls*{ap} and AUROC, similar to the comparison with metadata in appendix \ref{app:Detailed-Meta}.
The results show no evidence of an increased likelihood of underrepresented groups appearing earlier in the ranking, as AUROC stays around 50\% and \gls*{ap} is similar to the non-informed baseline, i.e., the percentage of samples belonging to the group.

\begin{table}[htbp]
    \centering
    \caption{
        Comparison of the \textsc{SelfClean} ranking with various demographic attributes.
        For reference, we include the prevalence of each group, also corresponding to the not-informed baseline performing best in terms of AP.
    }
    \resizebox{1.0\linewidth}{!}{%
    \begin{tabular}{lll rrr}
        \toprule
        \bfseries Dataset &
        \bfseries Attribute &
        \bfseries Value & 
        \bfseries $\text{Prevalence (\%)}$ & 
        \bfseries $\text{AUROC (\%)}$ & 
        \bfseries $\text{AP (\%)}$ \\
        \midrule
        \multirow{3}{*}{DDI} & \multirow{3}{*}{Skin Tone} 
            & Fitzpatrick Type 3\&4 & 36.7 & 46.8 & 35.4 \\
         &  & Fitzpatrick Type 1\&2 & 31.7 & 52.5 & 31.2 \\
         &  & Fitzpatrick Type 5\&6 & 31.6 & 50.9 & 35.9 \\
        \midrule
        \multirow{7}{*}{Fitzpatrick17k} & \multirow{7}{*}{Skin Tone} 
            & Fitzpatrick Type 2 & 29.0 & 53.2 & 31.1 \\
         &  & Fitzpatrick Type 3 & 20.0 & 47.5 & 19.1 \\
         &  & Fitzpatrick Type 1 & 17.8 & 52.8 & 18.9 \\
         &  & Fitzpatrick Type 4 & 16.8 & 45.4 & 15.2 \\
         &  & Fitzpatrick Type 5 & 9.2 & 46.3 & 8.5 \\
         &  & Fitzpatrick Type 6 & 3.8 & 50.8 & 3.8 \\
         &  & Fitzpatrick Type Unknown & 3.4 & 57.5 & 4.3 \\
        \midrule
        \multirow{6}{*}{CheXpert} & \multirow{6}{*}{Ethnicity} 
            & Non-Hispanic/Non-Latino & 72.9 & 50.0 & 72.8 \\
         &  & Unknown & 14.2 & 53.3 & 15.4 \\
         &  & Hispanic/Latino & 12.1 & 46.5 & 11.1 \\
         &  & Patient Refused & 0.3 & 43.5 & 0.2 \\
         &  & Not Hispanic & $<$ 0.1 & 35.1 & $<$ 0.1 \\
         &  & Hispanic & $<$ 0.1 & 9.1 & $<$ 0.1 \\
         \midrule
         \multirow{3}{*}{CheXpert} & \multirow{3}{*}{Gender} 
            & Male & 55.2 & 43.5 & 51.4 \\
         &  & Female & 44.3 & 56.4 & 48.5 \\
         &  & Unknown & $<$ 0.1 & 17.7 & $<$ 0.1 \\
        \midrule
        \multirow{23}{*}{CheXpert} & \multirow{23}{*}{Primary Race} 
        & White & 45.5 & 47.7 & 44.0 \\
        &  & Other & 12.9 & 46.4 & 11.9 \\
        &  & White, non-Hispanic & 10.0 & 55.7 & 11.5 \\
        &  & Asian & 9.5 & 51.7 & 9.7 \\
        &  & Unknown & 6.6 & 52.5 & 7.1 \\
        &  & Black or African American & 4.0 & 47.4 & 3.8 \\
        &  & Race and Ethnicity Unknown & 3.9 & 53.8 & 4.3 \\
        &  & Other, Hispanic & 1.7 & 49.6 & 1.6 \\
        &  & Asian, non-Hispanic & 1.2 & 56.0 & 1.4 \\
        &  & Native Hawaiian or Other Pacific Islander & 1.2 & 44.1 & 1.0 \\
        &  & Black, non-Hispanic & 0.8 & 55.5 & 1.0 \\
        &  & White, Hispanic & 0.5 & 53.7 & 0.6 \\
        &  & Other, non-Hispanic & 0.3 & 56.4 & 0.4 \\
        &  & Patient Refused & 0.2 & 44.4 & 0.2 \\
        &  & American Indian or Alaska Native & 0.2 & 46.5 & 0.2 \\
        &  & Pacific Islander, non-Hispanic & 0.1 & 51.4 & 0.2 \\
        &  & Native American, non-Hispanic & $<$ 0.1 & 60.7 & $<$ 0.1 \\
        &  & Black, Hispanic & $<$ 0.1 & 60.7 & $<$ 0.1 \\
        &  & Native American, Hispanic & $<$ 0.1 & 60.2 & $<$ 0.1 \\
        &  & Asian, Hispanic & $<$ 0.1 & 63.4 & $<$ 0.1 \\
        &  & White or Caucasian & $<$ 0.1 & 18.9 & $<$ 0.1 \\
        &  & Pacific Islander, Hispanic & $<$ 0.1 & 65.1 & 0.2 \\
        &  & Asian - Historical Conv & $<$ 0.1 & 58.7 & $<$ 0.1 \\
        \bottomrule
    \end{tabular}
    }
\end{table}

\newpage
\subsection{Detailed influence of dataset cleaning}
\label{app:Detailed-DatasetCleaning}

Table \ref{tab:DataCleaning-Detail} complements table \ref{tab:cleaninginfluencemain} with score ranges before differences.
Conclusions are drawn in section \ref{sec:discussion}.

\begin{table*}[htbp]
    \centering
    \scriptsize
    \caption{
    Influence of removing samples detected in the automatic cleaning mode with $\alpha = 0.10$ and $q = 0.05$ on downstream tasks.
    We report macro-averaged F1 scores for linear and $k$NN classifiers on \gls*{dino} features
    over 100 random training/evaluation splits with 80\% and 20\% fractions respectively. 
    We compute paired performance differences before and after cleaning the evaluation set,
    and before and after cleaning also the training set.
    We report the median and the intervals to the 5\% (subscript) and 95\% (superscript) percentiles.
    Additionally, we indicate significance of a paired permutation test on the difference sign with $^{*}p<0.05$, $^{**}p<0.01$, and $^{***}p<0.001$.
    }
    \label{tab:DataCleaning-Detail}
    \begin{tabular}{l  lll ll}
        \toprule
        & \multicolumn{5}{c}{\bfseries $k$NN Classifier} \\
        \cmidrule(lr){2-6}

        & \multicolumn{3}{c}{\bfseries Scores (\%)} 
        & \multicolumn{2}{c}{\bfseries Differences (\%pt.)} \\
        \cmidrule(lr){2-4}
        \cmidrule(lr){5-6}
        
        & Cont + Cont & Cont + Clean & Clean + Clean 
        & \multicolumn{1}{c}{Clean Eval} 
        & \multicolumn{1}{c}{Clean Train} \\
        \midrule
        DDI
            & $58.2^{+7.7}_{-8.3}$ & $59.2^{+7.5}_{-8.3}$ & $59.7^{+7.3}_{-8.8}$
            & $+1.2^{+1.9}_{-1.2} \ ^{***}$ 
            & $+0.0^{+1.7}_{-1.4} \ ^{***}$ \\
        HAM10000
            & $58.3^{+3.4}_{-4.9}$ & $58.3^{+3.7}_{-4.7}$ & $58.7^{+3.1}_{-4.6}$
            & $+0.2^{+0.5}_{-0.4} \ ^{***}$ 
            & $+0.2^{+1.3}_{-0.8} \ ^{**}$ \\
        Fitzpatrick17k
            & $60.2^{+1.8}_{-1.9}$ & $56.1^{+1.9}_{-2.2}$ & $56.1^{+2.0}_{-2.3}$
            & $-4.1^{+1.2}_{-1.3} \ ^{***}$ 
            & $+0.1^{+2.0}_{-1.7}$ \\
        \midrule
        Food-101N
            & $40.3^{+0.8}_{-0.9}$ & $40.4^{+0.7}_{-1.1}$ & $40.5^{+0.7}_{-1.1}$
            & $+0.1^{+0.1}_{-0.1} \ ^{***}$
            & $+0.1^{+0.2}_{-0.2} \ ^{***}$\\
        ImageNet-1k
            & $31.2^{+0.8}_{-0.9}$ & $30.8^{+0.9}_{-0.9}$ & $31.1^{+0.8}_{-0.9}$
            & $-0.4^{+0.1}_{-0.2} \ ^{***}$ 
            & $+0.4^{+0.3}_{-0.4} \ ^{***}$ \\
        \midrule
        & \multicolumn{5}{c}{\bfseries Linear Classifier} \\
        \cmidrule(lr){2-6}
        
        
        \bfseries Dataset &
        Cont + Cont & Cont + Clean & Clean + Clean 
        & \multicolumn{1}{c}{Clean Eval} 
        & \multicolumn{1}{c}{Clean Train} \\
        \midrule
        DDI
            & $59.2^{+9.6}_{-10.2}$ & $59.6^{+12.0}_{-11.2}$ & $58.9^{+9.0}_{-9.7}$
            & $+1.0^{+11.1}_{-11.2}$ 
            & $-0.7^{+7.7}_{-10.8}$ \\
        HAM10000
            & $62.6^{+4.2}_{-4.2}$ & $63.0^{+3.3}_{-4.0}$ & $62.8^{+3.2}_{-3.8}$
            & $+0.1^{+3.2}_{-3.5}$
            & $-0.1^{+3.9}_{-3.6}$\\
        Fitzpatrick17k
            & $52.8^{+2.6}_{-3.1}$ & $52.5^{+2.5}_{-4.1}$ & $52.6^{+2.9}_{-2.8}$
            & $-0.6^{+2.9}_{-3.6} \ ^{**}$ 
            & $+0.2^{+3.3}_{-3.9} \ ^{*}$ \\
        \midrule
        Food-101N
            & $50.0^{+0.9}_{-1.2}$ & $50.1^{+1.1}_{-1.0}$ & $50.4^{+0.8}_{-1.2}$
            & $+0.2^{+0.6}_{-0.5} \ ^{***}$ 
            & $+0.1^{+0.6}_{-0.5} \ ^{**}$ \\
        ImageNet-1k
            & $42.4^{+0.7}_{-0.9}$ & $42.0^{+0.9}_{-0.9}$ & $42.2^{+0.6}_{-1.0}$
            & $-0.4^{+0.6}_{-0.6} \ ^{***}$
            & $-0.0^{+0.9}_{-0.5}$ \\
        
        
        \bottomrule
    \end{tabular}
\end{table*}

\subsection{Investigation of VinDr-BodyPartXR near duplicates}
\label{app:Investigation-VDR}
\begin{figure}[htbp]
    \centering
    \begin{subfigure}[b]{0.4\textwidth}
        \centering
        \includegraphics[width=\linewidth]{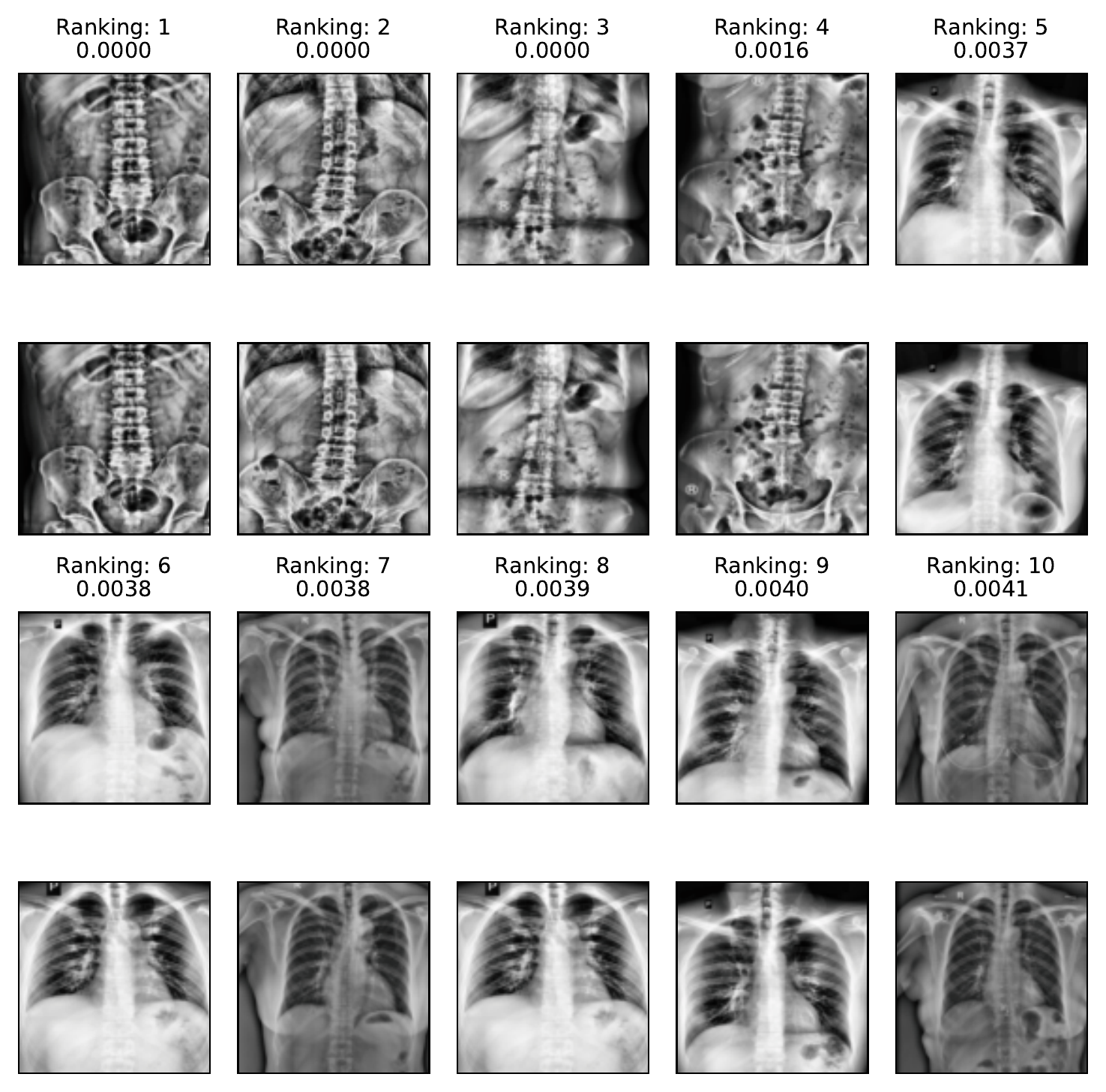}
        \caption{
            Ranking produced by \textsc{SelfClean} for near duplicates in the VinDr-BodyPartXR, of which the top-10 are shown along with the respective rank and score.
        }
        \label{fig:VDR-Duplicates}
    \end{subfigure}
    \hspace{5mm}
    \begin{subfigure}[b]{0.4\textwidth}
        \centering
        \includegraphics[width=\textwidth]{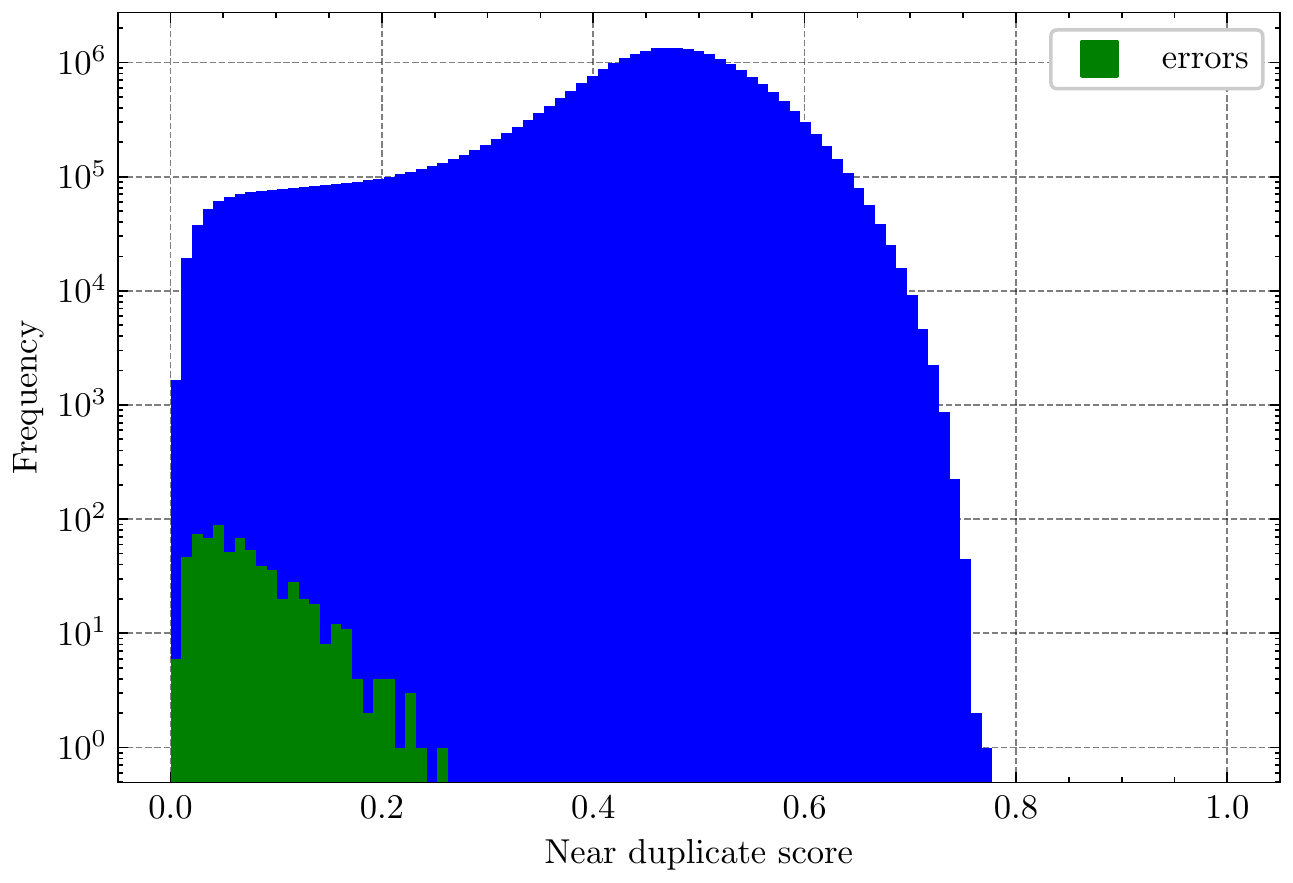}
        \caption{
            Histogram of the scores for VinDr-BodyPartXR with injected near duplicates using AUG.
            The green distribution shows synthetic issues, and blue is the overall score distribution.
        }
        \label{fig:VDR-Scores}
    \end{subfigure}
    \caption{Investigation of VinDr-BodyPartXR near duplicates.}
\end{figure}

In the synthetic experiments (see \ref{ssec:syntheticresults}) we observe particularly low \gls*{ap} for synthetic near duplicates with AUG for VinDr-BodyPartXR (VDR).
Here this discrepancy is further investigated.
Figure \ref{fig:VDR-Duplicates} illustrates the top-10 near duplicate candidates for VDR without synthetic contamination.
At least some of them are natural contamination that is not accounted for in the dataset's metadata, and others have highly standardized poses which may match more easily than synthetic contamination.
Figure \ref{fig:VDR-Scores} shows the score distribution of the injected duplicates in comparison to the overall distribution and illustrates that they lie in the earlier parts of the ranking.

\newpage
\section{Inspection effort saved}
\label{app:InspectionEffortSaved}

\looseness=-1
When potential data quality issues are verified by a human, it is valuable to quantify the reduction in inspection effort achieved through the ranking.
This reduction should be viewed as a function of the residual contamination that can be tolerated in the dataset, i.e., of the recall for data quality issues.
We quantify effort using the number of inspections required rather than the actual time spent, as this is a good proxy more directly related to the ranking.
Specifically, we calculate the \mbox{\gls*{fe}} needed to achieve a given recall by dividing the number of inspections required using the ranking by the number of inspections needed when candidate issues are sorted randomly.
The comparison baseline is random sorting, which always requires confirming a number of examples equal to the target recall times the number of potential issues, due to the uniform density of actual issues in the sequence.
The \acrlong*{fe} equals 1 when confirmation using the ranking is just as cumbersome as the baseline. 
It is $<$1 when the ranking is beneficial for cleaning, and $>$1 when it is detrimental.
The best and worst cases possible are obtained by a ranking algorithm that sorts all positive samples first or last respectively.
They obtain \glspl*{fe} equal to $\alpha_+$ and $[1-(1-\mathrm{R})\alpha_+]/\mathrm{R}$, where $\mathrm{R}$ is the recall and $\alpha_+$ is the contamination in the dataset, i.e., the number of actual data quality issues divided by the number of possible data quality issues.
Note that the \acrlong*{fe} saved by a method compared to another can easily be obtained by dividing the two corresponding \glspl*{fe}.

To summarize the inspection effort savings in a single number, we compute the \gls*{afe} over all possible recalls, i.e., the area under the \gls*{fe}--R curve.
To this end, we proceed as in the computation of average precision, and define

\begin{equation}
    \mathrm{AFE} = \sum_{i} (\mathrm{R}_{i+1} - \mathrm{R}_i)\mathrm{FE}_i.
\end{equation}

\begin{figure}[htbp]
  \centering
  \vspace*{-0.7em}
  \includegraphics[width=0.85\linewidth]{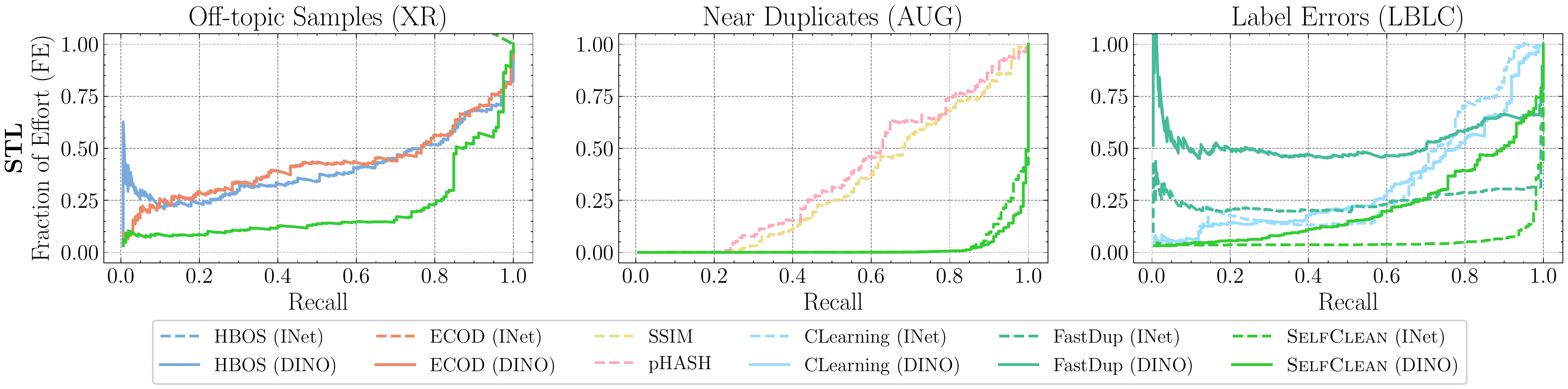}
  \caption{
    \looseness=-1
    \Gls*{fe}--R curves for a mixed-contamination strategy at 10\% level.
    The artificial dataset is created from STL by adding X-ray images (XR), injecting augmented duplicates (AUG), and changing labels at random (LBLC).
    The closer the curves are to zero, the less effort is needed to find data quality issues.
  }
  \label{fig:FE}
\end{figure}

\looseness=-1
In table \ref{tab:Results-AFE} we compare the best two competing approaches with \textsc{SelfClean} on a 10\% mixed-contamination dataset starting from STL in terms of \gls*{afe}.
In figure \ref{fig:FE} we further plot the \gls*{fe}--R curves for all approaches.
For competing methods which operate on extracted features, we compare performance using both supervised INet and self-supervised dataset-specific DINO features.
For both off-topic samples and near duplicate detection, the \gls*{afe} is significantly lower for \textsc{SelfClean} than for its competitors, indicating a large amount of time and effort saved in using it.
For label errors, \textsc{SelfClean} with self-supervised dataset-specific representation leads to a similar \gls*{afe} as competitors, which however may be aided by the similarity of ImageNet and STL in this specific case.

\begin{table*}[htbp]
    \centering
    \scriptsize
    \caption{
        \Acrfull*{afe} for the detection of synthetic data quality issues.
        Evaluation is performed on a 10\% mixed-contamination dataset starting from STL and creating off-topic samples (OT) using XR, near duplicates (ND) using AUG, and label errors (LE) using LBLC.
    }
    \label{tab:Results-AFE}
    \begin{tabular}{@{} l ll  rrr}%
        \toprule
        \parbox[t]{2mm}{\multirow{5}{*}{\rotatebox[origin=c]{90}{OT}}}
            & \bfseries Method
            & \bfseries Rep.
            & $\uparrow$ AUROC (\%) & $\uparrow$ AP (\%) & $\downarrow$ AFE (\%) \\
        \cmidrule{2-6}
            & HBOS \citep{goldstein_histogram-based_2012} & INet
                & 0.6 & 1.6 & 508.4 \\ 
            & ECOD \citep{li_ecod_2022} & INet 
                & 0.7 & 1.6 & 518.4 \\
            & \cellcolor{Light}\textsc{SelfClean} & \cellcolor{Light}INet 
                & \cellcolor{Light}0.7 & \cellcolor{Light}1.6 & \cellcolor{Light}472.8 \\
            & \cellcolor{Light}\textsc{SelfClean} & \cellcolor{Light}DINO
                & \cellcolor{Light}\textbf{86.9} & \cellcolor{Light}\textbf{24.4} & \cellcolor{Light}\textbf{20.2} \\ 
        \midrule
        \parbox[t]{2mm}{\multirow{5}{*}{\rotatebox[origin=c]{90}{ND}}}
            & \bfseries Method
            & \bfseries Rep.
            & $\uparrow$ AUROC (\%) & $\uparrow$ AP (\%) & $\downarrow$ AFE (\%) \\
        \cmidrule{2-6}
            & pHashing \citep{marr_theory_1997} &
                & 72.1 & 6.1 & 37.0 \\ 
            & SSIM \citep{wang_image_2004} &
                & 74.7 & 2.0 & 32.8 \\ 
            & \cellcolor{Light}\textsc{SelfClean} & \cellcolor{Light}INet
                & \cellcolor{Light}97.3 & \cellcolor{Light}26.1 & \cellcolor{Light}2.9 \\
            & \cellcolor{Light}\textsc{SelfClean} & \cellcolor{Light}DINO
                & \cellcolor{Light}\textbf{98.2} & \cellcolor{Light}\textbf{46.2} & \cellcolor{Light}\textbf{1.8} \\ 
        \midrule
        \parbox[t]{2mm}{\multirow{5}{*}{\rotatebox[origin=c]{90}{LE}}}
            & \bfseries Method
            & \bfseries Rep.
            & $\uparrow$ AUROC (\%) & $\uparrow$ AP (\%) & $\downarrow$ AFE (\%) \\
        \cmidrule{2-6}
            & CLearning \cite{northcutt_confident_2022} & INet
                & 76.6 & 6.9 & 33.2 \\ 
            & FastDup \cite{visuallayer2022fastdup} & INet
                & 88.1 & 0.4 & 24.3 \\ 
            & \cellcolor{Light}\textsc{SelfClean} & \cellcolor{Light}INet 
                & \cellcolor{Light}\textbf{98.3} & \cellcolor{Light}\textbf{63.4} & \cellcolor{Light}\textbf{5.2} \\ 
            & \cellcolor{Light}\textsc{SelfClean} & \cellcolor{Light}DINO 
                & \cellcolor{Light}85.3 & \cellcolor{Light}32.6 & \cellcolor{Light}21.5 \\ 
        \bottomrule
    \end{tabular}
\end{table*}

\newpage
\section{Validating algorithmic rankings with humans}
\label{app:Human-Verification}
\begin{figure}[htbp]
    \centering
    \includegraphics[width=0.7\linewidth]{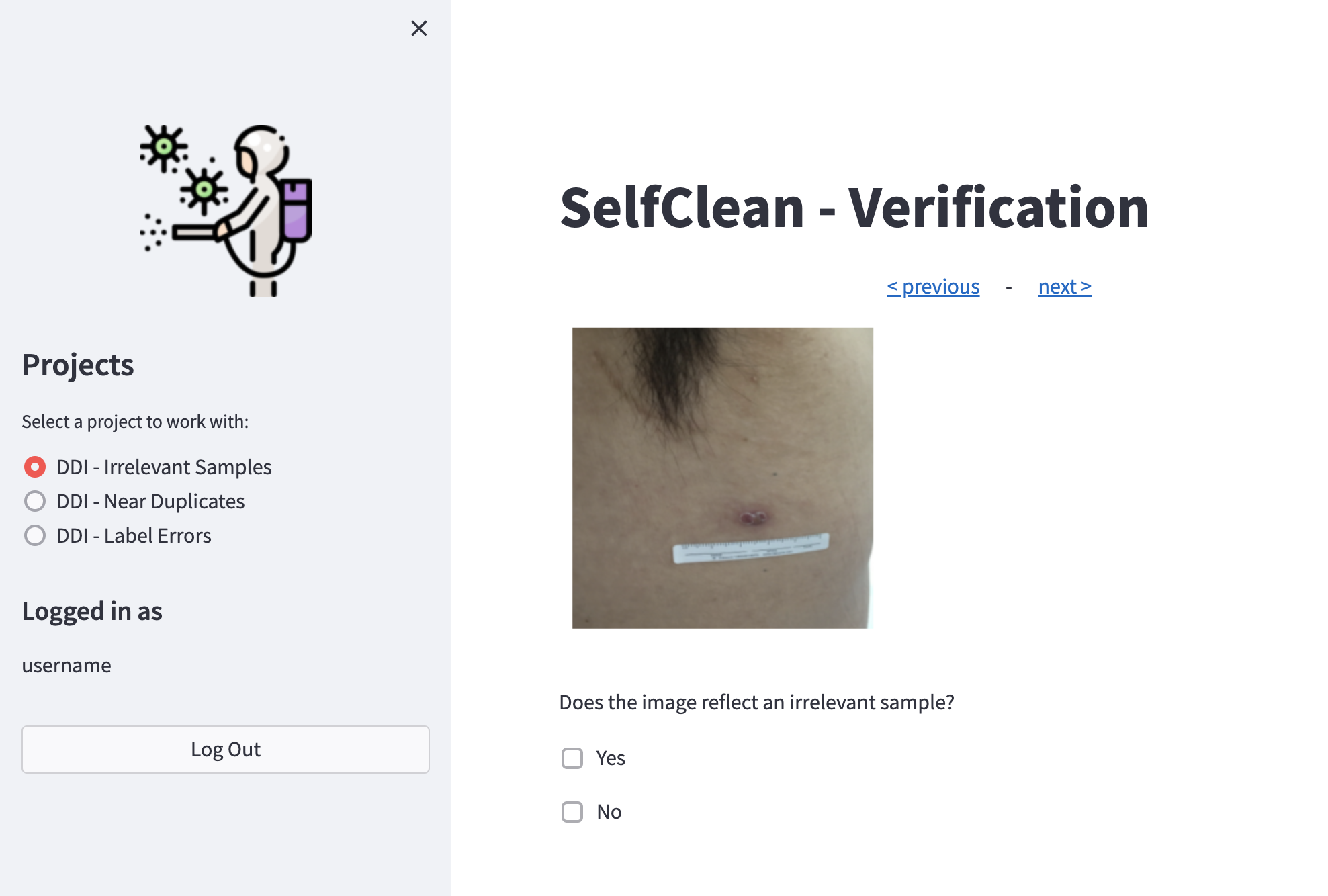}
    \caption{
        Screenshot of the verification tool used by medical experts to annotate data quality issues.
     }
    \label{fig:Verification-Tool}
\end{figure}

In this section, we describe the procedure used to confirm that, also according to human criteria, \textsc{SelfClean} assigns low ranks to problematic samples and high ranks to normal data, as discussed in the second part of section \ref{ssec:naturalresults}.
To this end, for each data quality issue type, we collect human verification for the first 50 images in the ranking and for 50 images randomly sampled from the dataset.
Crowd workers use the respective platform's tool\footnote[5]{\url{https://www.clickworker.com/}, accessed on the 28th of October.} for annotation, and medical expert annotators use a custom tool, which is shown in figure~\ref{fig:Verification-Tool}. 
The verification process starts with the selection of a dataset and data quality issue (e.g., the Fitzpatrick17k dataset and off-topic samples) and then proceeds with binary questions about single images or pairs thereof depending on the task.
Section \ref{app:Task-Descriptions} shows the task descriptions for each quality issue.
Note that the samples' ranks are not displayed to avoid potential bias.
Annotations were aggregated using majority voting for both crowd workers and medical experts.
Medical experts agreed with an average Krippendorff's alpha of 0.52, 0.97, and 0.55 for off-topic samples, near duplicates, and label errors, respectively.

We paid crowd workers 0.03 US dollars per annotation for images from ImageNet and Food-101N,
which roughly corresponds to 9 US dollars per hour.
Medical experts were not compensated financially but were instead acknowledged with co-authorship in a labeling consortium.

During annotation, we solely collected answers as binary labels along with anonymized annotator identification.
Thus, these annotations contain no personally identifiable information or offensive content.
In discussion with experts from the institutions of the co-authors, it was concluded that this verification process does not require IRB approval because the conducted study examines publicly available datasets and does not involve human subjects beyond binary annotations.

\subsection{Task descriptions}
\label{app:Task-Descriptions}

This section reports all task descriptions shown to the annotators:

\begin{itemize}
    \item Off-topic samples: 
        ``Your task is to judge if the image shown is irrelevant. 
        Select \textit{yes} when the image is not a valid input for the task at hand.''
        
    \item Near duplicates:
        ``Your task is to judge whether the two images shown together are pictures of the same object. 
        Note that pictures of the same object can be identical or different shots with the same object of interest.''
        
    \item Label errors:
        ``Your task is to judge whether the image's label is correct.
        Please select that the label is an error only if you think it is wrong and not when there is low uncertainty or ambiguity.''
\end{itemize}

\subsection{Detailed results}
\label{app:HumanValidation-Result}

\begin{figure}[htbp]
    \centering
    \begin{subfigure}[b]{0.24\textwidth}
        \centering
        \includegraphics[width=\textwidth]{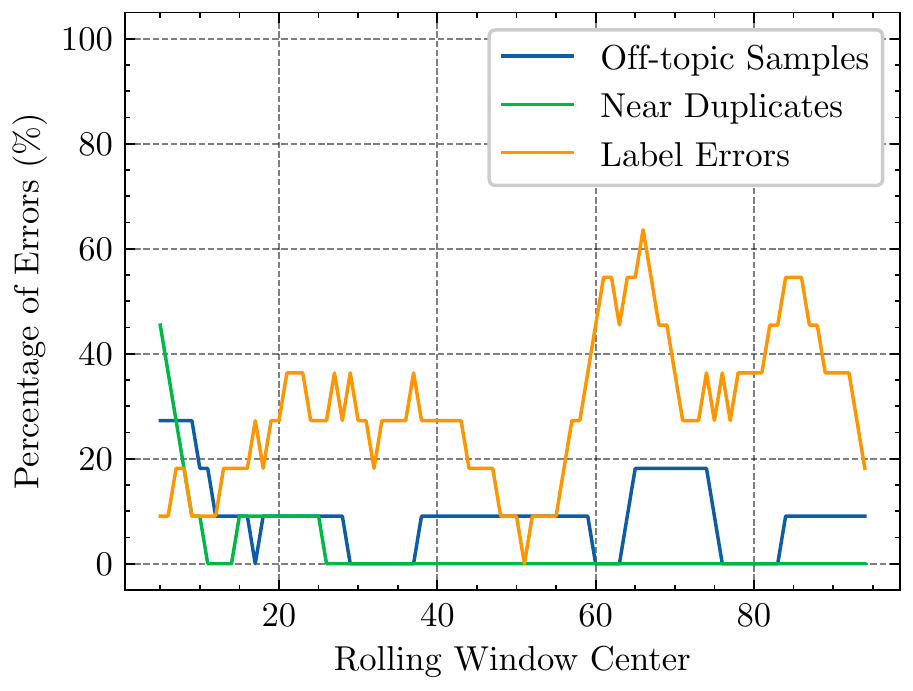}
        \caption{DDI}
    \end{subfigure}
    \begin{subfigure}[b]{0.24\textwidth}
        \centering
        \includegraphics[width=\textwidth]{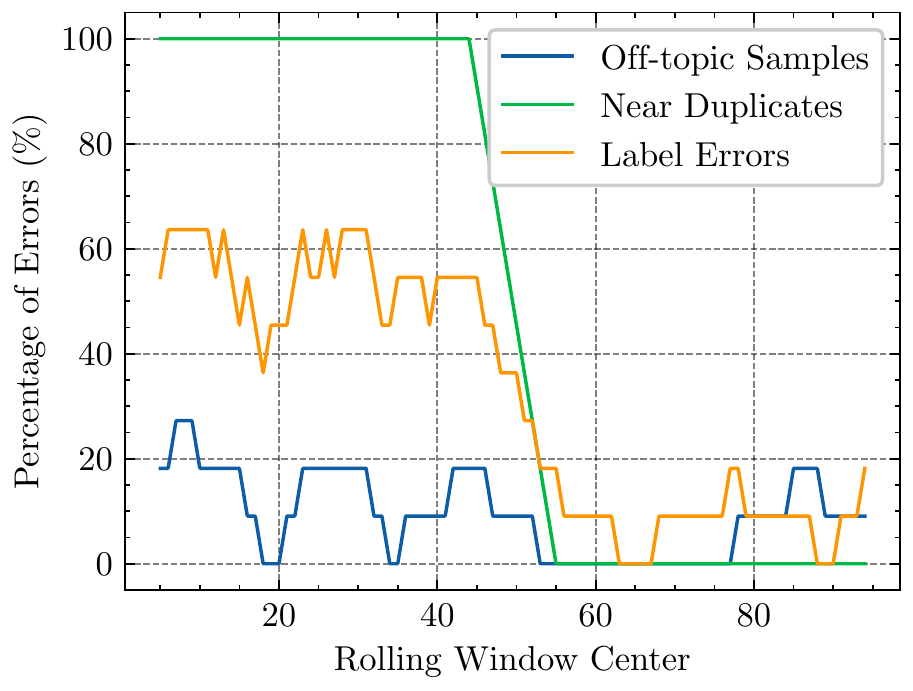}
        \caption{Fitzpatrick17k}
    \end{subfigure}
    \begin{subfigure}[b]{0.24\textwidth}
        \centering
        \includegraphics[width=\textwidth]{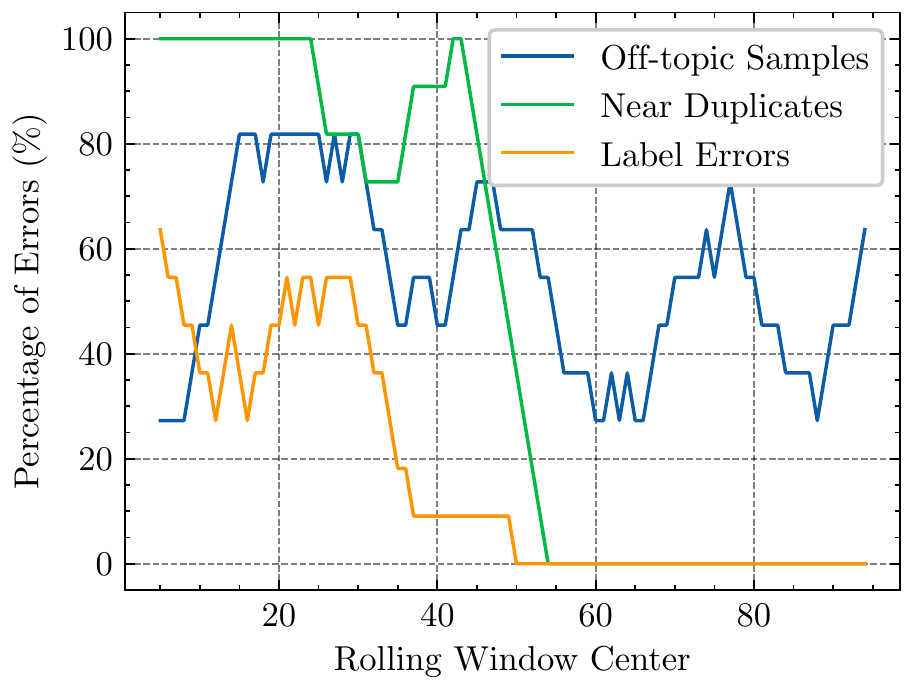}
        \caption{ImageNet-1k}
    \end{subfigure}
    \begin{subfigure}[b]{0.24\textwidth}
        \centering
        \includegraphics[width=\textwidth]{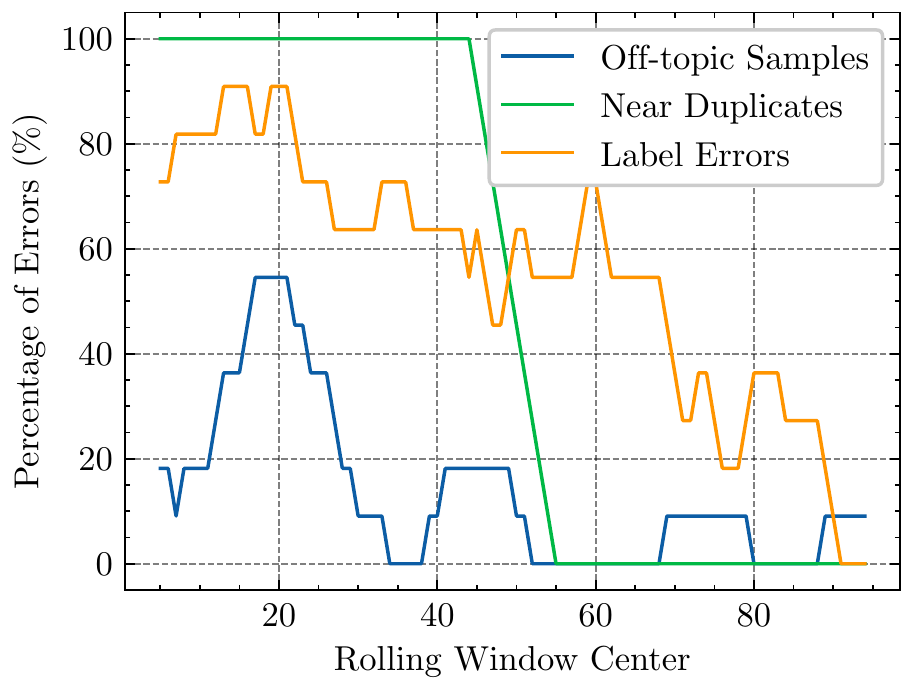}
        \caption{Food101N}
    \end{subfigure}
    \caption{
        Visualization of the percentage of quality issues found across the first 50 samples in the \textsc{SelfClean} ranking and in 50 random samples, using a rolling window of size 10.
        Results are reported across four datasets and for each issue type.
     }
    \label{fig:HumanValidation}
\end{figure}

\looseness=-1
In order to verify that problematic samples tend to appear first in the ranking provided by \textsc{SelfClean},
for each issue type, we first consider the first 50 images in the ranking against the 50 random ones, and then the first group of 25 in the ranking against the second group of 25.
We conduct one-sided Mann-Whitney $U$ statistical tests to verify that humans are more likely to identify data quality issues in samples that appear first in the \textsc{SelfClean} ranking.
In order to gain a more intuitive understanding, we also report the fraction of samples that were found to be problematic within the first 50 and the 50 random samples, and within samples ranked 1 through 25 and 26 through 50.
Finally, we visualize the distribution of human-confirmed problems through the ranking by plotting the fraction of confirmed problems in a rolling window of ten ranks in figure \ref{fig:HumanValidation}.

We observe significant alignment for near-duplicate detection throughout the considered datasets.
Label-error identification is significant in all cases but for DDI.
The different concentration of problems is mostly observed between images with low ranking and random samples, while the difference between samples 1-25 and 26-50 is less pronounced.
We observe that identifying label errors in a highly-curated dataset such as DDI is a nontrivial task which might exceed the design of the conducted experiment.
Finally, the detection of off-topic samples is the task where \textsc{SelfClean} achieves the lowest overall agreement with human annotators.
Nevertheless, these results suggest a significant separation of off-topic samples within the ranking in at least half of the cases.

\begin{table}[htbp]
    \centering
    \caption{
        Comparison of the percentage of issues found by humans in the 50 lowest-ranked samples with 50 random samples,
        and in samples 1 to 25 with samples 26 through 50.
        We report the percentage of issues in each sample and the corresponding $p$-value of a Mann–Whitney $U$ test,
        which represents the probability for the ranking to be unrelated to the position of problematic samples.
    }
    \label{app:HumanValidation-Table}
    \resizebox{\linewidth}{!}{%
        \begin{tabular}{l  l  ccc  ccc }
            \toprule
            & & \multicolumn{6}{c}{\bfseries Percentage of Human-Confirmed Problems} \\
            \cmidrule(lr){3-8}
            
            \bfseries Dataset &
            \bfseries Data Quality Issue &
            Lowest 1-50 (\%) & Random Sample (\%) & $p$-value &
            Lowest 1-25 (\%) & Lowest 26-50 (\%) & $p$-value \\
            \midrule
            DDI &
                Off-topic Samples &
                $12 $ & $8 $ & $0.25$ &
                $20 $ & $4 $ & $\mathbf{0.04}$ \\
            DDI &
                Near Duplicates &
                $12 $ & $0 $ & $\mathbf{0.006}$ &
                $24 $ & $0 $ & $\mathbf{0.005}$ \\
            DDI &
                Label Errors &
                $22 $ & $32 $ & $0.86$ &
                $20 $ & $24 $ & $0.63$ \\
            \midrule
            Fitzpatrick17k &
                Off-topic Samples &
                $14 $ & $4 $ & $\mathbf{0.04}$ &
                $12 $ & $16 $ & $0.65$ \\
            Fitzpatrick17k &
                Near Duplicates &
                $100 $ & $0 $ & $\mathbf{1.3\!\times\!10^{-23}}$ &
                $100 $ & $100 $ & $\texttt{undef}$ \\
            Fitzpatrick17k &
                Label Errors &
                $54 $ & $12 $ & $\mathbf{4.4\!\times\!10^{-6}}$ &
                $52 $ & $56 $ & $0.61$ \\
            \midrule
            ImageNet &
                Off-topic Samples &
                $62 $ & $48 $ & $0.08$ &
                $56 $ & $68 $ & $0.80$ \\
            ImageNet &
                Near Duplicates &
                $92 $ & $0 $ & $\mathbf{2.1\!\times\!10^{-20}}$ &
                $100 $ & $84 $ & $\mathbf{0.02}$ \\
            ImageNet &
                Label Errors &
                $36 $ & $0 $ & $\mathbf{1.6\!\times\!10^{-6}}$ &
                $48 $ & $24 $ & $\mathbf{0.04}$ \\
            \midrule
            Food-101N &
                Off-topic Samples &
                $24 $ & $4 $ & $\mathbf{0.002}$ &
                $36 $ & $12 $ & $\mathbf{0.02}$ \\
            Food-101N &
                Near Duplicates &
                $100 $ & $0 $ & $\mathbf{1.3\!\times\!10^{-23}}$ &
                $100 $ & $100 $ & $\texttt{undef}$ \\
            Food-101N &
                Label Errors &
                $72 $ & $34 $ & $\mathbf{7.6\!\times\!10^{-5}}$ &
                $80 $ & $64 $ & $0.61$ \\
            \bottomrule
        \end{tabular}
    }
\end{table}

\newpage
\section{Scoring for off-topic samples}
\label{app:irrelevantscoring}
This section describes how to construct a score based on hierarchical clustering, such that samples with a high probability of being off-topic have significantly lower values compared to the bulk of the data.
Note that, although in practice we use single-linkage agglomerative clustering, this heuristic construction can be applied to any distance-based hierarchical clustering and is formulated accordingly.

\textbf{Notation.}
Hierarchical clustering over a set of data points numbered $\{1,\ldots,N\}$ can be represented with a hierarchy of sets $\set{H}_n$ which specify clusters at each level $n$.
Let $n$ correspond to the number of clusters at a specific step $\set{H}_n = \{\set{C}_{1n}, \ldots, \set{C}_{nn}\}$,
where $\set{C}_{in}$ represents the $i$-th cluster at level $n$ in the hierarchy.
For instance, in agglomerative clustering, $n$ runs from $N$ to 1 as the algorithm proceeds and more data points are merged.
Without loss of generality, it is possible to reindex clusters such that indices of merged sets are always consecutive, and the other sets in $\set{H}_n$ do not change their relative order
\begin{equation}
    \set{C}_{in} =
    \begin{cases}
    \set{C}_{i(n+1)} &\text{if }i<i_n,\\
    \set{C}_{i(n+1)}\cup \set{C}_{(i+1)(n+1)} &\text{if }i=i_n,\\
    \set{C}_{(i+1)(n+1)} &\text{if }i>i_n,\\
    \end{cases}
    \qquad
    \text{for }\;i=1,\ldots,n\;\text{ and }\;n=1,\ldots,N,
\end{equation}
where from step $n+1$ to step $n$ clusters $i_n$ and $i_n+1$ are merged into cluster $i_n$.
The hierarchy of sets $\set{H}_n$ induces a dendrogram, i.e., a tree graph where each cluster is a node connected to its direct parent and children. 
Each element $n$ of the hierarchy (except for $\set{H}_N$, where every point is in a separate cluster) can also be associated with a distance $d_n$ which is the one at which the last two clusters were merged, $d_n = \dist(\set{C}_{i_n(n+1)}, \set{C}_{(i_n+1)(n+1)})$. 
To define a ranking, we sort the dendrogram such that at every step $|\set{C}_{i_n(n+1)}|\leq|\set{C}_{(i_n+1)(n+1)}|$, i.e., the cluster which contains the fewest leaves comes first, based on the idea that outliers are associated with merges containing fewer leaves \citep{jiang_two-phase_2001}.
In case of ties, the cluster created at the largest distance precedes the other. 

\begin{figure}[htbp]
  \centering
  \includegraphics[width=1.0\linewidth]{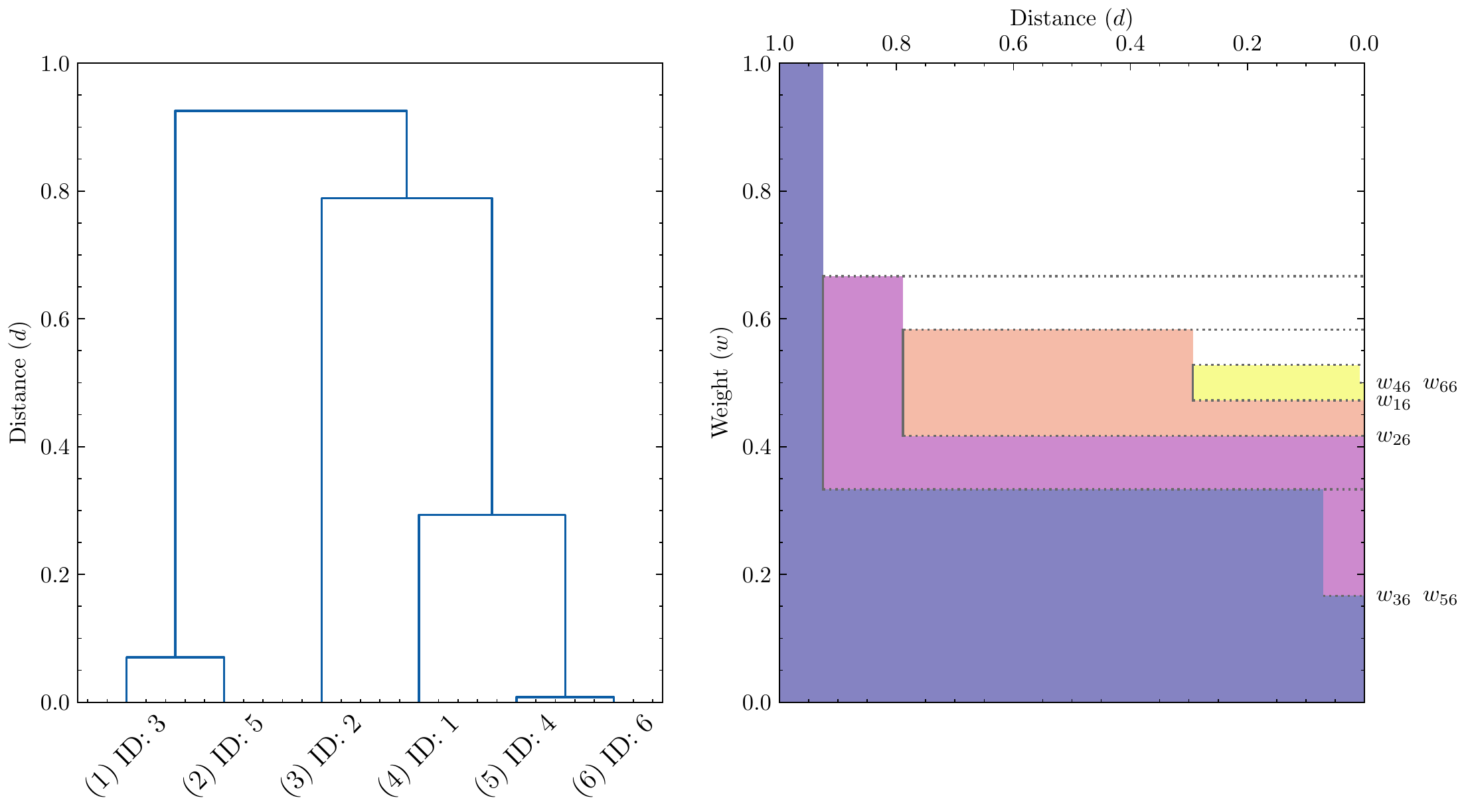}
  \caption{
     The left plot shows an example of a dendrogram for hierarchical clustering, and the right plot an illustration of the \acrfull*{LAD} scoring. 
     In the left plot, the x-axis shows the ranking of the different points in brackets and the corresponding identification number. 
     In the right plot, the right side of the y-axis shows the weights $w_{in}$ corresponding to equation \ref{eq:LAD}.
     }
  \label{fig:LAD-Illustration}
\end{figure}

\textbf{Scoring.}
To produce a score for each node in the dendrogram, natural building blocks are the merge distance, the sizes of the merged clusters, and their interactions \citep{tokuda_revisiting_2020}.
Accordingly, we define scores by drawing the dendrogram in a $[0,1]\!\times\![0,1]$ square where the horizontal axis is one minus the (merge) distance $d$ and the vertical axis is the weight $w_{in}$ of cluster $\set{C}_{in}$ which is defined recursively below.
Note that the possible values for the distance range from 0 to 1, which can generally be achieved with a transformation.
This guarantees that $1-d$ spans the same range.
This graphical construction is illustrated in the right panel of figure \ref{fig:LAD-Illustration}.
The score of each leaf is determined at each merge distance $d$ by the weight $w_{jn}$ of the cluster $\set{C}_{jn}$ it belongs to between merge distance $d_n$ and $d_{n-1}$. Formally, the off-topic sample score $s_{\text{OT}}(\vec{e}_i)$ is then given by the area under the curve $f_i(d)$ where
\begin{equation}
    f_i(d) = w_{jn}
    \quad\text{if}\quad d_{n} \le d < d_{n-1}
    \quad\text{and}\quad i\in \set{C}_{jn},
    \qquad n=1,\ldots,N,
\end{equation}
with $d_N = 0$ and $d_0 = 1$. 
For convenience, we define $p_{in} = |\set{C}_{in}| / N$ to be the probability of cluster $\set{C}_{in}$ and set $w_{0n}=0$ and $w_{11}=1$. 
To define the weights, we propose a rule which we call \gls*{LAD} and reads
\begin{equation} \label{eq:LAD}
w_{i(n+1)} =
\begin{cases}
    w_{in} &\text{if }i<i_n,\\
    w_{(i_n-1)n} + (w_{i_n n}-w_{(i_n-1)n})p_{i(n+1)}/p_{i_nn} &\text{if }i_n\leq i \leq i_n+1,\\
    w_{(i-1)n} &\text{if }i>i_n+1.
\end{cases}
\end{equation}
Essentially, at each split, children cluster $\set{C}_{i(n+1)}$ receives a weight $w_{i(n+1)}$ which is proportional to its relative size $p_{i(n+1)}/p_{i_nn}$ with respect to the parent cluster, while bound between the previous cluster weight $w_{(i_n-1)n}$ and the parent cluster weight $w_{i_nn}$.


\newpage
\section{Automatic cleaning}
\label{app:automaticcleaning}

In section \ref{sub:Dist-Indicators}, we defined scores that take extreme values for candidate issues.
Isolating data issues using such a score is essentially a one-dimensional outlier detection problem.
Here, we construct a procedure to detect outlier scores, which works well with \textsc{SelfClean}.
We then demonstrate that detected outliers are robust to the values of the two hyperparameters introduced by such procedure.

\subsection{Automatic cleaning procedure}

We start with the intuition that detecting problematic samples is straightforward if the scores are smoothly distributed for normal data, but are far from the bulk for data with quality issues.
However, all scores in this work range from 0 to 1, and increasingly extreme issues approach zero score without leaving large gaps on the score scale.
For this reason, we expand the neighborhoods of 0 and 1 using a logit transformation, $\tilde{s} = \log[s/(1-s)]$.
The transformed scores $\tilde{s}$ then range over the whole real axis enabling a better separation between normal and problematic samples.

Since the logit transformation has Jacobian $|\dd{\tilde{s}}/\dd{s}| = e^{-\tilde{s}}/(1+e^{-\tilde{s}})^2$, under broad assumptions we expect the dominant behavior of the transformed score distribution to drop at least as quickly as a logistic probability density function for $\tilde{s}\to\pm\infty$.
Note that this is the case even if the original score distribution is not just constant but presents an integrable power-law singularity for $s\to0, 1$.

\looseness=-1
To identify outlier samples, we first attempt to isolate a region on the left tail of the distribution that is free of issues.
To this end, we introduce a hyperparameter $\alpha$, the ``contamination rate guess'', which represents a generous estimate of the fraction of issues in the dataset.
For data quality issues where a score is associated to each sample, we simply drop the lowest $\lfloor\alpha_1 N\rfloor$ scores with $\alpha_1 = \alpha$, while when a score is associated to a pair of samples, we discard the lowest $\lfloor\alpha_1 N(N-1)/2\rfloor$ scores with $\alpha_1 = \alpha^2$.
Indeed, when there are no interactions (e.g., only pairs of near-duplicates) we expect $\alpha N$ abnormally low near-duplicate scores, but in the worst-case interaction scenario (e.g., all views of the same sample) we await $\alpha N(\alpha N-1)/2$ low out-of-distribution scores, which is equivalent to the above expression for $\alpha_1$ when $\alpha N \gg 1$.
Besides dropping the potentially problematic samples, we also select an upper score bound for the range of interest, since we aim at reproducing only the smooth \emph{left} tail of the distribution.
Reasonable choices are values between the lower score cutoff determined by $\alpha_1$ and the median, paying attention that enough data is included for sufficient robustness to noise.
For this reason, we choose the upper score cutoff to be the quantile corresponding to a fraction of data $\alpha_2$ which is the geometric mean between $\alpha_1$ and $1/2$, i.e., $\alpha_2^2 = \alpha_1/2$.
We observe that the range produces robust statistical information if the number of samples is sufficiently large and $\alpha\ll 1/2$, where in practice $\alpha\lesssim 1/4$ is already stable.

Following the outlined heuristic argument, we approximate the smooth component of the left tail of the distribution using a logistic distribution with suitably chosen scale and location parameters, which has probability density function
\begin{equation}
    \mathrm{pdf}(\tilde{s}; \mu, \sigma) =
    \frac{1}{\sigma}\mathrm{pdf}\Big(\frac{\tilde{s}-\mu}{\sigma}\Big),
    \qquad
    \mathrm{pdf}(\hat{s}) = \frac{e^{-\hat{s}}}{(1+e^{-\hat{s}})^2}.
\end{equation}
Given the score cutoffs $\bar{s}_1$ and $\bar{s}_2$ corresponding to the quantiles $\alpha_1$ and $\alpha_2$ of the empirically observed distribution, the scale $\sigma$ and location $\mu$ can be estimated as
\begin{equation}
    \sigma = \frac{\bar{s}_2 - \bar{s}_1}{\bar{s}(\alpha_2) - \bar{s}(\alpha_1)},
    \quad\;
    \mu = \frac{
        \bar{s}_1\bar{s}(\alpha_2) - \bar{s}_2\bar{s}(\alpha_1)
    }{
        \bar{s}(\alpha_2) - \bar{s}(\alpha_1)
    },
    \quad\;
    \bar{s}(\alpha_m) = \log\frac{\alpha_m}{1-\alpha_m}\quad\text{for}\; m=1,2.
\end{equation}
Here $\bar{s}(\alpha_m)$ indicates the percentage point function of the logistic distribution, i.e., the inverse of its cumulative distribution function.
Note that the whole estimation procedure for the left tail of the distribution relies exclusively on quantiles and is, therefore, naturally robust to outliers.

With an estimate of the smooth score distribution for normal data, we can identify abnormal samples by requesting that they be unlikely generated by the same random process.
This is achieved by demanding that the probability of obtaining a score below an outlier cutoff $s_\mathrm{cut}$ be less than a significance level $q$ times the expected fraction of outliers, which is $2\alpha/(N-1)$ in the case of pairs of samples and $\alpha$ otherwise.
We set the hyperparameter $q$ to $0.05$ corresponding to a 95\% one-sided confidence level and study the influence of this choice in section \ref{app:autocleanq}.
All samples with scores lower than the outlier cutoff will be then classified as problematic.

The strength of the aforementioned procedure lies in its ability to consistently detect outliers despite requiring multiple steps and introducing two additional hyperparameters. 
The number of outliers identified remains largely unaffected by reasonable choices for $\alpha$ and $q$.
The remaining parts of appendix \ref{app:automaticcleaning} are dedicated to showing that the procedure is intuitive and assumptions are empirically acceptable (\ref{app:autoexamples}), and to demonstrating that detected outliers exhibit low sensitivity to the contamination rate guess $\alpha$~(\ref{app:autocleanalpha}) and to the significance level $q$~(\ref{app:autocleanq}).

\subsection{Automatic cleaning examples}
\label{app:autoexamples}

\begin{figure}
    \centering
    \begin{subfigure}[b]{\textwidth}
        \centering
        \includegraphics[width=\textwidth]{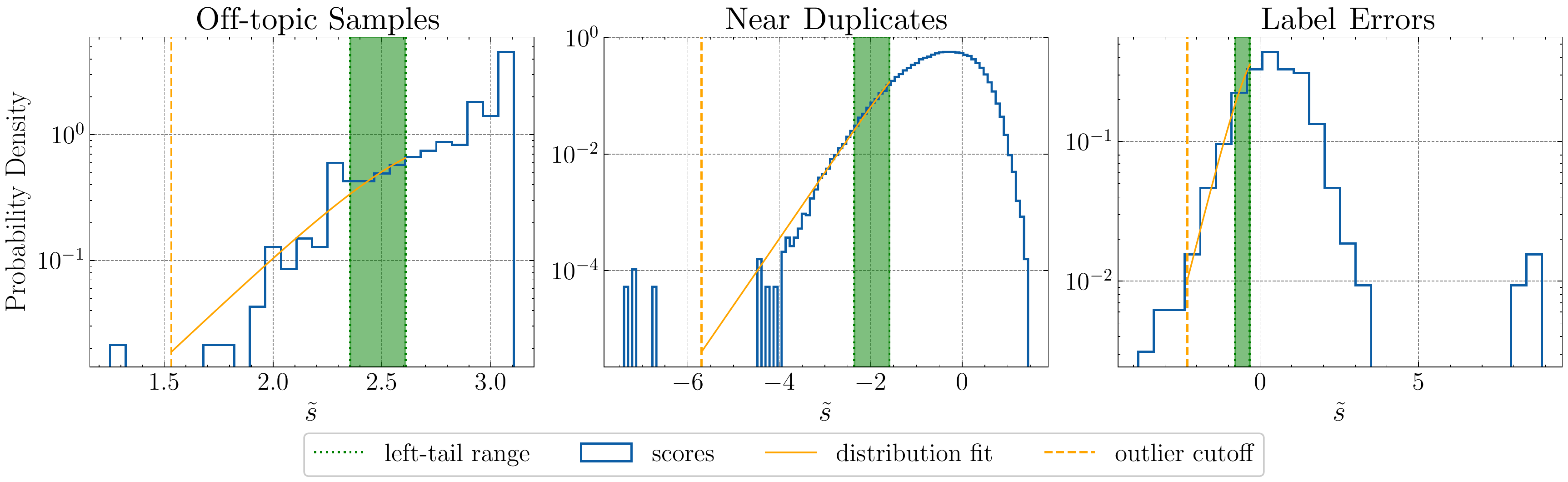}
        \caption{DDI \citep{daneshjou_disparities_2022}}
    \end{subfigure}
    \begin{subfigure}[b]{\textwidth}
        \centering
        \includegraphics[width=\textwidth]{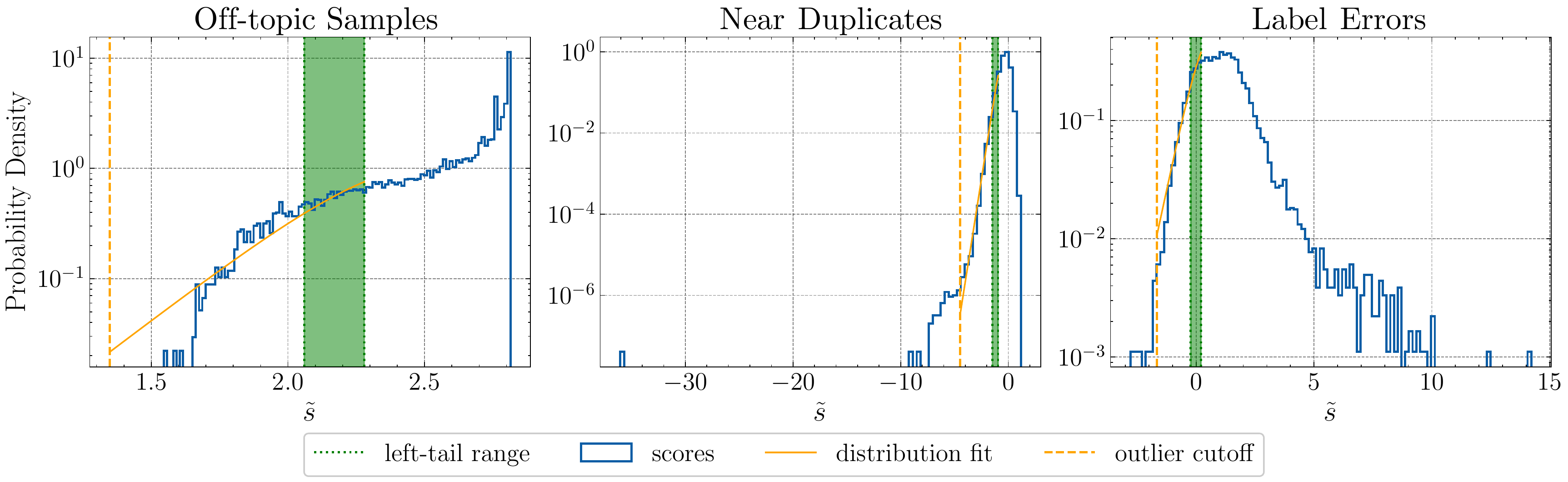}
        \caption{HAM10000 \citep{tschandl_ham10000_2018}}
    \end{subfigure}
    \begin{subfigure}[b]{\textwidth}
        \centering
        \includegraphics[width=\textwidth]{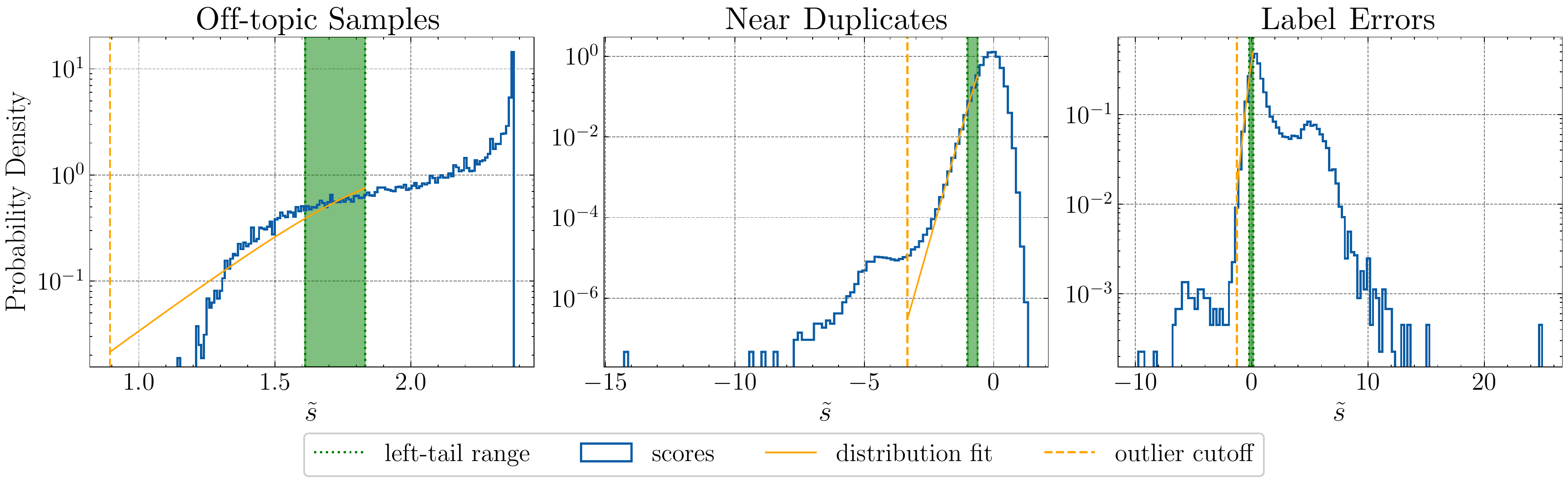}
        \caption{Fitzpatrick17k \citep{groh_evaluating_2021}}
    \end{subfigure}
    \begin{subfigure}[b]{\textwidth}
        \centering
        \includegraphics[width=\textwidth]{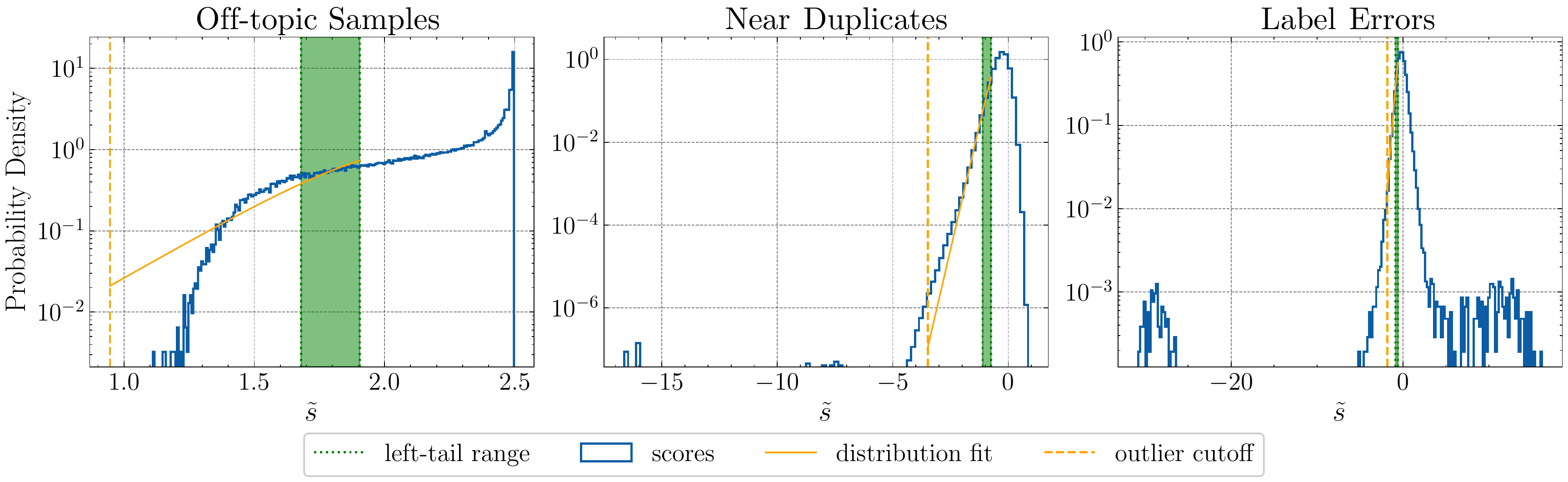}
        \caption{Food-101N \citep{lee_cleannet_2018}}
    \end{subfigure}
    \caption{
            Score histogram (blue) and associated left-tail distribution fit (solid orange)
            with outlier cutoff (dashed orange) for all considered issue types and representative datasets.
            The green shaded area represents the range $[\bar{s}_1, \bar{s}_2]$
            which is used to determine location and scale
            of the associated logistic distribution.
            The values $\alpha = 0.10$ and $q = 0.05$ are used throughout.
            }
    \label{fig:Distribution-Fits}
\end{figure}

In figure \ref{fig:Distribution-Fits}, we illustrate the fit to the left tails of distributions for representative datasets, together with the relevant range used to estimate scale and location and the position of the outlier cutoff to classify data quality issues.
We observe that the probability density function is a qualitatively good estimate of the density-normalized histograms in the expected range, i.e., for the smooth component of the histogram's left tail, within sampling uncertainties.
The fit quality is somewhat lower for off-topic samples, probably due to the score range which is all above $\tilde{s}=0$.
We also carried out experiments with a Gaussian functional form for score distribution tails and observed only minor changes, which resulted in a slightly reduced number of detected problems.

\subsection[Influence of the contamination rate guess]{Influence of the contamination rate guess $\alpha$}
\label{app:autocleanalpha}

\begin{figure}
    \centering
    \begin{subfigure}[b]{\textwidth}
        \centering
        \includegraphics[width=\textwidth]{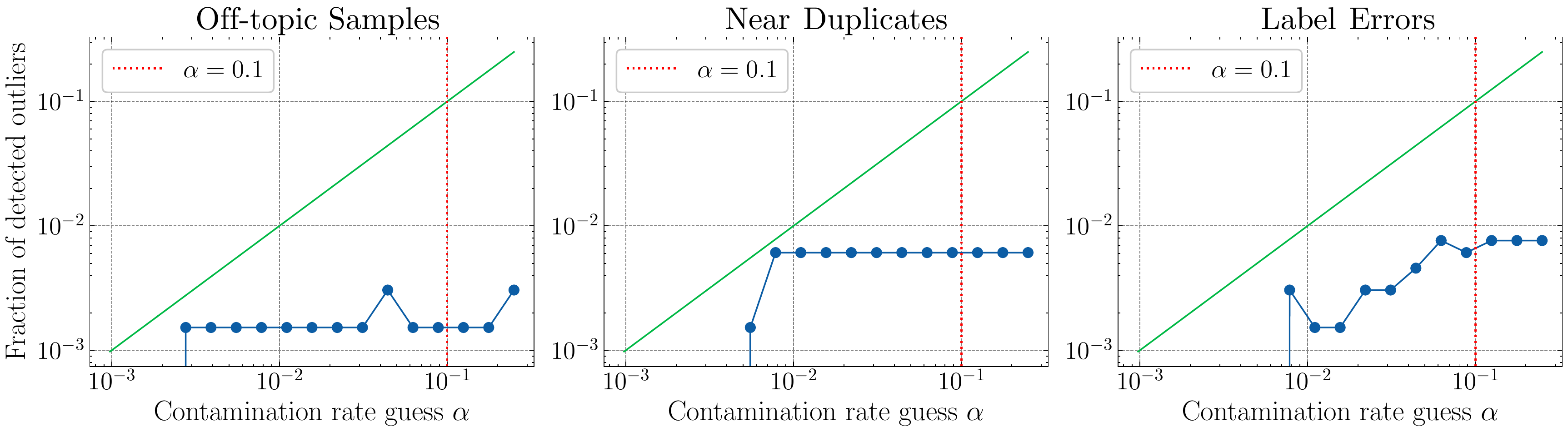}
        \caption{DDI \citep{daneshjou_disparities_2022}}
    \end{subfigure}
    \begin{subfigure}[b]{\textwidth}
        \centering
        \includegraphics[width=\textwidth]{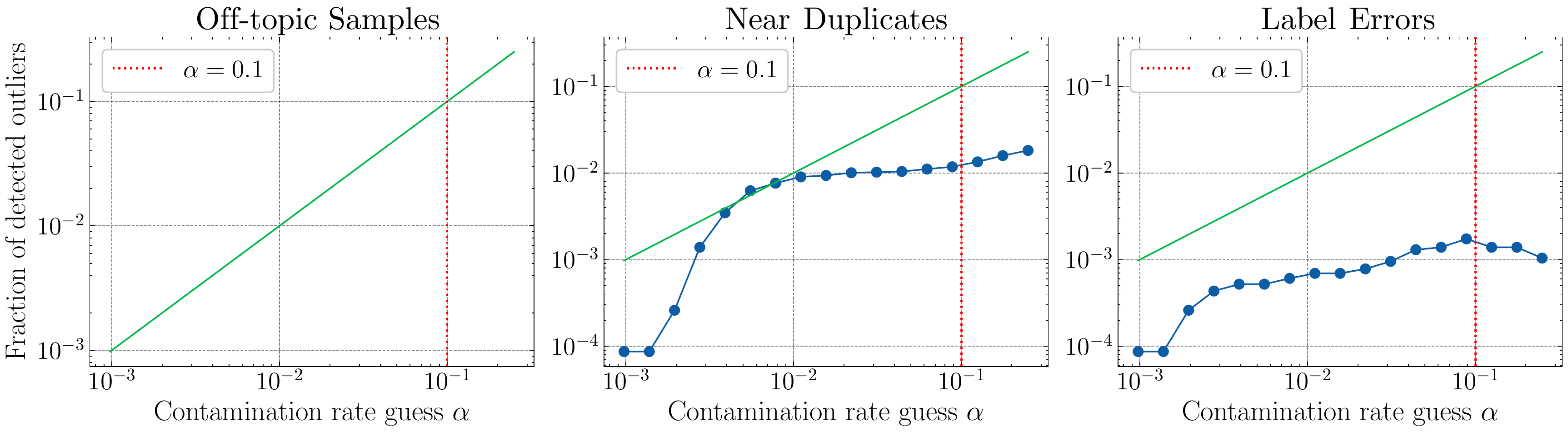}
        \caption{HAM10000 \citep{tschandl_ham10000_2018}}
    \end{subfigure}
    \begin{subfigure}[b]{\textwidth}
        \centering
        \includegraphics[width=\textwidth]{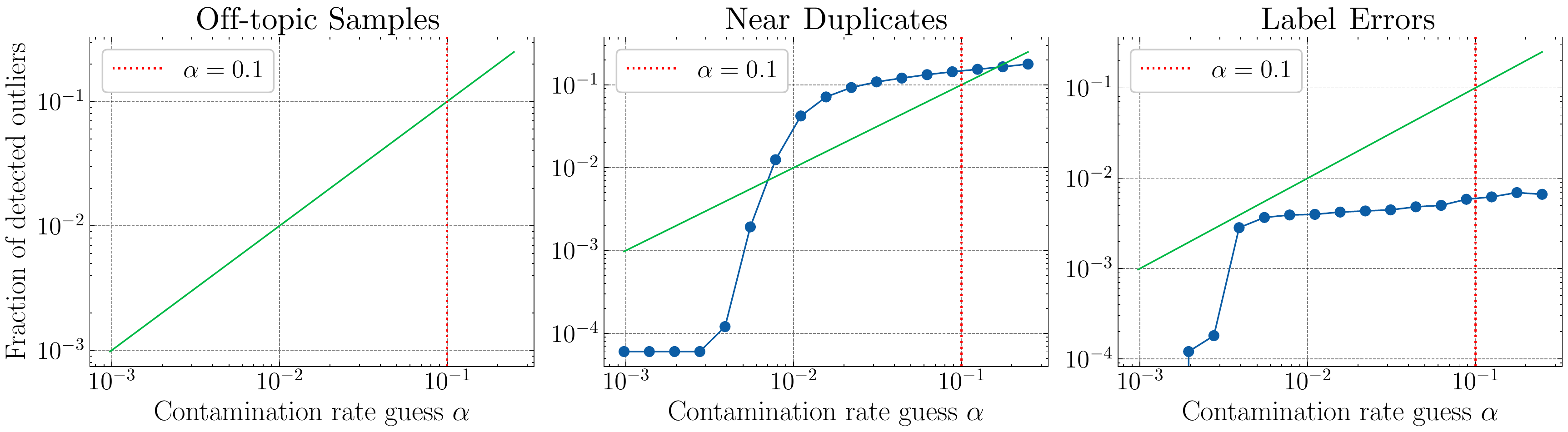}
        \caption{Fitzpatrick17k \citep{groh_evaluating_2021}}
    \end{subfigure}
    \begin{subfigure}[b]{\textwidth}
        \centering
        \includegraphics[width=\textwidth]{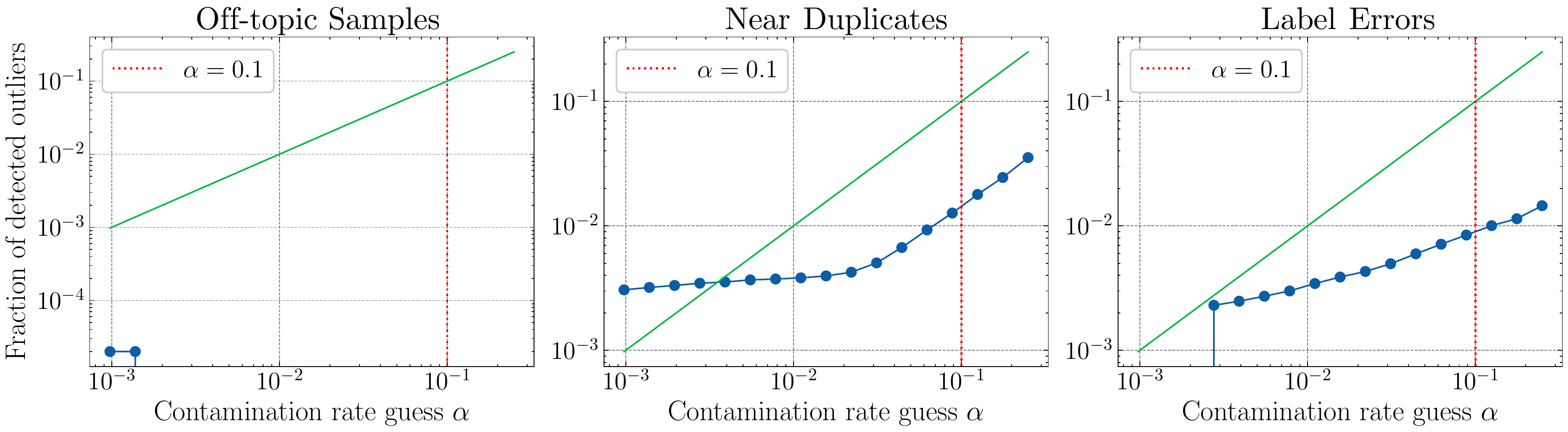}
        \caption{Food-101N \citep{lee_cleannet_2018}}
    \end{subfigure}
    \caption{
            Dependence of the fraction of detected data quality issues
            on the contamination rate guess~$\alpha$
            for all considered issue types and representative datasets,
            at a fixed significance level $q=0.05$.
            The observed behavior is reported in blue.
            It is outside of the lower margin of the plots
            when no problems are found.
            The green solid line represents a fraction of detected issues
            which is equal to the contamination rate guess for reference.
            The vertical dotted red line indicates the default value $\alpha=0.10$
            used in the rest of the paper.
            }
    \label{fig:AlphaSensitivity}
\end{figure}

In figure \ref{fig:AlphaSensitivity}, we analyze the sensitivity of the number of detected data quality problems with respect to the contamination rate guess $\alpha$ for all issue types and representative datasets analyzed in this paper.
In these plots, the significance level $q$ is fixed to its default value of $0.05$.
We observe that the fraction of found problems does not depend much on $\alpha$ over several orders of magnitude, suggesting a sensitivity to this hyperparameter that is approximately vanishing or at most logarithmic.
It is by virtue of this reduced dependence that we can fix $\alpha=0.10$ throughout the paper and that fully automatic cleaning is able to produce stable results with limited prior knowledge of dataset quality.

\subsection[Influence of the significance level]{Influence of the significance level $q$}
\label{app:autocleanq}

\begin{figure}
    \centering
    \begin{subfigure}[b]{\textwidth}
        \centering
        \includegraphics[width=\textwidth]{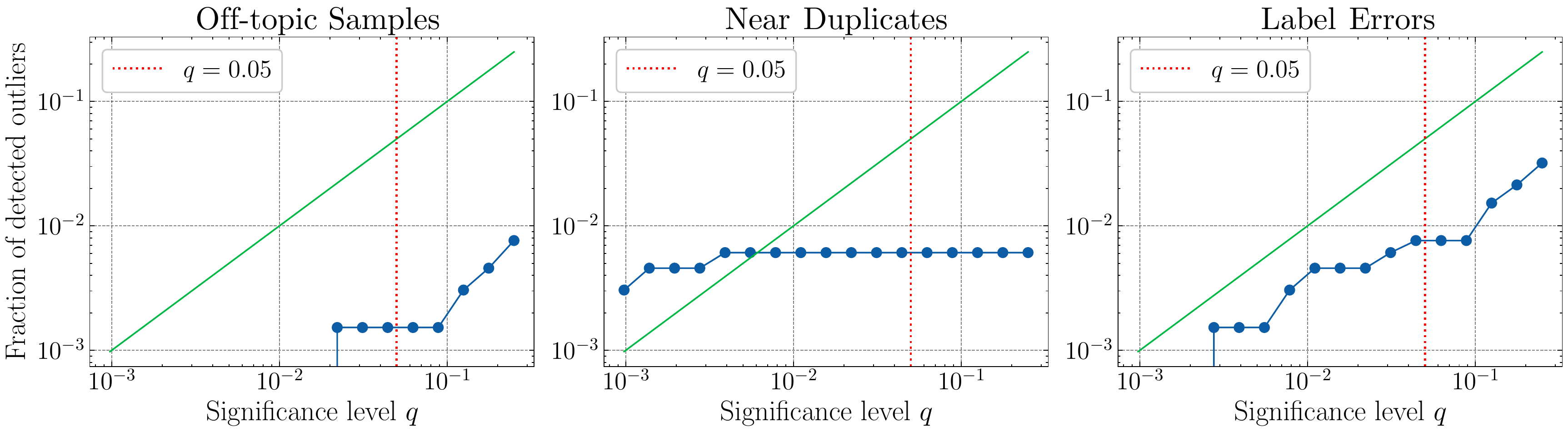}
        \caption{DDI \citep{daneshjou_disparities_2022}}
    \end{subfigure}
    \begin{subfigure}[b]{\textwidth}
        \centering
        \includegraphics[width=\textwidth]{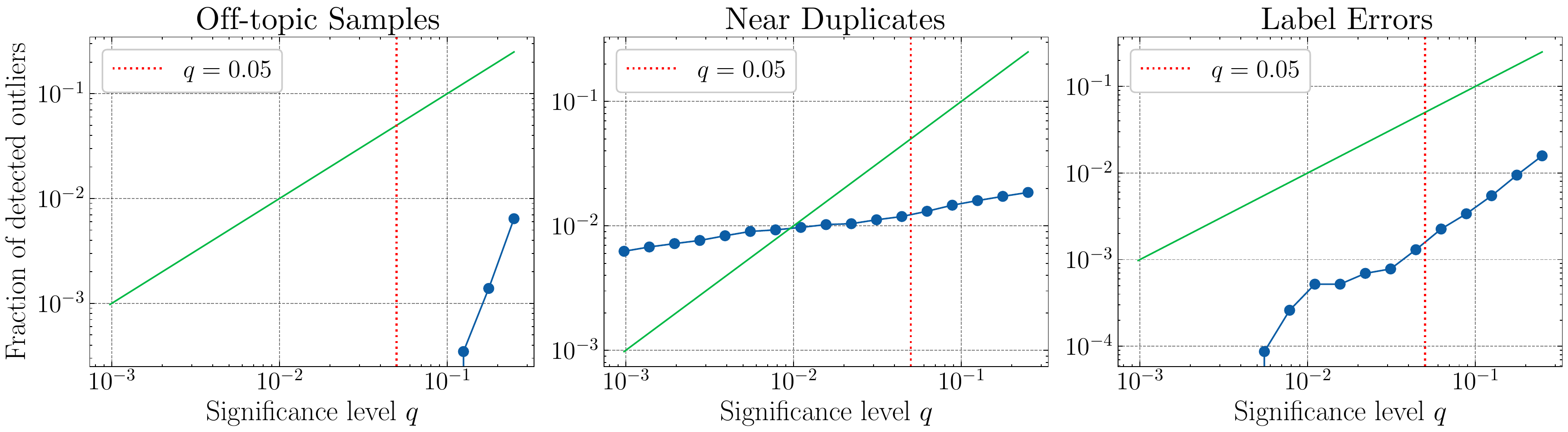}
        \caption{HAM10000 \citep{tschandl_ham10000_2018}}
    \end{subfigure}
    \begin{subfigure}[b]{\textwidth}
        \centering
        \includegraphics[width=\textwidth]{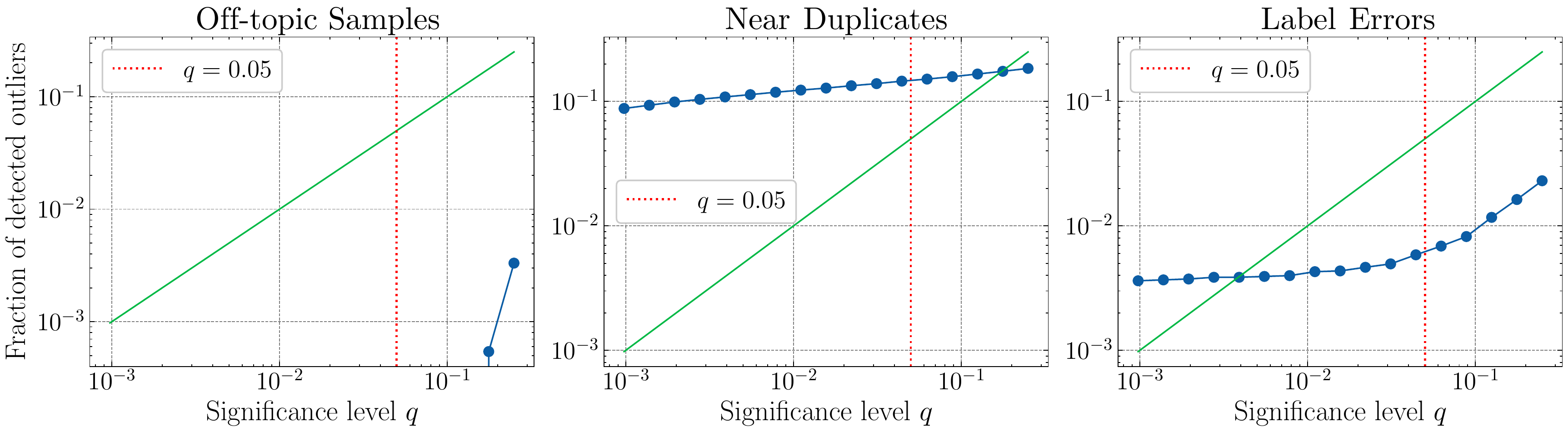}
        \caption{Fitzpatrick17k \citep{groh_evaluating_2021}}
    \end{subfigure}
    \begin{subfigure}[b]{\textwidth}
        \centering
        \includegraphics[width=\textwidth]{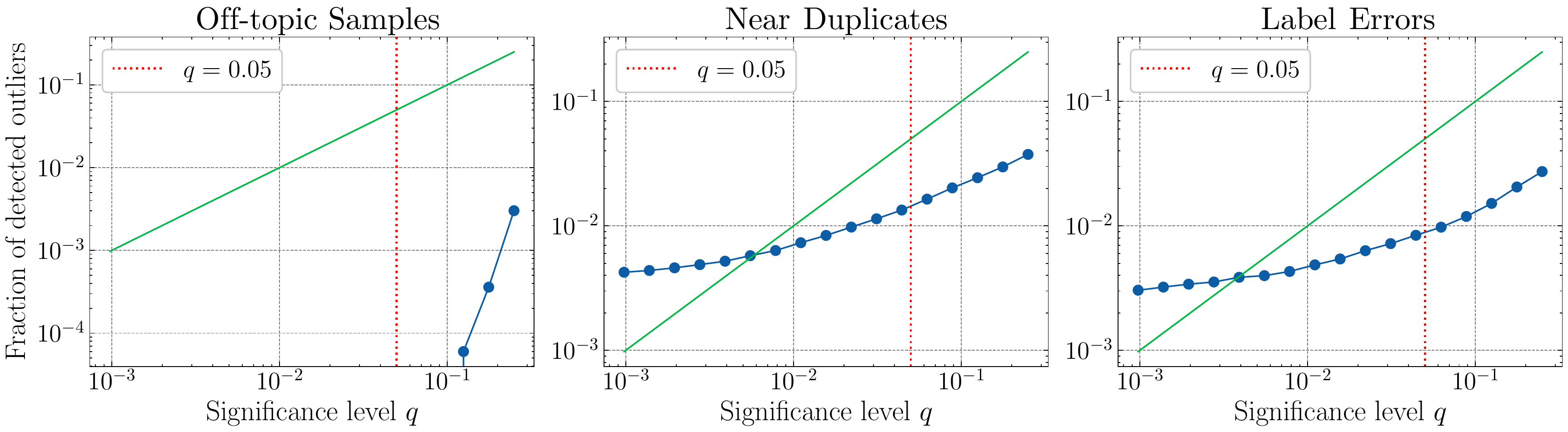}
        \caption{Food-101N \citep{lee_cleannet_2018}}
    \end{subfigure}
    \caption{
            Impact of the choice of the significance level~$q$
            on the fraction of detected data quality issues,
            across issue types and for representative datasets,
            for a fixed contamination rate guess $\alpha=0.10$.
            The observed dependency on $q$ is reported in blue,
            and it is below the lower margin of the plots
            when no problematic samples are found.
            The diagonal green solid line is just a reference to guide reading,
            and the dotted red line indicates the default choice $q=0.05$.
            }
    \label{fig:ThresholdSensitivity}
\end{figure}

In figure \ref{fig:ThresholdSensitivity} we report
the fraction of detected problematic samples as a function of the significance level $q$, for all considered issue types and representative datasets.
We can see that this hyperparameter essentially determines the number of outliers found, which is monotonically increasing with~$q$.
Moreover, the number of identified issues has, in most cases, a dependence on $q$ which is less than linear.
In some cases, especially when the number of detected outliers is below percent level or $q$ approaches $1$, we see more severe sensitivity to the specific value.
This may be because the empirical score distribution changes more abruptly than estimated by the logistic fit, as happens for off-topic samples, or because the region immediately below the lower score cutoff $\bar{s}(\alpha_1)$ (which corresponds to $q=1$) is densely populated almost by construction.
It is however clear that $q$ regulates how extreme scores need to be for a sample to be considered problematic.
A value of $q=10^{-3}$ will only select very apparent data quality issues, $q=1/4$ will almost certainly also include a significant fraction of valid samples, and the choice of $q=0.05$ strikes a compromise between precision and recall.

\newpage
\section{Inspection of benchmark datasets}
\label{app:Benchmark-Inspect}
This section reports the results of auditing multiple vision benchmarks using \textsc{SelfClean}.
In section \ref{app:Detailed-BenchmarkIssues}, we estimate the number of issues in fully automatic mode.
Sections thereafter illustrate the rankings by visualizing the top 15 samples of each issue type, namely off-topic samples, near duplicates, and label errors.
General conclusions are drawn in section \ref{sec:discussion}, while here we report more specific observations.

\begin{center}
    \begin{figure}[htbp]
      \centering
      \includegraphics[width=0.95\linewidth]{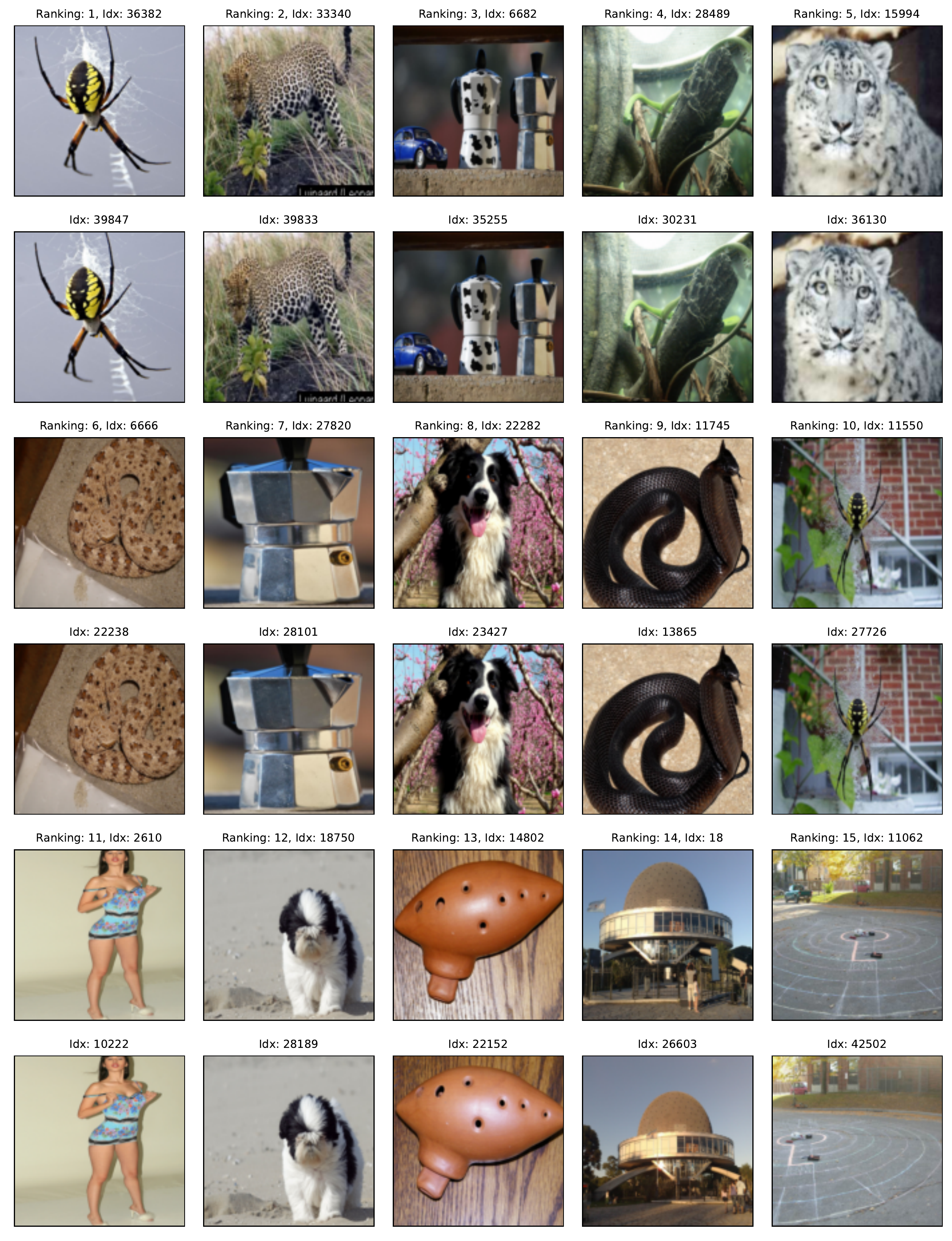}
      \caption{
        Ranking produced by \textsc{SelfClean} for near duplicates in the ImageNet-1k validation set, of which the top 15 are shown along with the respective rank and index.
    }
      \label{fig:ImageNet-Duplicates}
    \end{figure}
\end{center}

\begin{figure}[htbp]
  \centering
  \includegraphics[width=0.95\linewidth]{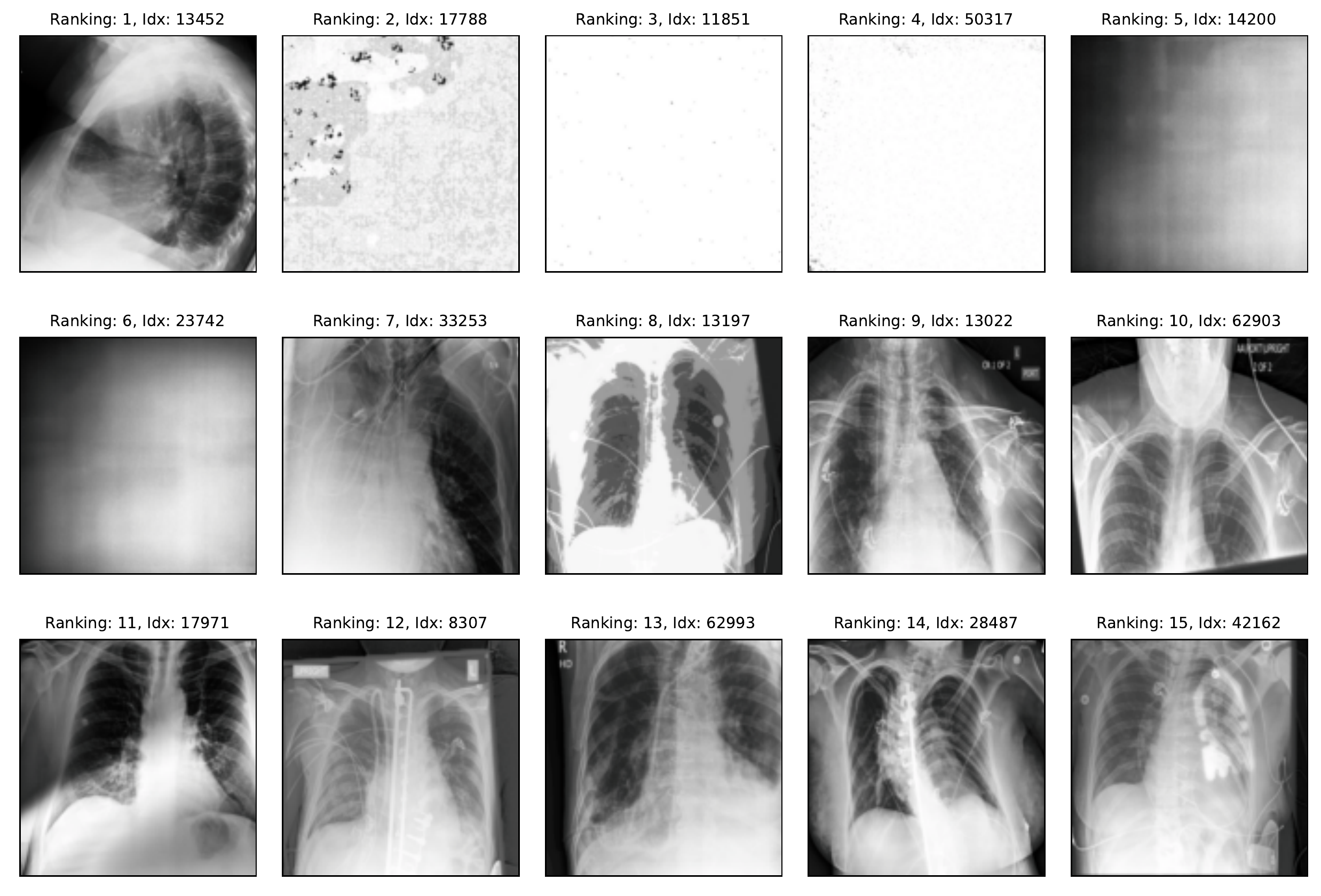}
  \caption{
    Ranking produced by \textsc{SelfClean} for off-topic samples in CheXpert, of which the top 15 are shown along with the respective rank and index.
}
  \label{fig:Chexpert-Irrelevant}
\end{figure}

\begin{figure}[htbp]
  \centering
  \includegraphics[width=0.95\linewidth]{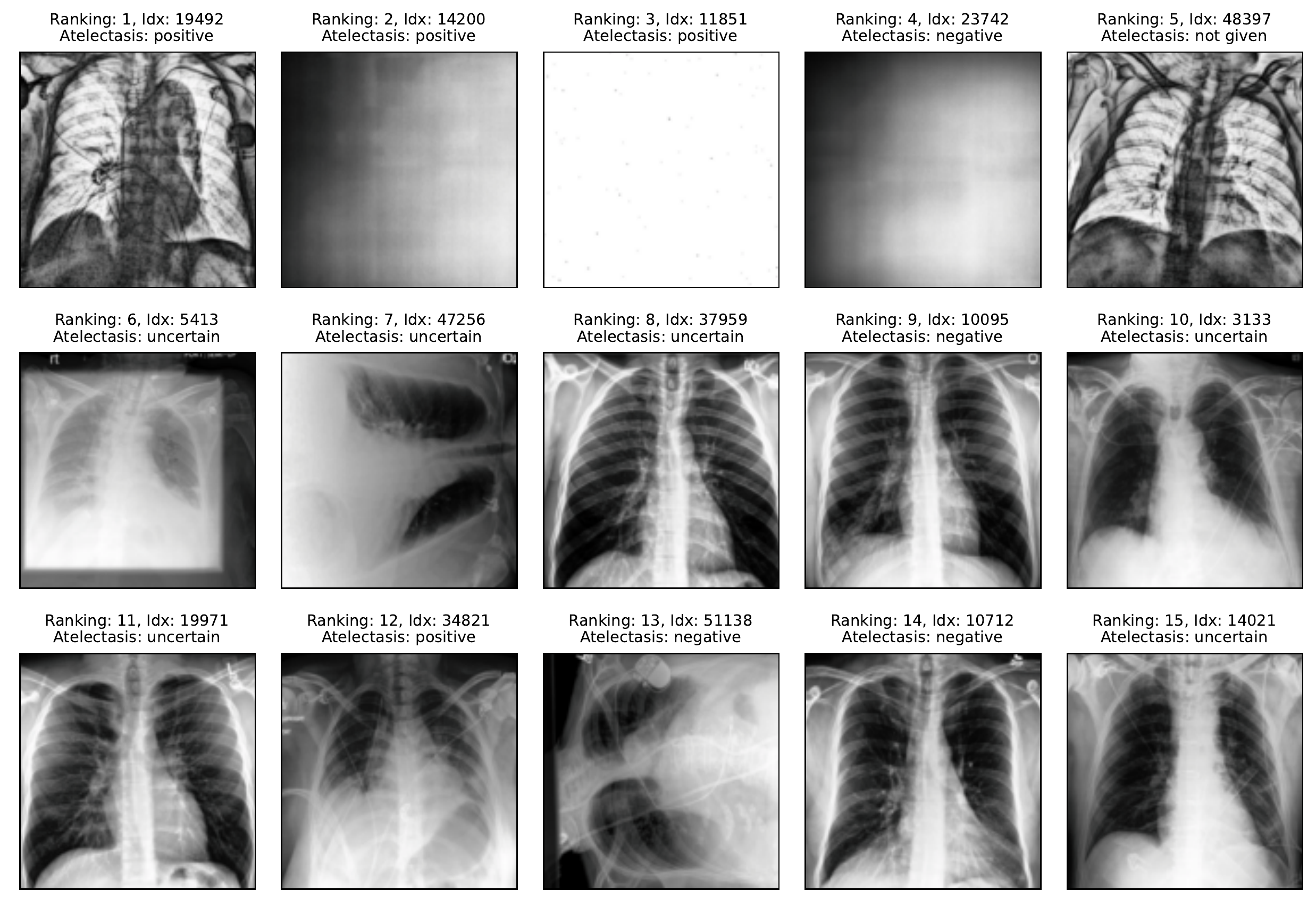}
  \caption{
    Ranking produced by \textsc{SelfClean} for atelectasis label errors in CheXpert, of which the top 15 are shown along with the respective rank, index, and original label.
}
  \label{fig:Chexpert-LabelError}
\end{figure}

\begin{center}
    \begin{figure}[htbp]
      \centering
      \includegraphics[width=0.95\linewidth]{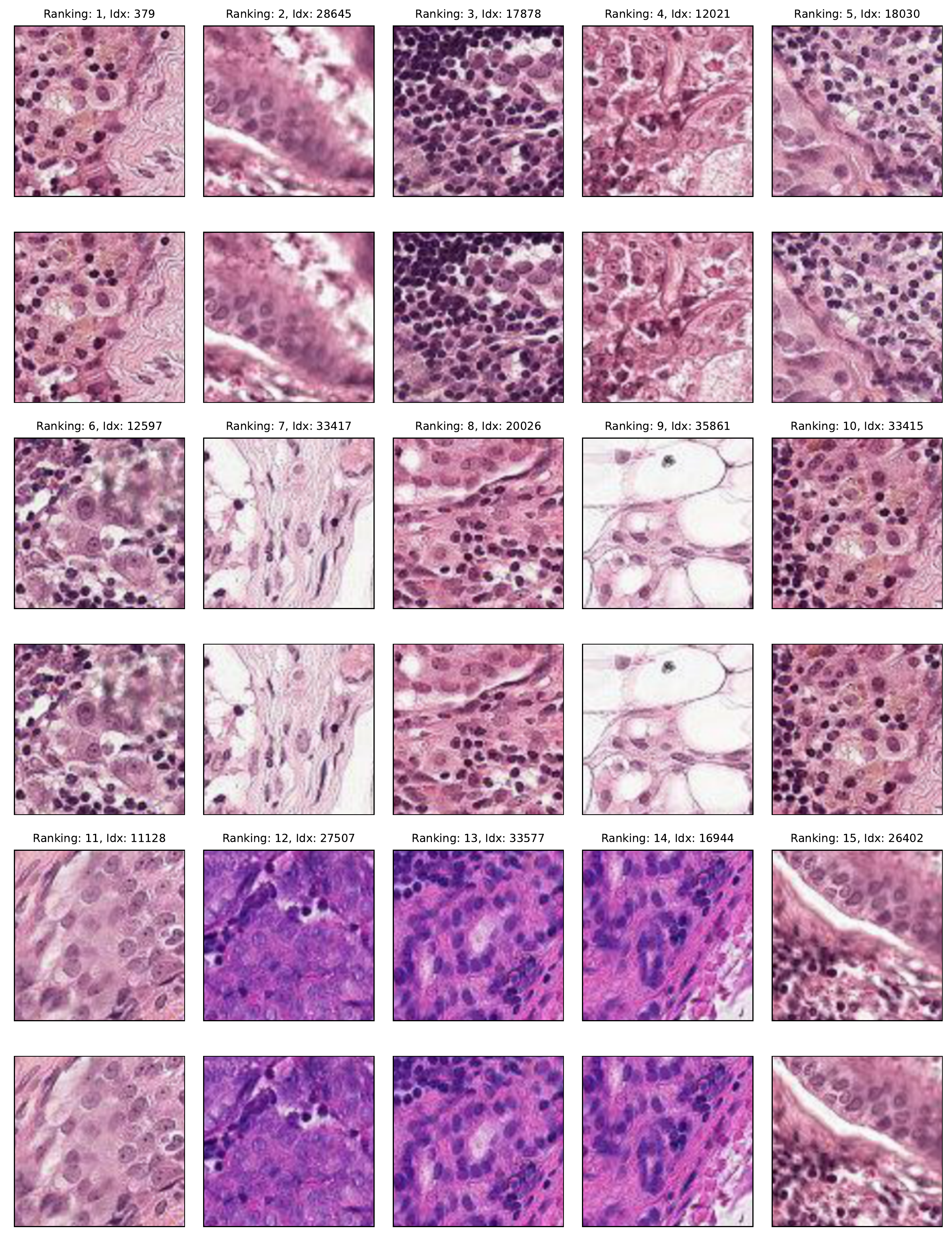}
      \caption{
        Ranking produced by \textsc{SelfClean} for near duplicates in PatchCamelyon, of which the top 15 are shown along with the respective rank and index.
      }
      \label{fig:PatchCamelyon-Duplicates}
    \end{figure}
\end{center}

\begin{figure}[htbp]
  \centering
  \includegraphics[width=0.95\linewidth]{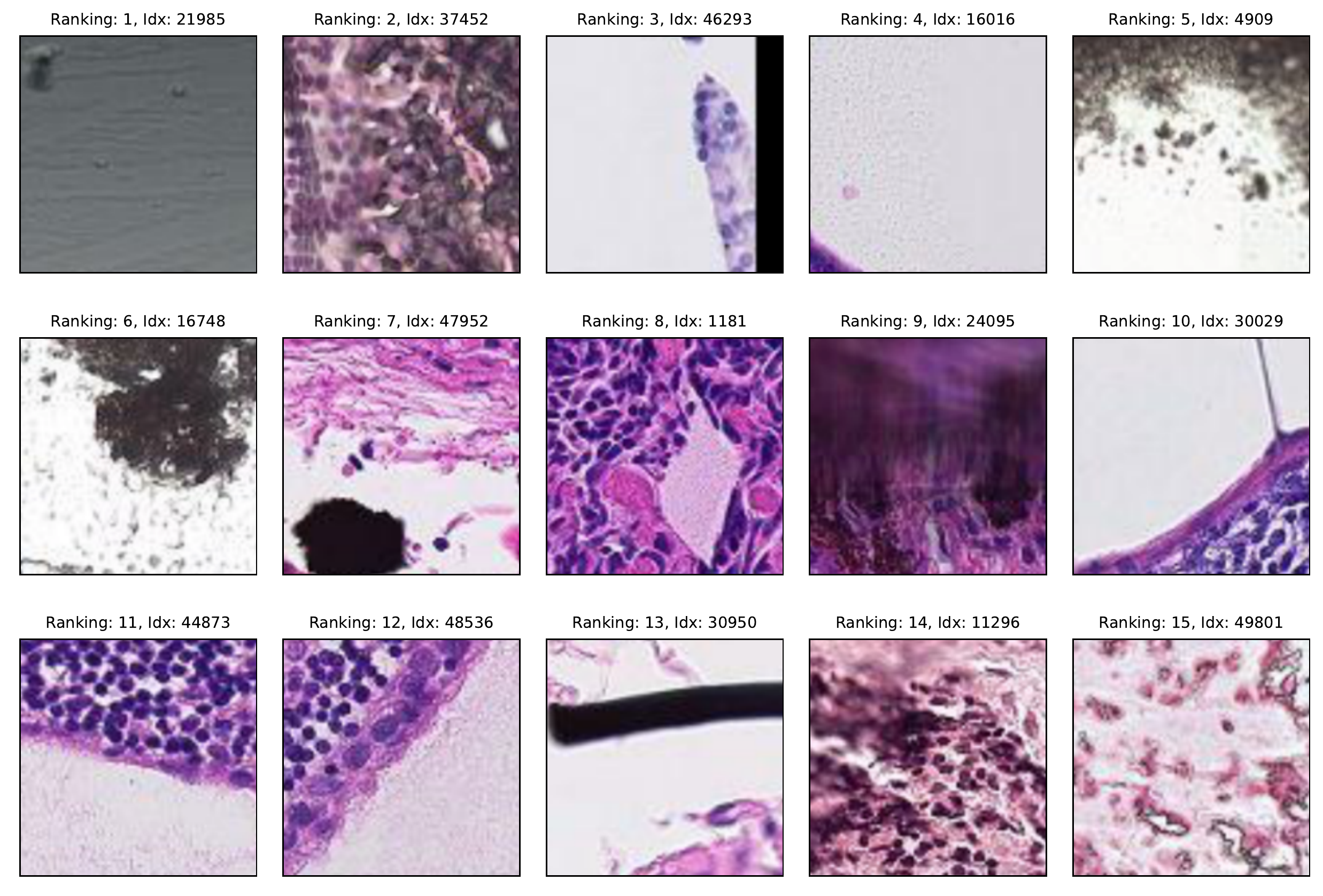}
  \caption{
    Ranking produced by \textsc{SelfClean} for off-topic samples in PatchCamelyon, of which the top 15 are shown along with the respective rank and index.
}
  \label{fig:PatchCamelyon-Irrelevant}
\end{figure}

\begin{center}
    \begin{figure}[htbp]
      \centering
      \includegraphics[width=0.95\linewidth]{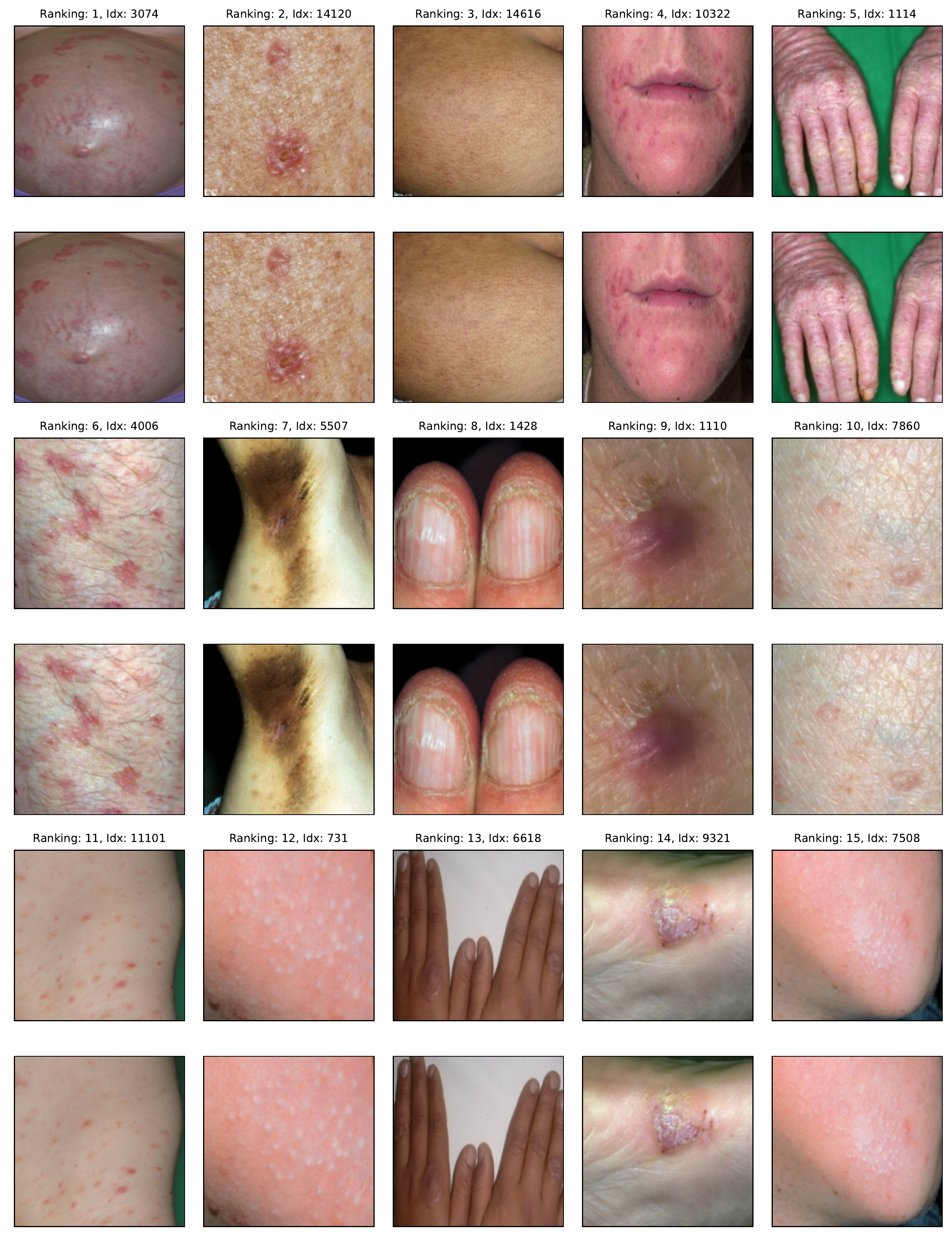}
      \caption{
        Ranking produced by \textsc{SelfClean} for near duplicates in the Fitzpatrick17k, of which the top 15 are shown along with the respective rank and index.
      }
      \label{fig:Fitzpatrick17k-Duplicates}
    \end{figure}
\end{center}

\begin{figure}[htbp]
  \centering
  \includegraphics[width=0.95\linewidth]{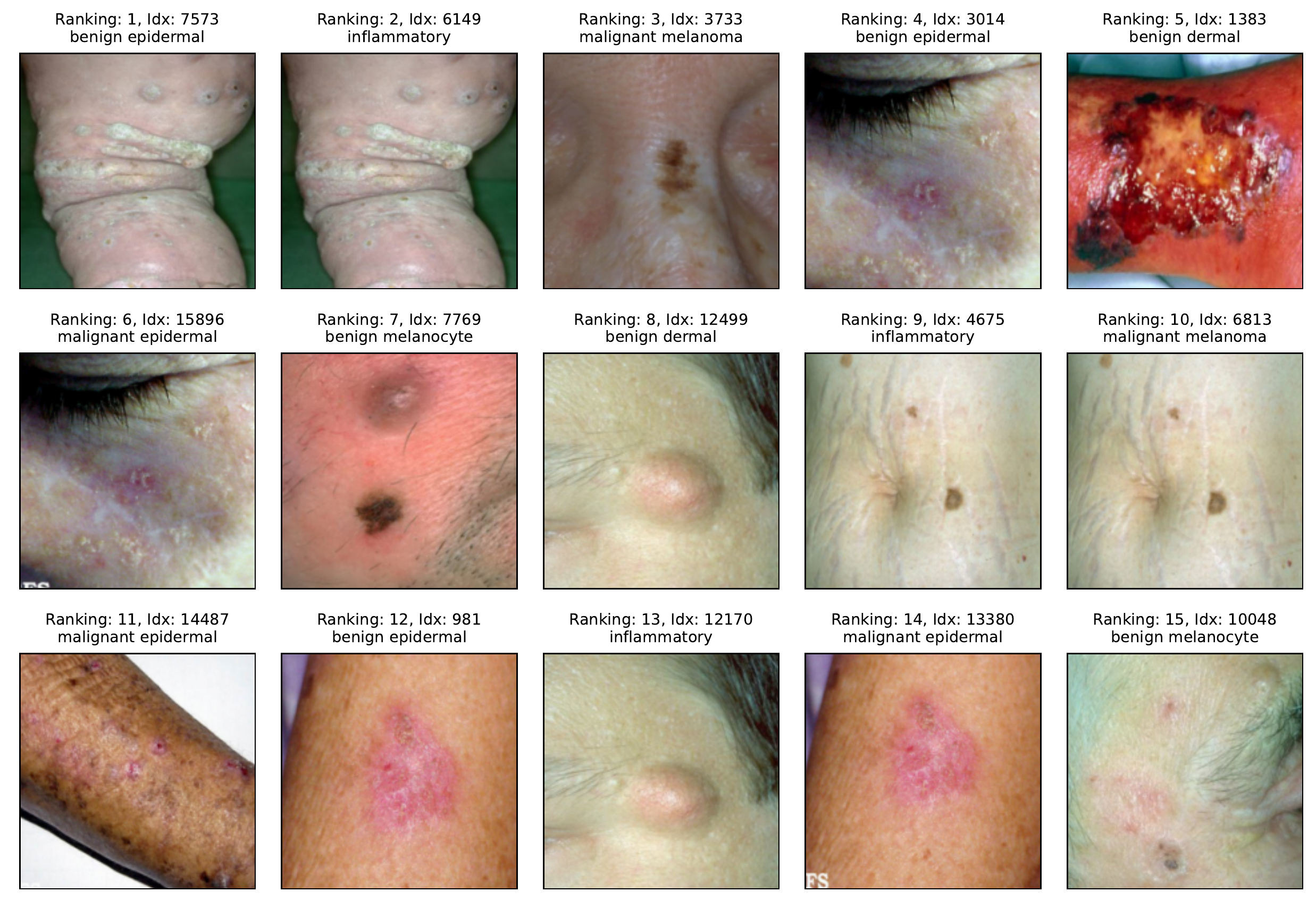}
  \caption{
    Ranking produced by \textsc{SelfClean} for label errors in the Fitzpatrick17k, of which the top 15 are shown along with the respective rank, index, and original label.
}
  \label{fig:Fitzpatrick17k-LabelError}
\end{figure}

\looseness=-1
Applying \textsc{SelfClean} to multiple benchmark datasets across different domains has led to different insights on why some of these data quality issues may occur.
Off-topic samples in the medical domain are often caused by device malfunctions, wrong configurations, tests, or other scanning errors (figure~\ref{fig:Chexpert-Irrelevant} Rank 2-8 and figure~\ref{fig:PatchCamelyon-Irrelevant} Rank 1, 5, and 9).
Near duplicates can often be traced to data acquisition problems, such as crawling both an image and its thumbnail (figure~\ref{fig:Fitzpatrick17k-Duplicates}) or the metadata failing to correctly flag that two images have a common origin (figure~\ref{fig:ImageNet-Duplicates} and \ref{fig:PatchCamelyon-Duplicates}).
The most apparent label errors are often near-duplicate samples with different labels (figure~\ref{fig:Fitzpatrick17k-LabelError} Rank 1\&2, 4\&6, 9\&10, and 8\&13), which indicate (understandable) difficulties in the annotation process, or off-topic samples with a label (figure~\ref{fig:Chexpert-LabelError} Rank 2-4), which easily arise in (semi-)automated annotation procedures.

\subsection{Estimation of issues in benchmark datasets}
\label{app:Detailed-BenchmarkIssues}

\begin{table*}[htbp]
    \centering
    \scriptsize
    \caption{
    Estimated percentage of data quality issues in vision benchmarks obtained using \textsc{SelfClean}'s automatic mode with $\alpha = 0.10$ and $q=0.05$.
    Images marked as originating from the same person, patient or lesion were excluded from the near-duplicate count whenever available.
    }
    \label{tab:EstimatedErrorsAppendix}
    \begin{tabular}{lr  rrr r}
        \toprule
        & & \multicolumn{3}{c}{\bfseries Estimated Issues} &  \\
        \cmidrule(lr){3-5}
        
        \bfseries Dataset &
        \bfseries Size &
        Off-topic Samples & Near Duplicates & Label Errors 
        & \multicolumn{1}{c}{\bfseries Total} \\
        \midrule
        \multicolumn{4}{l}{\textit{Medical Images}} \\
        DDI
            & 656
            & $1 \ (0.2\%)$ & $4 \ (0.6\%)$ & $5 \ (0.8 \%)$ 
            & $10 \ (1.5\%)$\\
        PAD-UFES-20
            & 2,298
            & $0 \ (0.0\%)$ & $0 \ (0.0\%)$ & $5 \ (0.4\%)$
            & $5 \ (0.4\%)$\\
        HAM10000
            & 11,526
            & $0 \ (0.0\%)$ & $1 \ (<\!0.1\%)$ & $17 \ (0.2\%)$ 
            & $18 \ (0.2\%)$\\
        VinDr-BodyPartXR
            & 16,086
            & $263 \ (1.6\%)$ & $20 \ (0.1\%)$ & $74 \ (0.5\%)$ 
            & $357 \ (2.2\%)$\\
        Fitzpatrick17k
            & 16,574
            & $18 \ (0.1\%)$ & $2{,}446 \ (14.8\%)$ & $103 \ (0.6\%)$ 
            & $2{,}567 \ (15.5\%)$\\
        ISIC-2019
            & 33,569
            & $0 \ (0.0\%)$ & $1{,}200 \ (3.6\%)$ & $97 \ (0.3\%)$
            & $1{,}297 \ (3.9\%)$\\
        CheXpert\footnotemark[6]
            & 223,414
            & $6 \ (<\!0.1\%)$ & $0 \ (0.0\%)$ & $303 \ (0.1\%)$ 
            & $309 \ (0.1\%)$\\
        PatchCamelyon
            & 327,680
            & $98 \ (<\!0.1\%)$ & $12{,}845 \ (3.9\%)$ & $589 \ (0.2\%)$ 
            & $13{,}532 \ (4.1\%)$\\
        \midrule
        \multicolumn{4}{l}{\textit{General Images}} \\
        STL-10
            & 5,000
            & $0 \ (0.0\%)$ & $7 \ (0.1\%)$ & $21 \ (0.4\%)$ 
            & $28 \ (0.5\%)$\\
        ImageNet-1k Validation
            & 50,000
            & $0 \ (0.0\%)$ & $36 \ (0.1\%)$ & $262 \ (0.5\%)$ 
            & $298 \ (0.6\%)$\\
        CelebA
            & 202,599
            & $2 \ (<\!0.1\%)$ & $810 \ (0.4\%)$ & $1{,}033 \ (0.5\%)$ 
            & $1{,}845 \ (0.9\%)$\\
        Food-101N
            & 310,009
            & $310 \ (0.1\%)$ & $4{,}433 \ (1.4\%) $ & $2{,}728 \ (0.9\%)$ 
            & $7{,}471 \ (2.4\%)$\\
        \bottomrule
    \end{tabular}
\end{table*}

\footnotetext[6]{Label errors refer to atelectasis detection only since the classification task admits multiple labels, and expert agreement is the highest for this condition.}

\newpage
\subsection{ImageNet-1k}

\begin{figure}[htbp]
  \centering
  \includegraphics[width=0.95\linewidth]{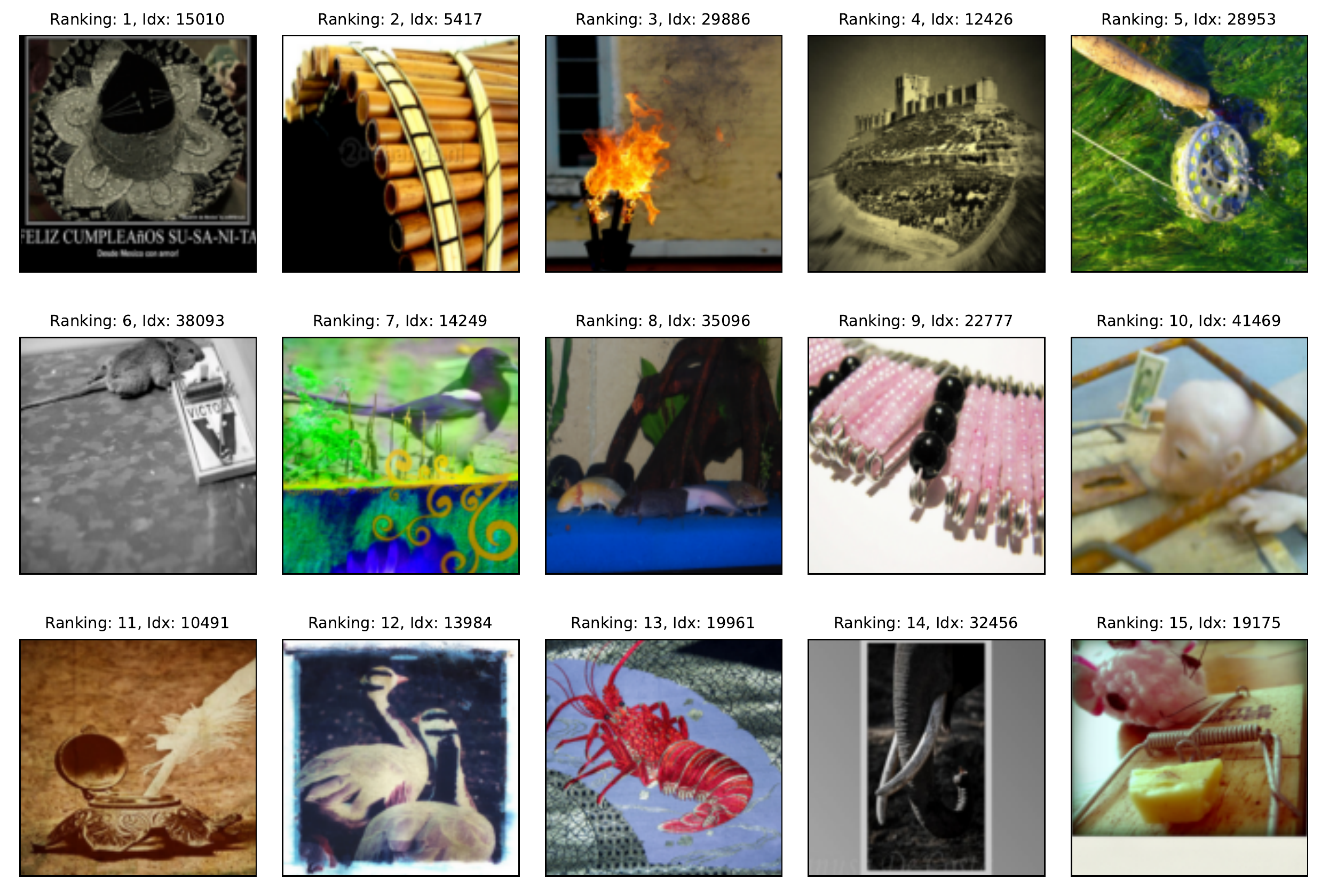}
  \caption{
    Ranking produced by \textsc{SelfClean} for off-topic samples in the ImageNet-1k validation set, of which the top 15 are shown along with the respective rank and index.
}
  \label{fig:ImageNet-Irrelevant}
\end{figure}

\begin{figure}[htbp]
  \centering
  \includegraphics[width=0.95\linewidth]{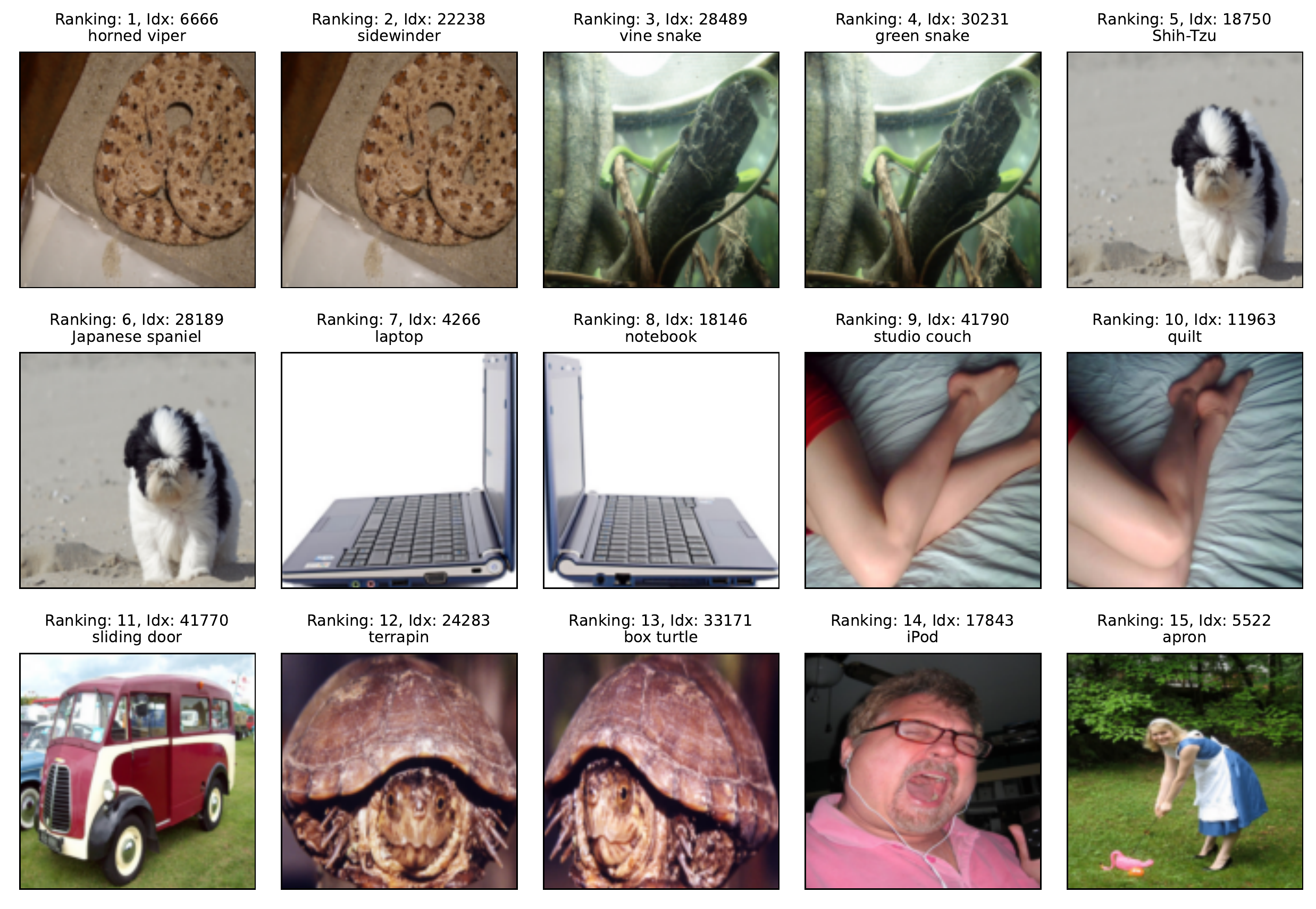}
  \caption{
    Ranking produced by \textsc{SelfClean} for label errors in the ImageNet-1k validation set, of which the top 15 are shown along with the respective rank, index, and original label.
}
  \label{fig:ImageNet-LabelError}
\end{figure}

\newpage
\subsection{CheXpert}

\begin{center}
    \begin{figure}[htbp]
      \centering
      \includegraphics[width=0.95\linewidth]{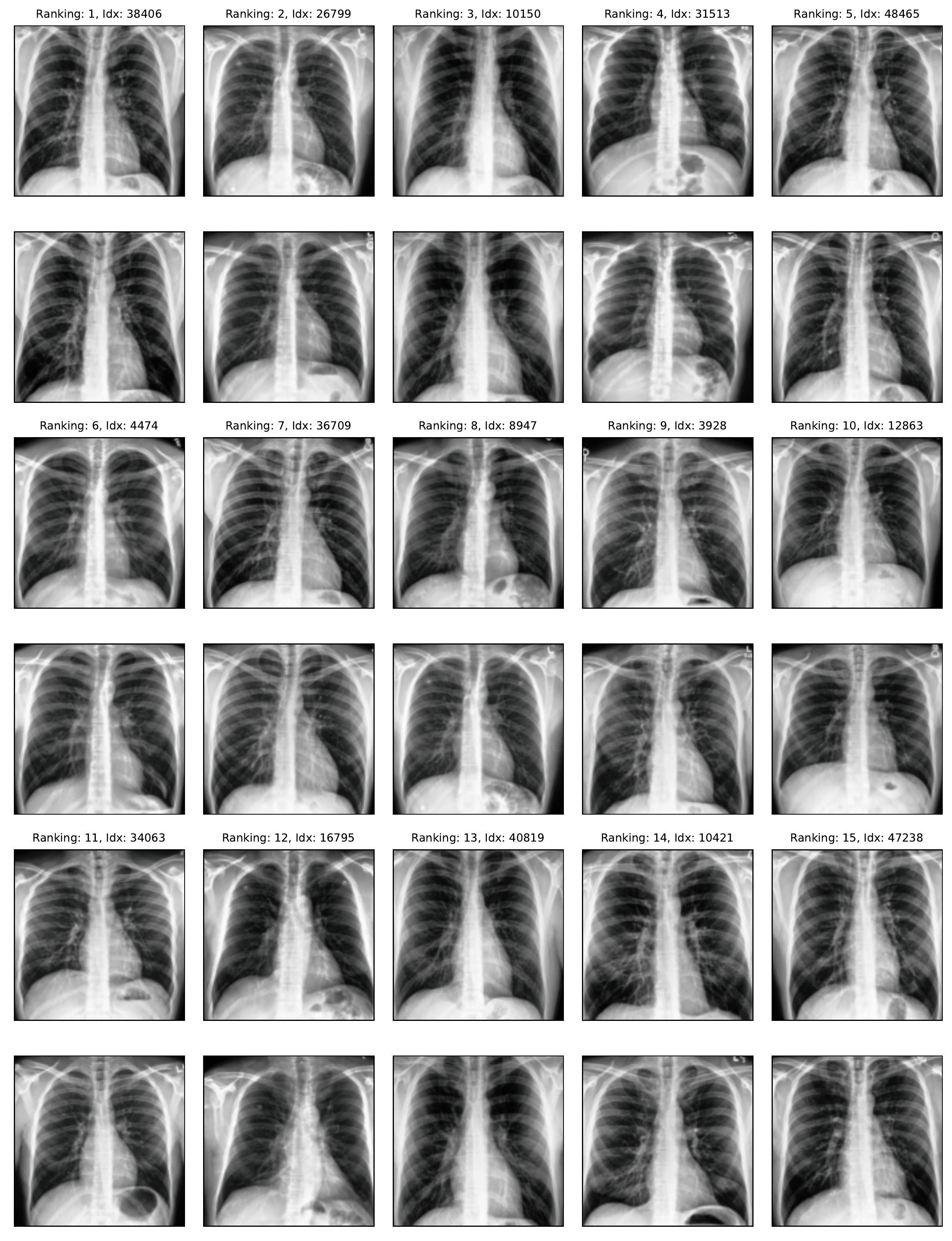}
      \caption{
        Ranking produced by \textsc{SelfClean} for near duplicates in CheXpert, of which the top 15 are shown along with the respective rank and index.
    }
      \label{fig:Chexpert-Duplicates}
    \end{figure}
\end{center}

\newpage
\subsection{PatchCamelyon}

\begin{figure}[htbp]
  \centering
  \includegraphics[width=0.95\linewidth]{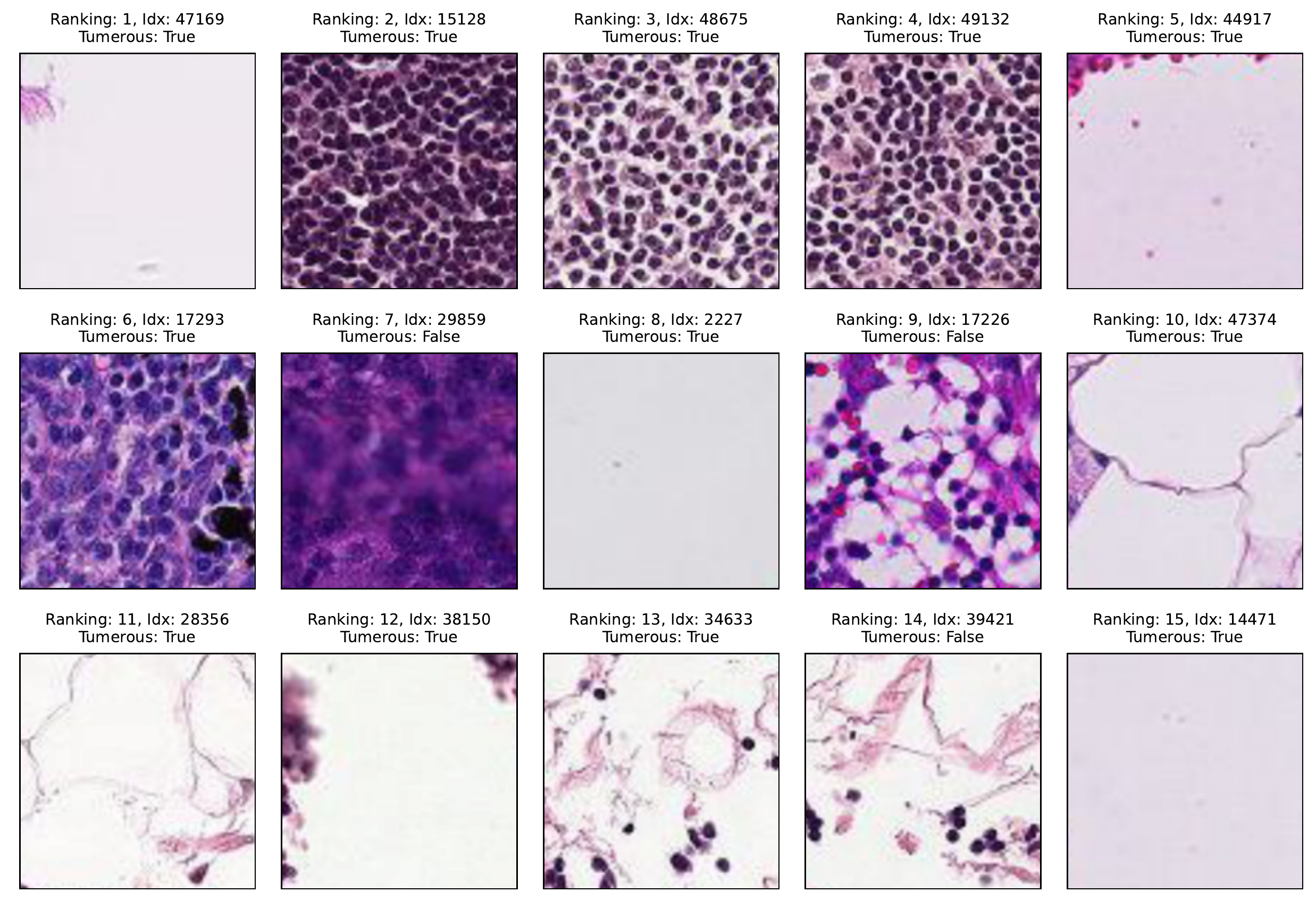}
  \caption{
    Ranking produced by \textsc{SelfClean} for label errors in PatchCamelyon, of which the top 15 are shown along with the respective rank, index, and original label, i.e., if the patch is tumerous.
}
  \label{fig:PatchCamelyon-LabelError}
\end{figure}

\newpage
\subsection{Fitzpatrick17k}

\begin{figure}[htbp]
  \centering
  \includegraphics[width=0.95\linewidth]{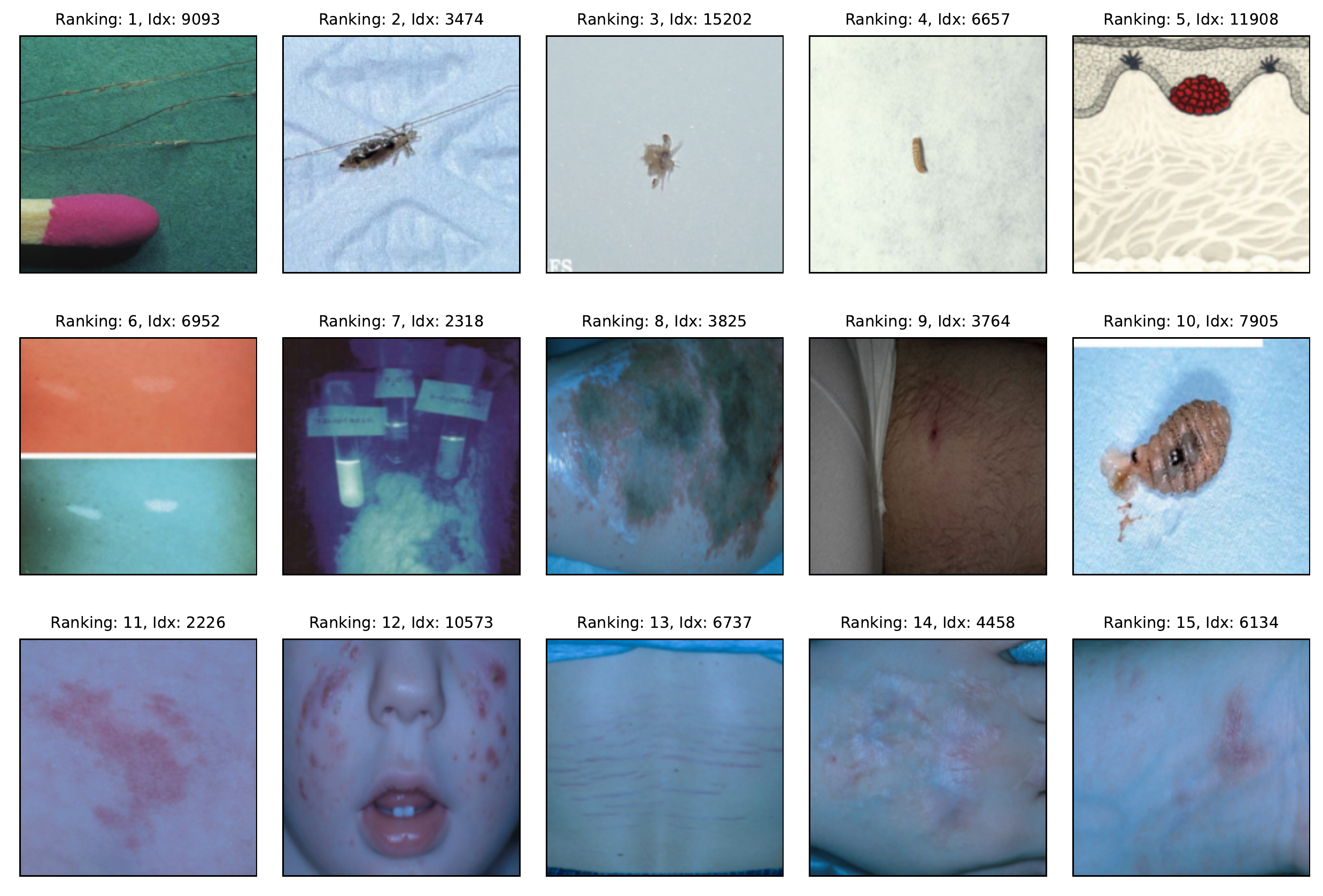}
  \caption{
    Ranking produced by \textsc{SelfClean} for off-topic samples in the Fitzpatrick17k, of which the top 15 are shown along with the respective rank and index.
}
  \label{fig:Fitzpatrick17k-Irrelevant}
\end{figure}

\end{document}


\frenchspacing

\maketitle

\section{Code Availability}
The implementation of \textsc{SelfClean} can be found at
\begin{center}
    \url{https://anonymous.4open.science/r/SelfClean-C961}.
\end{center}
Additionally to the implementation, the repository contains two example notebooks where \mbox{\textsc{SelfClean}} is applied to Imagenette and Oxford-IIIT Pet.

The code for the evaluation of \textsc{SelfClean} can be found at
\begin{center}
    \url{https://anonymous.4open.science/r/SelfClean-Evaluation-0E0F}.
\end{center}